\documentclass[10pt,journal,compsoc]{IEEEtran}
\usepackage{graphicx}

\usepackage{cite}

\pdfpxdimen=\dimexpr 1 in/192\relax  %
\usepackage{calc}

\newif\ifarxiv
\arxivtrue

\def\MYTITLE{Event-based Photometric Bundle Adjustment}

\usepackage{amsmath,amssymb,amsfonts}
\usepackage{cancel}

\usepackage{caption}
\usepackage{subcaption}
\usepackage{soul} %

\renewcommand{\paragraph}[1]{
 \textbf{#1.} 
}

\usepackage[dvipsnames]{xcolor}

\usepackage{threeparttable} %
\usepackage{makecell}
\usepackage{multirow}
\usepackage{rotating}

\usepackage{siunitx}
\usepackage{adjustbox} %

\usepackage{url}
\definecolor{myblue}{rgb}{0.12,0.49,0.85}

\def\pol{s} %
\def\prtl#1#2{\frac{\partial#1}{\partial#2}}

\def\numEvents{N_e} %
\def\numPixels{N_p} %
\def\numPoses{N_{\text{poses}}} %
\def\Warp{\mathbf{W}}

\def\bx{\mathbf{x}}

\def\Rot{\mathtt{R}}

\def\angvel{\boldsymbol{\omega}} %
\def\mP{\mathtt{P}} %

\def\imgpoint{\bx}

\def\bS{\mathbf{S}}

\def\bz{\mathbf{z}}

\def\mJ{\mathtt{J}}
\def\be{\mathbf{e}}%
\def\bb{\mathbf{b}}%
\def\Kint{\mathtt{K}}%
\def\balpha{\boldsymbol{\alpha}}%
\def\bbeta{\boldsymbol{\beta}}%
\def\rotperturb{\boldsymbol{\delta\phi}}%
\def\mE{\mathtt{E}}%
\def\br{\mathbf{r}}%
\def\bc{\mathbf{c}}%
\def\bq{\mathbf{q}}%

\def\Real{\mathbb{R}} %

\def\mA{\mathtt{A}}

\def\bp{\mathbf{p}}
\def\bP{\mathbf{P}}

\def\bX{\mathbf{X}}

\def\loss{g} %

\def\NA{\text{N/A}}

\def\emptycol{\hspace{-3ex}} %

\newcommand{\bnum}[1]{\bfseries #1}
\newcommand{\novalue}{{\textendash}}

\definecolor{light-gray}{gray}{0.6}
\newcommand\gframe[1]{{\color{light-gray}\frame{#1}}}
\newcommand\gframeRed[1]{{\color{red}\frame{#1}}}
\newcommand\gframeGreen[1]{{\color{green}\frame{#1}}}
\newcommand\gframeBlue[1]{{\color{blue}\frame{#1}}}
\newcommand\gframeYellow[1]{{\color{yellow}\frame{#1}}}

\def\playroom{\emph{playroom}}
\def\bicycle{\emph{bicycle}}
\def\city{\emph{city}}
\def\street{\emph{street}}
\def\town{\emph{town}}
\def\bay{\emph{bay}}

\def\shapes{\emph{shapes}}
\def\poster{\emph{poster}}
\def\boxes{\emph{boxes}}
\def\dynamic{\emph{dynamic}}
\def\sliderfar{\emph{slider\_far}}
\def\sliderdepth{\emph{slider\_depth}}

\def\bicycles{\emph{bicycles}}
\def\building{\emph{building}}
\def\miscellany{\emph{miscellany}}
\def\staircase{\emph{staircase}}

\def\building{\emph{building}}

\def\crossroad{\emph{crossroad}}
\def\atrium{\emph{atrium}}

\def\esmt{EKF-SMT}
\def\psmt{PF-SMT}
\def\rtpt{RTPT}
\def\cmaxgae{CMax-GAE}
\def\cmaxw{CMax-$\angvel$}
\def\cmaxslam{CMax-SLAM}
\def\EMBA{EMBA}

\def\LM{Levenberg-Marquardt}
\def\EPBA{EPBA}

\usepackage{pifont}%
\newcommand{\yesmark}{\ding{51}}%
\newcommand{\nomark}{\ding{55}}%

\ifarxiv
\usepackage[breaklinks=true,bookmarks=false,colorlinks,citecolor=myblue]{hyperref}
\else
\usepackage[breaklinks=true,bookmarks=false,colorlinks=false,citecolor=black,hidelinks]{hyperref}
\fi
\hypersetup{
  pdftitle={\MYTITLE},
  pdfsubject={Computer Vision, Robotics},
  pdfauthor={Shuang Guo, Guillermo Gallego},
  pdfkeywords={Event Cameras, Ego-motion, Asynchronous Sensor, High Temporal Resolution}
}
\usepackage{booktabs}

\ifarxiv
\usepackage{orcidlink}
\makeatletter
\long\def\@IEEEtitleabstractindextextbox#1{\parbox{0.922\textwidth}{#1}}
\makeatother
\fi

\author{Shuang~Guo\ifarxiv\orcidlink{0000-0002-0142-0678}\fi~and Guillermo~Gallego\ifarxiv\orcidlink{0000-0002-2672-9241}\fi%
\IEEEcompsocitemizethanks{
\IEEEcompsocthanksitem The authors are with the Dept.~of Electrical Engineering and Computer Science, Technische Universit\"at Berlin, and with the Robotics Institute Germany, Berlin, Germany. 
G. Gallego is also with the Science of Intelligence (SCIoI) Excellence Cluster and with the Einstein Center Digital Future (ECDF), Berlin, Germany. 
\IEEEcompsocthanksitem Funded by the Deutsche Forschungsgemeinschaft (DFG, German Research Foundation) under Germany’s Excellence Strategy – EXC 2002/1 ``Science of Intelligence'' – project number 390523135.
\ifarxiv
\else
\IEEEcompsocthanksitem 
Project page with code: \url{https://github.com/tub-rip/epba}.%
\fi
}%
}

\usepackage[capitalize]{cleveref}
\newif\ifaisy
\aisyfalse %
\ifaisy
\crefname{section}{Section}{Sections}
\crefname{table}{Table}{Tables}
\crefname{figure}{Figure}{Figures}
\else 
\crefname{section}{Sec.}{Secs.}
\crefname{table}{Tab.}{Tabs.} 
\crefname{figure}{Fig.}{Figs.}
\fi
\Crefname{section}{Section}{Sections}
\Crefname{table}{Table}{Tables}
\Crefname{figure}{Figure}{Figures}

\newif\ifclearsectionlook
\clearsectionlookfalse

\ifarxiv
\usepackage[absolute]{textpos}
\fi

\begin{document}
\title{\MYTITLE}

\ifarxiv
\definecolor{somegray}{gray}{0.5}
\newcommand{\darkgrayed}[1]{\textcolor{somegray}{#1}}
\begin{textblock}{11}(2.5, 0.4)
\begin{center}
\darkgrayed{This paper has been accepted for publication at the\\
IEEE Transactions on Pattern Analysis and Machine Intelligence, 2025.
\copyright IEEE}
\end{center}
\end{textblock}
\fi

\IEEEtitleabstractindextext{%
\begin{abstract}
We tackle the problem of bundle adjustment (i.e., simultaneous refinement of camera poses and scene map) for a purely rotating event camera.
Starting from first principles, we formulate the problem as a classical non-linear least squares optimization. 
The photometric error is defined using the event generation model directly in the camera rotations and the semi-dense scene brightness that triggers the events.
We leverage the sparsity of event data to design a tractable \LM{} solver that handles the very large number of variables involved.
To the best of our knowledge, our method, which we call Event-based Photometric Bundle Adjustment (\EPBA{}), is the first event-only photometric bundle adjustment method that works on the brightness map directly and exploits the space-time characteristics of event data, without having to convert events into image-like representations.
Comprehensive experiments on both synthetic and real-world datasets demonstrate \EPBA{}'s effectiveness in decreasing the photometric error (by up to 90\%), 
yielding results of unparalleled quality.
The refined maps reveal details that were hidden using prior state-of-the-art rotation-only estimation methods.
The experiments on modern high-resolution event cameras show the applicability of \EPBA{} to panoramic imaging in various scenarios (without map initialization, at multiple resolutions, and in combination with other methods, such as IMU dead reckoning or previous event-based rotation estimation methods).
We make the source code publicly available.
\end{abstract}

}
\maketitle
\IEEEdisplaynontitleabstractindextext

\ifclearsectionlook \tableofcontents \fi

\ifarxiv
\section*{Video and Source Code}
Project page: \url{https://github.com/tub-rip/epba}.
\fi

\ifclearsectionlook\cleardoublepage\fi \section{Introduction}
\label{sec:intro}

\IEEEPARstart{B}{undle} Adjustment (BA) is the problem of jointly refining the sensor motion and the reconstructed scene map that best fit the acquired visual data in terms of an objective function (e.g., reprojection or photometric error) \cite{Triggs00,Alismail16accv}.
Refinement redistributes the errors among all variables of the problem, increasing consistency and robustness.
Hence, it is a paramount topic in photogrammetry, robotics and computer vision, enabling accurate positioning and measurement technology with various sensors (cameras, LiDARs, etc.).
It finds multiple applications, such as geodetic mapping \cite{Agarwal14ram}, image stitching (e.g., mosaicing) \cite{Brown06ijcv}, visual odometry (VO) \cite{Engel17pami}, simultaneous localization and mapping (SLAM) \cite{Kaess12ijrr} and 3D scanning for virtual reality \cite{Furukawa10pami,Tola11mva}. %
BA with standard, frame-based cameras is a developed topic \cite{Hartley03book,Szeliski10book,Wu11cvpr,Klenk19thesis},
but it is inherently limited by the information acquired by the cameras. 
Some of these limitations (e.g., dynamic range, motion blur) can be overcome with novel sensors.

\begin{figure}[t]
    \centering
    \includegraphics[width=\linewidth]{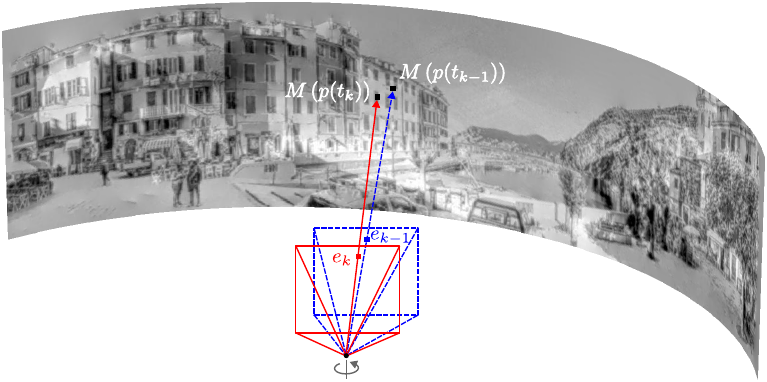}
    \caption{\EPBA{} jointly refines the camera rotations and panoramic intensity map by minimizing a photometric error.
    Each event contributes an error term by computing the difference of the intensities at two nearby map points.
}
\label{fig:eyecatcher}
\end{figure}

Event cameras are novel bio-inspired visual sensors that measure per-pixel brightness changes \cite{Lichtsteiner08ssc,Son17isscc,Finateu20issccShort}, called ``events''. 
In contrast to the images (i.e., frames) produced by standard cameras, the output of an event camera is an asynchronous and sparse stream of data \cite{Posch14ieee} (triggered in the order of thousand or million events per second). 
This working principle endows event cameras with potential advantages (high dynamic range (HDR), high temporal resolution, low latency, low power consumption, etc.) that can be leveraged to overcome difficult scenarios for standard cameras. 
However, new methods are required to unlock such advantages, rethinking computer vision in terms of the space-time nature of the event data \cite{Gallego20pami}.

\begin{figure*}[t]
    \centering
    \includegraphics[width=0.8\linewidth]{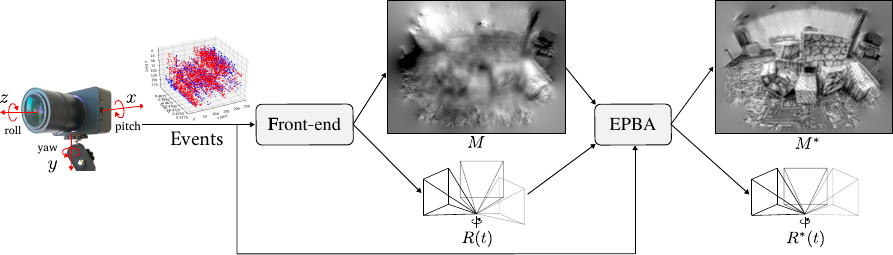}
    \caption{Event-based Photometric Bundle Adjustment (\EPBA{}) can be used as the back-end of a rotational visual odometry / SLAM system to jointly refine the camera rotations $R(t)$ and panoramic intensity map $M$ produced by a front-end.
    }
\label{fig:pipeline}
\end{figure*}

Event-based BA arises in VO/SLAM methods with event cameras (often in the form of sliding-window optimization), 
where joint refinement of the unknowns is needed to improve ego-motion estimation \cite{Rebecq17bmvc,Chin19cvprw,Wang23ral}. %
Adopting frame-based terminology \cite{Engel17pami}, we may categorize event-based BA approaches as feature-based (``indirect'') or photometric-based (``direct'').
However this topic is still in its infancy, as most event-based VO/SLAM systems lack a global refinement step.
Instead, they operate in a parallel-tracking-and-mapping %
manner \cite{Weikersdorfer13icvs,Kim14bmvc,Kim16eccv,Rebecq17ral,Zhou20tro}, with each subsystem relying on the output of the other subsytems as input to work properly. 

Some of the challenges faced by event-based estimation methods are: 
noise (e.g., due to pixel manufacturing mismatch and dynamic effects) \cite{Guo22pami}, 
and data association (i.e., identifying which events are triggered by the same scene point) \cite{Gehrig19ijcv}.
Each event carries little information, and many events need to be observed to collect sufficient information for reliable association.
Additionally, BA poses the challenge of simultaneously estimating correlated variables, 
which is performed in a high-dimensional search space, making optimization costly and prone to local minima.

So far, these challenges have been mostly tackled in a feature-based manner \cite{Rebecq17bmvc,Chin19cvprw,Wang23ral}. 
The solution consists in detecting sparse keypoints in the events (possibly preceded by an events-to-image conversion \cite{Chin19cvprw,Rebecq17bmvc} to reutilize frame-based detectors)
and feeding them to well-known geometric BA backends (e.g., \cite{Triggs00,Leutenegger15ijrr}). 
However, ($i$) this discards the large amount of information contained in the events (as shown by image reconstruction methods \cite{Rebecq19pami,Zhang22pami}) 
and ($ii$) is not yet effective: due to noise and the dependency of events on motion, current event features/keypoints are not as accurate and stable as frame-based ones, so their use in VO/SLAM has been scarce \cite{Kueng16iros,Klenk24threedv}.
Another solution consists in leveraging grayscale information from colocated frames \cite{Rosinol18ral,Hidalgo22cvpr}, 
in which case the BA (feature-based \cite{Rosinol18ral} or photometric-based \cite{Hidalgo22cvpr}) is just borrowed from frame-based systems (i.e., operates on frames, which can suffer from motion blur and low dynamic range).

On the other hand, feature-based approaches have been surpassed by direct methods in frame-based VO \cite{Engel17pami}.
Likewise, event-based direct methods, which take into account the information of each event, not just those that conform to a feature's definition, 
are the state of the art in some motion-estimation tasks \cite{Gallego18cvpr,Nunes21pami,Peng21pami,Gu21iccv}. 
These ideas suggest that it should be possible to achieve BA with direct methods while exploiting the unique characteristics of events, 
namely that they are continuously (asynchronously) triggered by edges as the camera moves, 
and that each event is a relative brightness measurement (i.e., a difference of two absolute measurements) (\cref{fig:eyecatcher}).

As done in other event-based tasks \cite{Rebecq17ral}, we approach this one with increasing level of complexity, 
in a three degrees-of-freedom (3-DOF) VO/SLAM scenario \cite{Weikersdorfer13icvs,Kim14bmvc,Reinbacher17iccp,Kim21ral}, e.g., rotational motion.
The reasons to focus on this scenario are multiple:
($i$) The scenario is interesting in its own right, as it finds applications in panorama creation \cite{Kim14bmvc} (e.g., for smartphones), 
space situational awareness (star tracking \cite{Chin20wacv}, sky or Earth mapping \cite{Cohen18amos,McHarg22spie,Arja23cvprw}), and VO/SLAM in dominantly-rotational motion cases (e.g., rotating satellites \cite{Chin19cvprw}).
($ii$) It allows us to focus on exploring the possibilities of a new method and how good results it can deliver without having to worry about some of the scene parameters (e.g., depth). 
($iii$) It allows us to gain insights about the problem (event-based BA), %
developing a solid foundation and deriving takeaways that can guide us to take on more complex scenarios with confidence\footnote{
Historically, 3-DOF motion scenarios have been an inspiration to develop basic, extensible tools. %
For example, Kim et al. solved the event-based 3-DOF SLAM problem \cite{Kim14bmvc} before extending its method to 6-DOF \cite{Kim16eccv}.
Contrast maximization (CMax) was introduced for angular velocity estimation \cite{Gallego17ral} and has been straightforwardly extended to tackle many other problems, including optical flow (the most complex type of image plane motion) \cite{Shiba24pami,Paredes23iccv,Hamann24eccv}.}.
($iv$) It allows us to carry out a comprehensive evaluation of how well prior motion estimators behave when used to initialize the solution of a new problem (event-based BA), gaining further knowledge about the interaction between old and new system modules.

In this paper, we propose a method called Event-based Photometric Bundle Adjustment (\EPBA{}) to tackle the BA problem for event cameras,
starting from the simple but rich and practical scenario of a purely rotational motion \cite{Cook11ijcnn,Kim14bmvc,Reinbacher17iccp,Gallego17ral,Kim18phd,Chin19cvprw,Kim21ral,Guo24tro} (see \cref{fig:pipeline}).
Stemming from first principles, we leverage the event generation model (EGM), without short-time linearizations, to define the photometric error.
Then we formulate the event-based BA problem as a non-linear least squares (NLLS) optimization in the camera motion and a panoramic intensity map.
Due to the sparsity of event data, only a portion of map pixels are observed and need to be refined,
which naturally leads to a semi-dense photometric map.
Sparsity is further exploited to design a tractable \LM{} (LM) solver that deals with the large number of variables involved.

To the best of our knowledge, \EPBA{} is the first event-only photometric BA approach that works on the intensity map, directly.
In the experiments, we run \EPBA{} to refine the camera motions and maps obtained by several state-of-the-art event-based rotation estimation front-end methods \cite{Kim14bmvc,Reinbacher17iccp,Gallego17ral,Kim21ral}, on both synthetic and real-world datasets.
The results show that \EPBA{} is able to obtain the jointly optimal camera motion and scene map in terms of photometric error.
Remarkably, the refined maps reveal details that were concealed by the front-end methods,
which suggests that the usage of EGM is key to unlock the potential of event cameras to record the rich photometric content of a scene.

EPBA builds on a large body of research on BA regarding robust objective functions and solvers.
Nevertheless, there is novelty:
($i$) the BA problem and solution here proposed is, to the best of our knowledge, new in the context of event cameras (see \cref{sec:related}).
($ii$) the sparsity pattern is specific of the problem addressed (not shared by frame-based BA), due to the measurement model (event generation model). 
The semi-dense character of the recovered map is distinctively new, which is due to the direct parametrization of the problem in terms of absolute brightness. 
($iii$) EPBA is able to run on the newest high-resolution event cameras (VGA and 1 Mpixel resolution), and to produce delicate panoramas from scratch (without an initial map).
($iv$) EPBA can also be used as a map-only BA in combination with a recent method \cite{Guo24tro}, providing high-quality results.

Our contributions are summarized as follows:
\begin{enumerate}
    \item We propose the first event-only photometric rotational bundle adjustment approach that jointly refines the camera motion as well as a panoramic intensity map (\cref{sec:method}).
    \item We perform the first study that exploits the sparsity of event data to design an NLLS solver for event-based BA problems.
    The solver can recover semi-dense photometric maps from events (\cref{sec:method:rotmotion}).
    \item We conduct comprehensive experiments on synthetic and real-world datasets (\cref{sec:experim}).
    The results show that our method is capable of achieving the jointly optimal camera rotations and map in terms of photometric error.
    \item  We demonstrate \EPBA{}'s applicability to panoramic imaging using high-resolution event cameras (\cref{sec:experim:wild,sec:experim:super_resolution,sec:experim:map_only}), as well as in high-speed, HDR and low-light scenarios (\cref{sec:experim:fast_hdr}), without requiring map initialization.
    \item We release the source code\ifarxiv. \else~(upon acceptance).\fi 
\end{enumerate}

The rest of the paper is organized as follows:
\cref{sec:related} reviews prior work on the topic, 
\cref{sec:method} introduces the \EPBA{} method,
which is thoroughly tested in \cref{sec:experim}.
\Cref{sec:limitations} discusses the limitations and \cref{sec:conclusion} draws conclusions.
Additional results and mathematical derivations are given in the supplementary material.

\ifclearsectionlook\cleardoublepage\fi \section{Related Work}
\label{sec:related}

\subsection{Event-based Rotation Estimation}
\label{sec:related:rotation}
The capabilities of event cameras to estimate rotational motion in challenging scenarios (e.g., high speed, HDR), have been investigated in several works, the most relevant of which are summarized in \cref{tab:eslam:methods}.

Soon after the invention of the Dynamic Vision Sensor (DVS) \cite{Lichtsteiner08ssc}, Cook et al. \cite{Cook11ijcnn} proposed a method consisting of a network of Interacting Visual Maps (IVM) to recover several visual quantities of interest from the event data. 
The method assumed a purely rotating event camera and estimated its angular velocity, optical flow, brightness gradient map and brightness map on the image plane that fitted the input events. 
The network operated by message passing with local update rules between the visual maps.

\begin{table}[t]
\centering
\caption{\emph{Event-based rotational VO/SLAM methods}. 
The columns indicate: 
the type of method (\textbf{D}irect or \textbf{I}ndirect --feature-based),
whether the method has a global refinement step (i.e., back-end \cite{Cadena16tro}),
whether the method exploits the event generation model (\textbf{L}inearized --LEGM-- \cite{Gehrig19ijcv} or not, i.e., \textbf{N}on-linear),
and the type of map.
\label{tab:eslam:methods}
}
\begin{adjustbox}{max width=1.0\linewidth}
\setlength{\tabcolsep}{3pt}
\renewcommand{\arraystretch}{1.1}
\begin{tabular}{lccccll}
\toprule
\textbf{System} & \textbf{Year} & \textbf{D/I} & \textbf{Refine} & \textbf{EGM} & \textbf{Map type} & \textbf{Remarks}\\
\midrule
IVM \cite{Cook11ijcnn} & 2011 & D & \nomark & \yesmark (L) & Grayscale & Reconstructs an image\\ %
\psmt{} \cite{Kim14bmvc} & 2014 & D & \nomark & \yesmark (L) & Grayscale & Reconstructs a panorama\\ %
\rtpt{} \cite{Reinbacher17iccp} & 2017 & D & \nomark & \nomark & Edge map & Probabilistic map\\
\cmaxw{} \cite{Gallego17ral} & 2017 & D & \nomark & \nomark & Local IWE & Visual gyroscope\\
\esmt{} \cite[Ch.5]{Kim18phd} & 2018 & D & \nomark & \yesmark (L) & Grayscale & Reconstructs a panorama\\ %
Chin et al.~\cite{Chin19cvprw} & 2019 & I & \yesmark & \nomark & Sparse points & Converts to frames\\
\cmaxgae{} \cite{Kim21ral} & 2021 & D & \nomark & \nomark & 3D-point set & Local \& global alignment\\
CMax-SLAM \cite{Guo24tro} & 2024 & D & \yesmark & \nomark & Panoramic IWE & Refines camera motion\\ %
EMBA \cite{Guo24eccv} & 2024 & D & \yesmark & \yesmark (L) & Grayscale grad. & Refines motion \& $\nabla$map\\
\textbf{EPBA} (this work) & 2024 & D & \yesmark & \yesmark (N) & Grayscale & Refines motion \& map\\
\bottomrule
\end{tabular}
\end{adjustbox}
\end{table}

Later, Kim et al. \cite{Kim14bmvc} proposed a simultaneous mosaicing and tracking (SMT) method consisting of two Bayesian filters operating in parallel (\psmt{} -- particle filter SMT). 
It estimated the camera motion and a grayscale panoramic intensity map of the scene. 
Subsequently, the camera tracker was replaced by an extended Kalman filter (EKF) in \cite{Kim18phd}, yielding \esmt{}.

Also working in parallel, but using a non-linear least squares (NLLS) formulation, a panoramic tracking and probabilistic mapping was developed by Reinbacher et al. \cite{Reinbacher17iccp} (\rtpt{}).
The tracking part leveraged direct alignment techniques \cite{Engel17pami}. %
The panoramic map of the scene stored a probability akin to the spatial event rate at each point: 
the higher the value, the more likely events are to be produced when camera pixels cross that map point. 
It is based on \cite{Weikersdorfer13icvs}. %

Contrast Maximization (CMax) was invented in \cite{Gallego17ral} to estimate the camera's angular velocity (i.e., \cmaxw{}). 
It warps events on the image plane and aligns them via a focus function \cite{Gallego18cvpr,Gallego19cvpr,Stoffregen19cvpr} that measures the goodness of fit between the events and the candidate rotational motion trajectories. 
The resulting motion-compensated (i.e., sharp) image of warped events (IWE) acts as a local edge map of the scene.
The work inspired Kim et al. \cite{Kim21ral} to jointly estimate angular velocity and absolute orientation (\cmaxgae{}), by using local and global event alignment stages.

\subsection{Bundle Adjustment with Event Cameras}
All of the methods described in \cref{sec:related:rotation} are short-term, i.e., front-ends of SLAM systems.
They lack a bundle adjustment (BA) refinement module, i.e., a SLAM back-end \cite{Cadena16tro} that reduces the propagation of errors between tracking and mapping parts of the system, which is desirable to improve accuracy and consistency.

Surveying the literature (\cref{tab:eslam:methods}), \cite{Chin19cvprw} introduced a BA approach for an event-based rotational motion system; 
but it was feature-based %
(after converting events into frames \cite{Chin19cvprw} or fitting line segments and extracting their end points \cite{Chin20wacv}) and was tested only on synthetic star-tracking data. 
Recently, a rotational SLAM pipeline consisting of both front-end and back-end, called \cmaxslam{}, has been developed \cite{Guo24tro}. 
The front-end is similar to \cite{Gallego17ral}, while the back-end maximizes the contrast of a panoramic IWE.
Expanding the survey to rigid-body motions, some SLAM systems have a back-end module, 
but they are either feature-based \cite{Rosinol18ral,Wang23ral} 
or based on grayscale images \cite{Hidalgo22cvpr}, thus borrowing the back-end from frame-based approaches \cite{Engel17pami,Alismail16accv}.

Therefore, to the best of our knowledge, event-only photometric (i.e., direct) BA is a novel, still unexplored topic, 
which we address with increasing complexity by starting from the rotational motion case to establish a foundation\footnote{
Concurrent to this journal work, 
the method \EMBA{} appeared at a conference \cite{Guo24eccv}.
It extends the filter-based method SMT to handle data from longer time intervals via batch optimization, 
and it adopts the LEGM as measurement model to design the loss function.
Optimization is carried out over the camera rotations and the \emph{spatial gradient} of the panoramic map.
However, it suffers from inaccuracies caused by LEGM's linearization errors, which produces slightly blurred maps.
Our experiments (\cref{sec:experim}) include a quantitative comparison with \cite{Guo24eccv}.}.

\emph{Map type:} 
Comparing the methods in \cref{tab:eslam:methods}, they differ in the scene representation, i.e., the type of scene map built. 
Some methods, like \psmt{} and \esmt{} estimate the richest form of scene map (absolute brightness panorama), based on the LEGM and assuming that the contrast sensitivity/threshold $C$ of the event camera is known.
Others maintain some form of edge map of the scene (e.g., spatial event rates or IWEs, instead of grayscale intensities), which does not require event polarity data and has been proven to be enough for camera localization \cite{Reinbacher17iccp}. 
Our goal when tackling the photometric bundle adjustment problem is to recover the camera trajectory orientation and the scene brightness, i.e., the richest form of scene map (\cref{fig:pipeline}). 
Hence, our approach maintains a similar type of map as \psmt{}, but with a key difference: we directly estimate the brightness map, instead of its spatial gradient. 
This implies that fewer variables are needed to represent the map (one value per pixel instead of two), 
and that only map pixels with events enter into the optimization, further reducing the number of variables to a semi-dense brightness map.

\emph{Loss type:} Related to the previous point, as shown in \cref{tab:eslam:methods}, the approaches that have a grayscale map representation use it in combination with the event generation model (EGM).
The natural loss function to be optimized is the photometric error conveyed by each event, where the EGM is used as measurement model \cite{Hidalgo22cvpr}. %
However, while all previous methods use the short-time linearized EGM \cite{Gehrig19ijcv}, we use the original (non-linear), more accurate version, which is one of the main reasons behind the unprecedented high-quality results.

\emph{Comparison with \cmaxslam{}:}
Our proposal comprises several similarities and differences with respect to the event-only back-end module of \cmaxslam{} \cite{Guo24tro}. 
Both back-ends apply direct methods, maintain global (panoramic) maps that help reduce drift, 
and have a continuous-time trajectory representation that allows them to define a smooth warping model between the image plane and the panoramic map(s).
However, the back-ends have key differences in regards to the type of loss function, map, and search space:
\cmaxslam{} optimizes a contrast/focus loss via a panoramic IWE, whereas our BA minimizes photometric error via a grayscale map. 
Formulating the problem as an NLLS allows us to leverage Gauss-Newton--type methods to converge quickly to the solution, while CMax does not have a similar NLLS formulation.
\cmaxslam{} searches for the best camera rotations, which implicitly define a panoramic IWE; hence the best (edge-) map is obtained as a by-product, with little control over it. 
Instead, photometric BA searches explicitly for both camera rotations and scene map. 
The search space is considerably larger, but it allows for further control over both variables.

Finally, note that \cmaxslam{} and our method are not mutually exclusive: 
one could run \cmaxslam{} to obtain an accurate camera motion that is used to initialize the full- or map-only photometric BA method introduced in this paper in order to get an HDR grayscale mosaic of the scene.

\ifclearsectionlook\cleardoublepage\fi \section{Event-based Photometric BA}
\label{sec:method}

\begin{figure}[t]
     \centering
     \begin{subfigure}{0.48\linewidth}
         \centering
         \includegraphics[width=\linewidth]{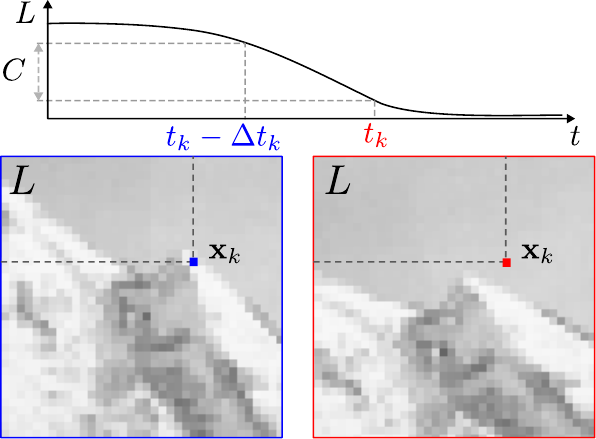}
         \caption{EGM at the sensor~\eqref{eq:EGM}.}
         \label{fig:EGM:image}
     \end{subfigure}\;
     \begin{subfigure}{0.48\linewidth}
         \centering
         \includegraphics[trim={0 30px 20px 0},clip,width=\linewidth]{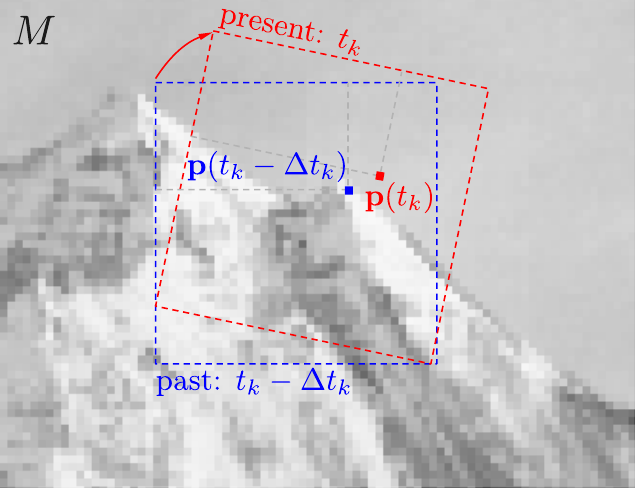}
         \caption{EGM on the scene map \eqref{eq:EGMmap}.}
         \label{fig:EGM:pano}
     \end{subfigure}
        \caption{
        An event represents a (temporal) brightness change at image pixel $\imgpoint_k$,
        or a (spatial) brightness change between two map points.
        \label{fig:EGM}
        }
\end{figure}

\def\figWidth{0.145\linewidth}
\begin{figure*}[t]
	\centering
    {\footnotesize
    \setlength{\tabcolsep}{1pt}
	\begin{tabular}{
    >{\centering\arraybackslash}m{0.3cm}
	>{\centering\arraybackslash}m{0.38\linewidth} 
	  >{\centering\arraybackslash}m{\figWidth}
    >{\centering\arraybackslash}m{\figWidth}
    >{\centering\arraybackslash}m{\figWidth}
    >{\centering\arraybackslash}m{\figWidth}
    }
    
    \rotatebox{90}{\makecell{Initial map}}
	& \includegraphics[width=\linewidth]{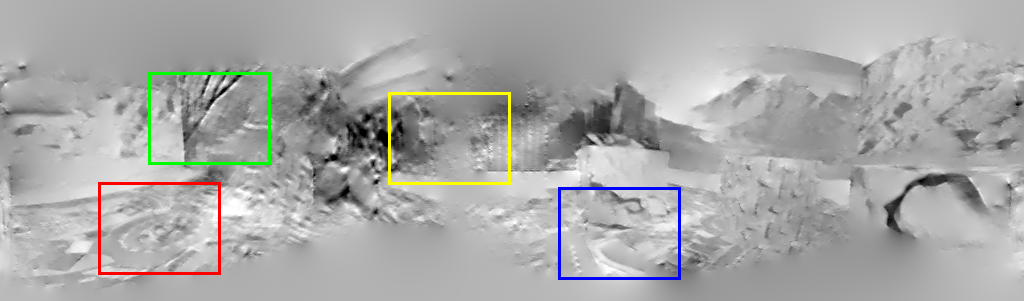}
    & \gframeRed{\includegraphics[width=\linewidth]{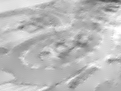}}
    & \gframeGreen{\includegraphics[width=\linewidth]{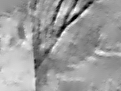}}
    & \gframeYellow{\includegraphics[width=\linewidth]{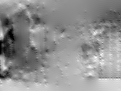}}
    & \gframeBlue{\includegraphics[width=\linewidth]{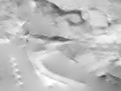}}
    \\[-0.5ex]

    \rotatebox{90}{\makecell{Semi-dense map}}
    & \gframe{\includegraphics[width=\linewidth]{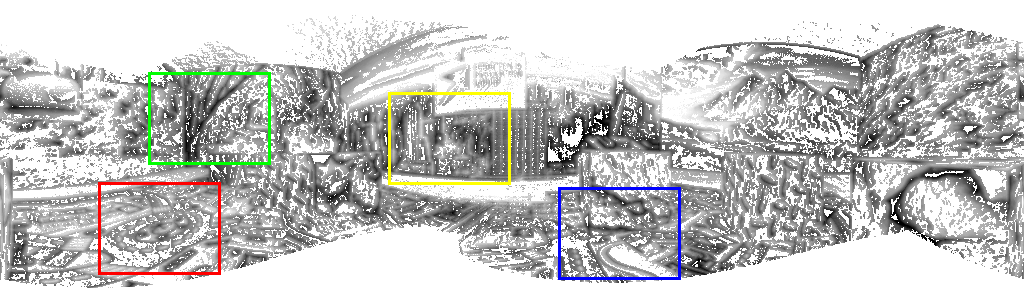}}
    & \gframeRed{\includegraphics[width=\linewidth]{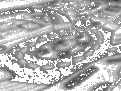}}
    & \gframeGreen{\includegraphics[width=\linewidth]{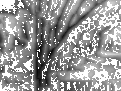}}
    & \gframeYellow{\includegraphics[width=\linewidth]{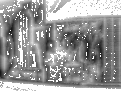}}
    & \gframeBlue{\includegraphics[width=\linewidth]{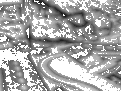}}
    \\[-0.5ex]

    \rotatebox{90}{\makecell{Final map}}
    & \includegraphics[width=\linewidth]{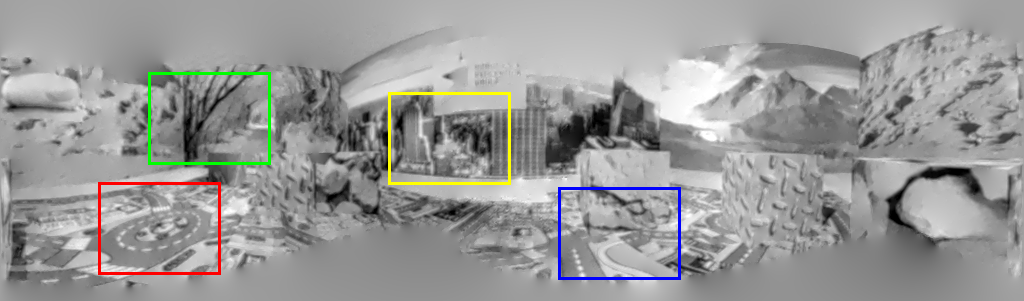}
    & \gframeRed{\includegraphics[width=\linewidth]{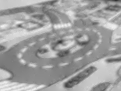}}
    & \gframeGreen{\includegraphics[width=\linewidth]{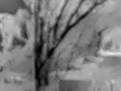}}
    & \gframeYellow{\includegraphics[width=\linewidth]{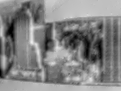}}
    & \gframeBlue{\includegraphics[width=\linewidth]{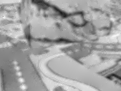}}
    \\
        
	\end{tabular}
	}
    \vspace{-1ex}
    \caption{Initial map (top), semi-dense map (middle) and densified map (bottom) for the \playroom{} sequence, 
    along with four insets (right columns).
    \label{fig:map_comp}
    }
\end{figure*}

This section first reviews the measurement model of event cameras (\cref{sec:method:EGM}).
Then it formulates the problem of event-based photometric BA in a general case (\cref{sec:method:formulation}).
The problem is then detailed for rotational camera motions (\cref{sec:method:error_term}), 
describing the error terms, their linearization and the partitioning and sparsity techniques leveraged to implement a tractable \LM{} solver.

\subsection{Event Generation Model (EGM)}
Let $L$ represent the logarithmic intensity of light at the sensor \cite{Gallego20pami}, 
and $M$ be the logarithmic intensity (radiance) on the surface $\bS \subset \Real^3$ of objects in the scene.
We formulate the well-known idealized version of the event-triggering condition and then rewrite it in terms of the scene elements.

\label{sec:method:EGM}
\textbf{EGM on the image plane}.~Each pixel of an event camera independently measures brightness changes, 
producing an event $e_k \doteq (\imgpoint_k,t_k,\pol_k)$ as soon as the logarithmic intensity change $\Delta L$ reaches a predefined contrast threshold $C$ \cite{Gallego20pami}.
The EGM is given by:
\begin{equation}
    \Delta L \doteq L(\imgpoint_k,t_k) - L(\imgpoint_k,t_k - \Delta t_k) = \pol_k C,
    \label{eq:EGM}
\end{equation}
where the event polarity $\pol_k \in \left\{ +1,-1 \right\}$ indicates the sign of the change \cite{Lichtsteiner08ssc}, 
and $\Delta t_k$ is the time elapsed since the last event at the same pixel $\imgpoint_k$.
This is illustrated in \cref{fig:EGM:image}: induced by camera motion, the brightness at pixel $\imgpoint_k$ decreases from white (snow) at $t_k-\Delta t_k$ to gray (sky) at time $t_k$, thus producing a negative event $e_k$.

\textbf{EGM in the scene}. Assuming ($i$) that the event is due to the relative motion between the camera and the scene and ($ii$) that there is a geometric and photometric model of the scene, we may cast \eqref{eq:EGM} from the image plane into an equation in terms of the scene elements.
Let $\bS(\imgpoint,t)\in \Real^3$ be the surface point that projects onto pixel $\imgpoint$ at camera viewpoint $\mP(t)$ 
(classically, $\bS(\imgpoint,t)$ is the point in space where the ray back-projected through pixel $\imgpoint$ from camera viewpoint $\mP(t)$ intersects the surface of objects in the scene).
As photometric model, assume that light travels without attenuation between a scene point $\bS$ and its projection $\imgpoint$ on the image plane \cite{Yezzi03ijcv2}, that is, the intensity of the scene map is $M(\bS(\imgpoint,t)) \doteq L(\imgpoint,t)$.
Then, $\bS(\imgpoint_k,t_k)$ and $\bS(\imgpoint_k,t_k - \Delta t_k)$ are the scene points that project on the same camera pixel, $\imgpoint_k$, at times $t_k - \Delta t_k$ and $t_k$, respectively,
and the intensities in \eqref{eq:EGM} transfer to these 3D points, 
yielding an equation in terms of map intensities:
\begin{equation}
    \Delta L \equiv M\bigl(\bS(\imgpoint_k,t_k)\bigr) - M\bigl(\bS(\imgpoint_k,t_k - \Delta t_k)\bigr) = \pol_k C.
    \label{eq:EGMmap}
\end{equation}

Hence, \eqref{eq:EGMmap} naturally associates each event with two map points (whose intensity difference is $\pol_k C$).
This is illustrated in \cref{fig:EGM:pano}:
as the camera moves, the pixel $\imgpoint_k$ scans the scene,
with the radiance decreasing from the white snow value $M(\bS(\imgpoint_k,t_k - \Delta t_k))$ to the gray sky  $M(\imgpoint_k,\bS(t_k))$ 
(assume there is a sphere of infinite radius that provides values for non-surface objects like the sky \cite{Yezzi03ijcv2,Jin07jmiv}).
The path a pixel takes across the scene is effectively described by the concatenation of segments. 
These segments are defined between pairs of \emph{consecutive} points: the start and end points of each segment correspond to the time of the previous event $t_k - \Delta t_k$ and the time of the current event $t_k$ (at the same pixel $\imgpoint_k$), respectively.

It is also illustrated in \cref{fig:eyecatcher}: 
as the camera rotates, the ray through pixel $\imgpoint_k$ (of event $e_k$) touches two map points (with panoramic coordinates $\bp(t_k-\Delta t_k)$ and $\bp(t_k)$), whose intensity difference (akin \eqref{eq:EGMmap}) describes $e_k$.

\subsection{General Formulation}
\label{sec:method:formulation}

\paragraph{Objective / Loss function}
\label{sec:method:formulation:objective}
Stemming from \eqref{eq:EGMmap}, each event represents a brightness change of predefined size $C$, 
and the brightness change may be written in terms of the camera motion and the scene map.
Therefore, assuming $C$ is known, a natural idea is to formulate the BA problem as finding the motion and scene parameters $\bP$ that minimize the least-squares error terms implied by \eqref{eq:EGMmap}:
\begin{equation}
\loss(\bP) \doteq \sum_{k=1}^{\numEvents}( z_{k}(\bP) - \pol_k C )^{2},
\label{eq:ObjFunc}
\end{equation}
where $z_{k} \equiv \hat{\Delta L}$ acts as a prediction for $\Delta L = \pol_k C$,
and $\numEvents$ is the number of events involved.
The scene parameters comprise the variables that describe the 3D map: shape (surfaces $\bS$) and appearance (intensity $M$).

Equation~\eqref{eq:ObjFunc} is a non-linear least squares (NLLS) function of the state $\bP$. 
It can be interpreted as the loss function corresponding to a maximum likelihood formulation of the problem when noise in $\Delta L$ is zero-mean Gaussian.
This is a reasonable design choice that is supported by empirical evidence \cite[Fig.6]{Lichtsteiner08ssc}.

Stacking the per-event error terms $\epsilon_k \doteq z_{k}(\bP) - \pol_k C$ into a vector $\be \in \Real^{\numEvents}$, the problem can be rewritten as: 
\begin{equation}
\label{eq:NLLSGenericProblem}
\min_{\bP}\loss(\bP),\quad \text{with}\quad \loss = \|\be\|^2 = \be^{\top}\be,
\end{equation}
where $\be(\bP)$ is the photometric error (or ``residual'') vector.

\paragraph{Solution approach}
\label{sec:method:formulation:solution}
The effective approach to minimize NLLS objectives is Gauss-Newton's (GN) method and its variants, such as Levenberg-Marquardt (LM) \cite{Hartley03book,Barfoot15book}. 
It consists in linearizing the error vector, solving the normal equations to compute a parameter update $\Delta\bP^{\ast}$, and iterate until (local) convergence.

Specifically for GN, assuming current values for the camera motion and scene parameters $\bP_{\text{op}}$, 
i.e., an ``operating point'' (abbreviated ``op'') in a high-dimensional space, and a perturbation $\Delta\bP$ around the operating point, the errors are linearized:
\begin{equation}
\be \approx \be_{\text{op}} + \mJ_{\text{op}}\Delta\bP,
\label{eq:ErrorsLinearized}
\end{equation}
where $\be_{\text{op}} \doteq \be(\bP_{\text{op}})$, and $\mJ_{\text{op}}$ is the derivative of the error with respect to the parameters $\bP$.
The optimal perturbation $\Delta\bP^{\ast}$ satisfies the system of normal equations, i.e.,
\begin{equation}
\mJ_{\text{op}}^{\top}\mJ_{\text{op}} \Delta\bP^{\ast} = -\mJ_{\text{op}}^{\top}\be_{\text{op}}
\quad\Leftrightarrow\quad
\mA\, \Delta\bP^{\ast} = \bb,
\label{eq:NormalEqs}
\end{equation}
which are used to update the ``operating point'' and iterate.

While this approach may appear as a classic one, there are several challenges involved in formulating the problem:
($i$) designing a meaningful and well-behaved loss, 
($ii$) identifying suitable parametrization and perturbation schemes,
($iii$) designing efficient approximations and solvers for a tractable implementation (e.g., solving very large systems of equations).
We address these challenges in the next section.

\subsection{Formulation for a Rotational Event Camera}
\label{sec:method:error_term}
\label{sec:method:rotmotion}

The general approach in \cref{sec:method:formulation} can be instantiated on different problems, such as front-to-parallel motion \cite{Weikersdorfer13icvs}, rotational motion \cite{Kim14bmvc}, etc.
For a purely rotating event camera, 
we let $M$ be a 2D brightness map of the scene (in logarithmic scale, which we omit for brevity),
and the map points can be described using 2D coordinates (e.g., on a panorama \cite{Kim14bmvc},
\cref{fig:EGM:pano,fig:map_comp}).
Writing out explicitly the dependency of $\epsilon_{k}$ with respect to the unknowns,
\begin{equation}
\label{eq:ObsOrigEGM}
\epsilon_{k} \doteq M\bigl(\bp(t_{k})\bigr)-M\bigl(\bp(t_{k}-\Delta t_{k})\bigr) - \pol_k C.
\end{equation}
There is a chain of transformations from the event location to the error entry (for simplicity we drop the subscript $k$):
\begin{equation}
\label{eq:chainoftransformations}
\bx\stackrel{\Kint^{-1}}{\mapsto}\bX\stackrel{\Rot(t)}{\mapsto}\bX'(t)\stackrel{\pi}{\mapsto}\bp(t)\mapsto M(\bp(t))\mapsto\epsilon
\end{equation}
The event pixel location $\bx$ is back-projected to $\bX \in \Real^3$, rotated $\bX'=\Rot(t)\bX \in \Real^3$, and projected onto a (panoramic) map point $\bp$.
Though it is possible to define a map on the sphere, $\bX'(t)\equiv \bS(\imgpoint,t)$, for simplicity we consider a 2D rectangular representation (e.g., equirectangular projection $\pi$).
Compounding transformations, $\bx$ is transferred (i.e., warped) to the map point $\bp$ according to the camera orientation $\Rot(t)$, the intrinsic calibration $\Kint$ and the type of projection model $\pi$ used to represent the map: $\bx\mapsto\bp$, i.e., 
\begin{equation}
\label{eq:mappointwarp}
    \bp(t) \doteq \Warp(\bx;\Rot(t),\Kint,\pi).
\end{equation}
Then, the intensities at two map points are read, as depicted in \cref{fig:EGM:pano} (also in \cref{fig:eyecatcher}), 
and used to compute \eqref{eq:ObsOrigEGM}, which leads to the photometric error \eqref{eq:NLLSGenericProblem}.

\subsubsection{Parameterization, Operating Point and Perturbations}
\label{sec:method:perturbation}
While in general the camera orientation trajectory $\Rot(t)$ and the map $M$ are functions defined in infinite-dimensional spaces,
we approximate them to be implemented on a computer.
Let $\balpha$ and $\bbeta$ be the finite number of parameters that are used to approximate the trajectory $\Rot(t;\balpha)$ and the map $M(\bbeta)$. 
The trajectory is approximated using a spline representation that interpolates $\Rot(t)$ linearly using two neighboring control poses $\{\Rot_{i},\Rot_{i+1}\}\subset\balpha$.
Thus $\balpha$ represents the control poses that specify the trajectory \cite{Guo24tro}.
Likewise the 2D function $M:\Real^{2}\to\Real$ is approximated by a panoramic intensity image with pixels $\bbeta$ (\cref{fig:map_comp}).
This is in stark contrast with prior works (\cref{sec:related}), which use the spatial gradient $\nabla M$ to parameterize the map \cite{Kim14bmvc,Kim18phd,Guo24eccv}. 

To linearize the errors \eqref{eq:ErrorsLinearized} we consider pose perturbations in the Lie-group sense (control poses in the Lie group and perturbations in the Lie algebra \cite{Barfoot15book}), and pixel perturbations directly in brightness space.
Specifically, the perturbations of the camera pose at time $t$ (not necessarily a control pose) and the brightness map (continuous variables) are:
\begin{equation}
\label{eq:poseandmapperturbation}
\Rot(t) =\exp(\delta\boldsymbol{\varphi}^{\wedge})\,\Rot_{\text{op}}(t), 
\qquad
M = M_{\text{op}}+\Delta M,
\end{equation}
where we use the exponential map (notation from \cite{Barfoot15book}).

The ``operating point'' in the search space consists of the current camera orientation trajectory (parameterized by $\numPoses$ control poses) and the map (e.g., brightness values):
\begin{equation}
\bP_{\text{op}} = \{\Rot_{1}^{\text{op}},\ldots,\Rot_{\numPoses}^{\text{op}},\bbeta_{1}^{\text{op}},\ldots,\bbeta^{\text{op}}_{\numPixels} \}.
\label{eq:OperPointCamAndMap}
\end{equation}

The $\numPoses$ camera control poses and the $N_{p}$ map pixels 
are perturbed according to
\begin{equation}
\label{eq:Perturb}
\Rot_{i} = \exp(\rotperturb_{i}^{\wedge})\,\Rot_{i}^{\text{op}},\qquad 
\bbeta_n = \bbeta_n^{\text{op}} + \Delta\bbeta_n.
\end{equation}
Collecting terms, the perturbation vector is partitioned as
\begin{equation}
\label{eq:PerturbVecPartitioning}
\Delta\bP = (\Delta\bP_{\balpha}^{\top},\Delta\bP_{\bbeta}^{\top})^{\top},
\end{equation}
with $\Delta\bP_{\balpha} \!=\! (\rotperturb_{1}^{\top},..,\rotperturb_{\numPoses}^{\top})^{\top}$, 
$\Delta\bP_{\bbeta} \!=\! (\Delta\bbeta_{1},..,\Delta\bbeta_{N_{p}})^{\top}$.

\begin{figure}[t]
     \centering
     \begin{subfigure}[t]{0.69\linewidth}
         \centering
         \includegraphics[width=\linewidth]{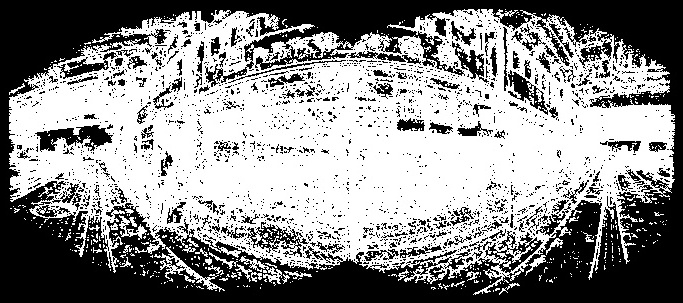}
         \caption{Semi-dense valid pixels.}
         \label{fig:sparsity:mask}
     \end{subfigure}
     \begin{subfigure}[t]{0.30\linewidth}
         \centering
         \includegraphics[width=\linewidth]{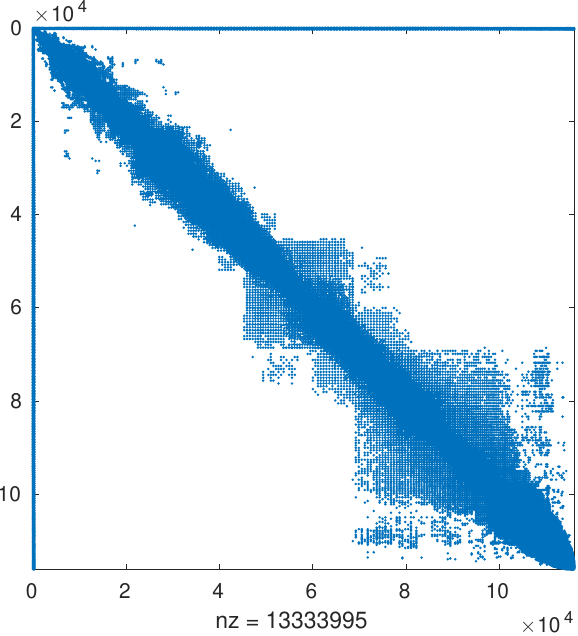}
         \caption{Sparsity.}
         \label{fig:sparsity:pattern}
     \end{subfigure}
        \caption{Set of semi-dense valid pixels (left) and corresponding ``arrowhead'' sparsity pattern of the normal equation's matrix (\bicycle{} sequence, \cmaxw{} trajectory).
        \label{fig:sparsity}}
\end{figure}

\subsubsection{Linearization of the Photometric Error}
\label{sec:suppl:linearization}

Perturbing the camera motion and the scene map we aim to arrive at an expression like \eqref{eq:ErrorsLinearized}:
\begin{equation}
\be\approx\be_{\text{op}}+\mJ_{\text{op},\balpha}\Delta\bP_{\balpha}+\mJ_{\text{op},\bbeta}\Delta\bP_{\bbeta},
\label{eq:LinearizedErrorPartitionedHigh}
\end{equation}
where $\mJ_{\text{op},\balpha}\doteq\left.\prtl{\be}{\bP_{\balpha}}\right|_{\text{op}}$
and $\mJ_{\text{op},\bbeta}\doteq\left.\prtl{\be}{\bP_{\bbeta}}\right|_{\text{op}}$.
Thus, we only consider the first-order terms (i.e., discard higher order ones). 
Here, $\mJ_{\text{op},\balpha}$ is an $N_{e}\times 3\numPoses$ matrix, and $\mJ_{\text{op},\bbeta}$ is an $N_{e}\times \numPixels$ matrix,
where $N_{e}$ is the number of events and $\numPixels$ is the number of valid panorama pixels (\cref{fig:sparsity}).

Let us write the linearization of each error term in \eqref{eq:LinearizedErrorPartitionedHigh}. 
Stemming from \eqref{eq:ObsOrigEGM}, for each error entry we obtain: 
\begin{align}
\epsilon_{k} & \approx \underbrace{ M_{\text{op}}(\bp_{\text{op}}(t_{k}))
-M_{\text{op}}(\bp_{\text{op}}(t_{k}-\Delta t_{k}))-\pol_{k}C }_{\epsilon_{\text{op},k}}\nonumber \\
 & \quad\underbrace{-\bq_{\text{op}}^{\top}(t_{k})\delta\boldsymbol{\varphi}+\bq_{\text{op}}^{\top}(t_{k}-\Delta t_{k})\delta\tilde{\boldsymbol{\varphi}}}_{\text{linear in }\Delta\bP_{\balpha}}\nonumber \\
 & \quad\underbrace{+\Delta M(\bp_{\text{op}}(t_{k}))-\Delta M(\bp_{\text{op}}(t_{k}-\Delta t_{k}))}_{\text{linear in }\Delta\bP_{\bbeta}},
 \label{eq:LinErrorOrigEGM}
\end{align}
where (see Appendix \ref{sec:suppl:linear})
\begin{align}
    \bq_{\text{op}}^{\top}(t) & \doteq \left(\nabla M_{\text{op}} (\bp_{\text{op}}(t))\right)^{\top}\mE_{\text{op}}(t) \\
    \mE_{\text{op}}(t) & \doteq \left.\prtl{\pi}{\bz}\right|_{\bz_{\text{op}}}\bz_{\text{op}}^{\wedge} \\
    \pi & \;\;\text{is the equirectangular projection } \mathbb{R}^3\to\mathbb{R}^2 %
\end{align}
\begin{align}
    \bz(t) & = \Rot(t)\Kint^{-1}\bx^{h}\\
    \bz_{\text{op}}(t) & \doteq \Rot^{\text{op}}(t)\Kint^{-1}\bx^{h}\\
    \bx^{h} &= (x,y,1)^{\top} \text{are homogeneous coordinates of } \bx\\
    {}^{\wedge} & \;\;\text{is the hat (skew-symmetric) operator \cite{Barfoot15book}} \\
    \delta\boldsymbol{\varphi} & \;\;\text{is the perturbation of } \Rot(t_k)\\
    \delta\tilde{\boldsymbol{\varphi}} & \;\;\text{is the perturbation of } \Rot(t_k-\Delta t_{k}).
\end{align}

Equation \eqref{eq:LinErrorOrigEGM} states that the predicted contrast in \eqref{eq:ObsOrigEGM} depends on: 
the event camera orientations at two different times $\{ t_{k}$, $t_{k}-\Delta t_{k} \}$ and the map intensities at two different pixel locations $M\bigl(\bp(t_{k})\bigr)$ and $M\bigl(\bp(t_{k}-\Delta t_{k})\bigr)$ (\cref{fig:EGM:pano}).
Due to the linear spline rotation interpolation, $\delta\tilde{\boldsymbol{\varphi}}$ uses the two control rotations closest to time $t_{k}-\Delta t_{k}$.
These need not be the same ones as those of $\delta\boldsymbol{\varphi}$ (at time $t_{k}$). 
We leverage \cite{Barfoot15book,Sommer20cvpr,Gallego14jmiv} to efficiently calculate the analytical derivatives of the errors with respect to the control poses.

\subsubsection{Problem Size, Partitioning and Sparsity}

\emph{Problem size}:
It is clear that for problems of moderate size, with millions of events, thousands of pixels, and hundreds of camera control poses, Jacobian matrix storage is intractable.
For example, for an input sequence with $\numEvents = 10^6$ events,  
and a panoramic map of $1024 \times 512$~px, 
if we assume $25\%$ of map pixels are valid pixels ($\numPixels = 1.3 \times 10^4$)
and there are $\numPoses = 50$ control poses, the number of elements in $\mJ_{\text{op}}$ will be $1.315 \times 10^{10}$. 
Even if we maintain $\mJ_{\text{op}}$ using 32-bit floating point precision (4 bytes), the RAM memory required by $\mJ_{\text{op}}$ is approximately 48.9 GB, which is already beyond the capabilities of standard computers.
While one could save memory by storing matrix $\mJ_{\text{op}}$ in sparse format, 
it does not have a simple sparsity pattern, and accessing its non-zero entries is time-consuming.
Since the size of matrix $\mA$ in the normal equations \eqref{eq:NormalEqs} only depends on the number of unknowns, which is significantly smaller than the size of $\mJ_{\text{op}}$, the computation of $\mJ_{\text{op}}$ is circumvented by directly calculating $\mA$ in an efficient way.
This is in line with ideas from large-scale frame-based BA (e.g., \cite{Brown06ijcv,Ni07iccv,Wu11cvpr}).

\emph{Partitioning}: 
The state $\bP$ of the BA problem has two parts: the camera rotations and the scene map. 
This allows us to partition \eqref{eq:PerturbVecPartitioning} and the normal equations \eqref{eq:NormalEqs} in blocks:
\begin{equation}
\left(\begin{array}{cc}
\mA_{11} & \mA_{12}\\
\mA^\top_{12} & \mA_{22}
\end{array}\right)\left(\begin{array}{c}
\Delta\bP_{\balpha}^{\ast}\\
\Delta\bP_{\bbeta}^{\ast}
\end{array}\right)
=\left(\begin{array}{c}
\bb_{1}\\
\bb_{2}
\end{array}\right),
\label{eq:NormalEqsFirstPartitioning}
\end{equation}
where $\mA_{11} \doteq \mJ_{\text{op},\balpha}^{\top}\mJ_{\text{op},\balpha}$ only depends on the derivatives with respect to the camera rotations,
$\mA_{22} \doteq \mJ_{\text{op},\bbeta}^{\top}\mJ_{\text{op},\bbeta}$ only depends on the derivatives with respect to the scene map,
and $\mA_{12} \doteq \mJ_{\text{op},\balpha}^{\top}\mJ_{\text{op},\bbeta}$. 
There is a large difference in state dimensions: the size of $\mA_{11}$ (rotations) is significantly smaller than that of $\mA_{22}$ (map pixels).
This fact can be leveraged for efficient solution of block-partitioned systems using well-known tools.

\emph{Sparsity}:
Besides block-partitioning, we can exploit sparsity to implement a tractable \LM{} solver for this problem.
As shown in \cref{fig:sparsity}, due to the sparsity of event data, only a part of map points is observed (the so-called ``valid pixels''), which leads to a \emph{semi-dense} map updating scheme and a sparse $\mA_{22}$.
With the tailored LM solver in \cref{sec:suppl:form_normalEq}, \EPBA{} is able to work on sequences of 
$\approx 50$ M events, with a map size of $8192 \times 4096$~px and $\approx 75$ control poses (e.g., \crossroad{} in \cref{fig:wild_experim}).

\subsubsection{\LM{} Solver in Cumulative Form}
\label{sec:suppl:form_normalEq}
We build the normal equations \eqref{eq:NormalEqs}, \eqref{eq:NormalEqsFirstPartitioning}, without the need to store the full $\mJ_{\text{op}}$,
by directly computing the system matrix $\mA$ and vector $\bb$ in a cumulative way from the linearization of each error term \eqref{eq:LinErrorOrigEGM}.

\paragraph{Left hand side matrix $\mA \in \Real^{(3 \numPoses + \numPixels)^2}$}
Let $\br_k^{\top}$ be the $k$-th row of $\mJ_{\text{op}}$, which stores the derivatives of error term $\epsilon_k$.
Following the partitioning in \eqref{eq:NormalEqsFirstPartitioning}, we can further write $\br_k^{\top} = (\br_{k,\balpha}^{\top}, \br_{k,\bbeta}^{\top})$,
where $\br_{k,\balpha}$ and $\br_{k,\bbeta}$ are the camera pose part and map part of $\br_k$, respectively.
Then we rewrite the system matrix as the sum of the outer product of each row:
\begin{equation}
    \mA \doteq \mJ_{\text{op}}^{\top}\mJ_{\text{op}} 
    = \sum_{k=1}^{\numEvents} \br_k \br_k^{\top} 
    \!=\! \sum_{k=1}^{\numEvents} \mA_k 
    \!=\! \sum_{k=1}^{\numEvents}
    \begin{pmatrix}
    {\mA_{11}}_k & {\mA_{12}}_k \\
    {\mA_{12}^{\top}}_k & {\mA_{22}}_k
    \end{pmatrix}\!,
    \label{eq:LHSMatCumulative}
\end{equation}
where ${\mA_{11}}_k \doteq {\br_{k,\balpha}} {\br_{k,\balpha}}^{\top}$,
${\mA_{12}}_k \doteq {\br_{k,\balpha}} {\br_{k,\bbeta}}^{\top}$
and ${\mA_{22}}_k \doteq {\br_{k,\bbeta}} {\br_{k,\bbeta}}^{\top}$.
Hence, the contribution of each event to $\mA$ is additive, which offers a cumulative way to form the system matrix.
As mentioned in \cref{sec:method:error_term}, each error term depends on the intensities at two map points.
This leads to a sparse structure of ${\mA_{22}}_k$.
For simplicity, if each warping operation is from a sensor pixel onto a single panoramic map pixel (nearest neighbor), then every 
${\mA_{22}}_k$ only has four non-zero elements, with two $+1$ lying on the diagonal and two $-1$ symmetrically located on the upper/lower triangular parts. 
This determines the sparsity pattern of $\mA$, which can be exploited to speed up the solution of the normal equations.

\paragraph{Right hand side vector $\bb \in \Real^{3 \numPoses + \numPixels}$}
\label{sec:suppl:LM_details:RHS}
Similarly, let $\bc_n$ be the $n$-th column of $\mJ_{\text{op}}$.
Following the partitioning in \eqref{eq:NormalEqsFirstPartitioning}, we can rewrite
\begin{equation}
    \begingroup %
    \setlength\arraycolsep{2pt}
    \mJ_{\text{op}} = 
    \begin{pmatrix}
        \bc_{1,\balpha}, & \ldots, & \bc_{3 \numPoses,\balpha}, 
        & \bc_{1,\bbeta}, & \ldots, & \bc_{\numPixels,\bbeta}
    \end{pmatrix},
    \label{eq:JopColumns}
    \endgroup
\end{equation}
where $\bc_{i,\balpha} = \left.\frac{\partial \be}{\partial {\bP_{i,\balpha}}}\right|_{\text{op}}$ 
and $\bc_{j,\bbeta} = \left.\frac{\partial \be}{\partial {\bP_{j,\bbeta}}}\right|_{\text{op}}$ 
store the derivatives of the whole error vector $\be$ with respect to each component of the pose/map state.
Inserting \eqref{eq:JopColumns} into \eqref{eq:NormalEqs} we obtain the cumulative formula for each element of $\bb$:
${\bb_1}_i = -\bc^\top_{i,\balpha} \be_{\text{op}}$ and ${\bb_2}_j = -\bc^\top_{j,\bbeta} \be_{\text{op}}$, that is,
\begin{equation}
    {\bb_1}_i \!=\! -\sum_{k=1}^{\numEvents} \left.\frac{\partial \epsilon_k}{\partial {\bP_{i,\balpha}}}\right|_{\text{op}}\!\!\epsilon_{\text{op},k}, \;
    {\bb_2}_j \!=\! -\sum_{k=1}^{\numEvents} \left.\frac{\partial \epsilon_k}{\partial {\bP_{j,\bbeta}}}\right|_{\text{op}}\!\!\epsilon_{\text{op},k}.
\label{eq:RHSVecCumulative}
\end{equation}

Formulas \eqref{eq:LHSMatCumulative} and \eqref{eq:RHSVecCumulative} accumulate the contribution of each event to the normal equations \eqref{eq:NormalEqs}.
Thanks to the proposed cumulative method for computing the normal equations, there is no need to update the non-zero elements of the sparse $\mA_{22}$ by index, which would be very inefficient for the data structure of sparse matrices.
Instead, one can just maintain all the non-zero elements (values and indices) individually, and assemble them all at once into a sparse matrix, after all events are processed.

\paragraph{Solving and Updating} 
Adopting a \LM{} approach, the augmented normal equations are:
\begin{equation}
    \left(\mA + \lambda\, \text{diag}(\mA)\right) \Delta \bP^{\ast} = \bb,
    \label{eq:augmentedNormalEqs}
\end{equation}
where $\lambda$ is an exploration-exploitation parameter that varies between iterations:  
if the cost decreases (resp. increases) at the new operating point, $\lambda$ will be decreased (resp. increased) by a factor of 10 \cite{Hartley03book}.

The normal equations \eqref{eq:augmentedNormalEqs} can be solved using advanced techniques for linear systems of equations, 
which have been used in frame-based BA literature \cite{Hartley03book,Ni07iccv}. %
In particular, we consider the Cholesky decomposition described in \cite{Barfoot15book} (a direct method)
and the conjugate gradient (CG) method \cite{Hager06survey} (an iterative method).
The Cholesky decomposition can be accelerated by exploiting the sparsity pattern of $\mA_{22}$ by means of the approximate minimum degree permutation (AMD) matrix reordering algorithm \cite{Amestoy04tms}.
However, for very large problems the Cholesky decomposition becomes too expensive, 
and the iterative CG solver, which does not even require storing matrix $\mA$, becomes the method of choice.

After solving for the optimal perturbations in \eqref{eq:augmentedNormalEqs}, the operating camera rotations and scene map are updated \eqref{eq:Perturb}.
The iterative process (linearize--solve--update) is repeated until convergence.

\subsubsection{Robust Objective Function}
\label{sec:method:robust}

Building on well-known techniques from classical BA, we increase the robustness of the method against noise by considering Huber and Cauchy loss functions \cite{Hartley03book,Barfoot15book}:
\begin{equation}
\begin{split}
\label{eq:robustlossfunctions}
\text{Huber: } & \rho(\epsilon_k) = {\begin{cases}{\epsilon^{2}_k}&{\text{for }}|\epsilon_k| < \delta ,\\ 
\left(2|\epsilon_k|-\delta \right)\delta, &{\text{otherwise.}}\end{cases}} \\[0.4ex]
\text{Cauchy: } & \rho(\epsilon_k) = b^2 \log (1 + \epsilon^2_k / b^2).
\end{split}
\end{equation}
We thus replace $\sum_k \epsilon_k^2$ in \eqref{eq:NLLSGenericProblem} by $\sum_k \rho(\epsilon_k)$.
These functions limit the influence of event data with large errors (e.g., ``outliers''), by assigning loss values that grow slower than quadratic.
The normal equations and the LM approach have to be adapted, accordingly \cite{Barfoot15book,Brown06ijcv}. %
Both functions \eqref{eq:robustlossfunctions} have a scale hyperparameter ($\delta$ or $b^2$) that allows us to control the shape of the loss. 
The hyperparameters are determined via sensitivity analyses (\cref{sec:experim:sensitivity}); we set $\delta = 0.05$ and~$b^2 = 1/50$.

\subsubsection{Map Densification}
\label{sec:method:map_smoothing}

The LM solver only refines valid pixels (semi-dense), 
which may lead to intensity discontinuities (artifacts) in the map (see the middle row of \cref{fig:map_comp}).
To overcome this issue, we perform a Poisson in-painting and smoothing of the semi-dense map.
Specifically, we apply spatial convolution kernels 
$\nabla_x = (-0.5, 0, 0.5)$ and $\nabla_y = (-0.5, 0, 0.5)^{\top}$
to the refined semi-dense map $M$.
The output gradient maps $M_x$ and $M_y$ are initialized to zero, 
and they are updated only at pixels where the semi-dense map $M$ fully overlaps with the non-zero values of each kernel mask.
Then we reconstruct a densified map, $M_\text{final}$, by solving Poisson's equation \cite{Agrawal06eccv}:
\begin{equation}
    \nabla^2 M_\text{final} = \frac{\partial M_x}{\partial x} + \frac{\partial M_y}{\partial y}.
    \label{eq:poisson}
\end{equation}
An example of the final, refined map is shown in the bottom row of \cref{fig:map_comp}.

\ifclearsectionlook\cleardoublepage\fi \section{Experiments}
\label{sec:experim}

\begin{figure*}[t]
     \centering
     \begin{subfigure}[b]{0.40\linewidth}
         \centering
         \includegraphics[width=\linewidth]{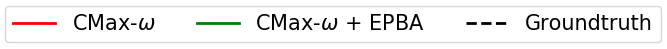}
     \end{subfigure}
     
    \begin{subfigure}[b]{0.245\linewidth}
         \centering
         \includegraphics[width=\linewidth]{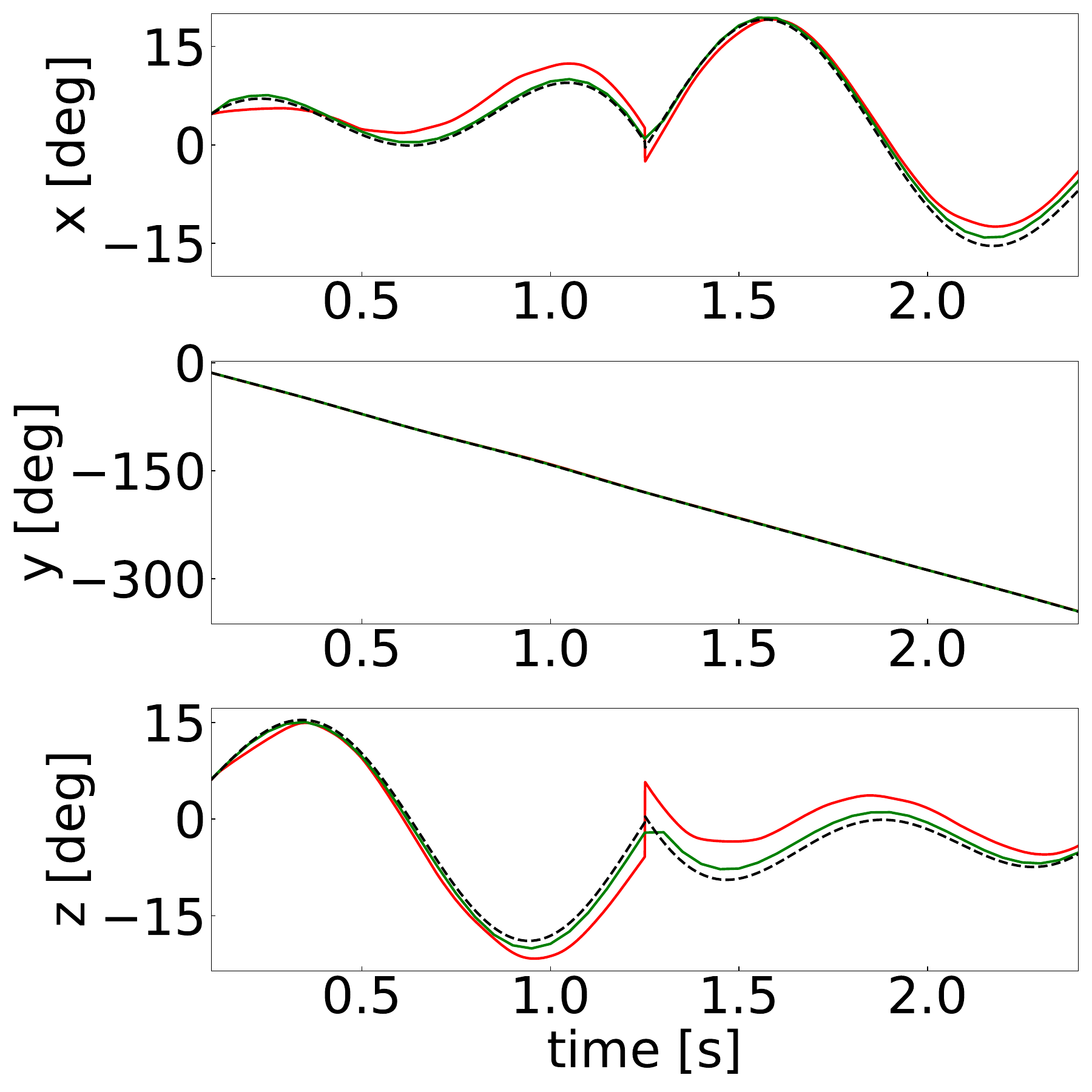}
         \caption{\playroom{} (synth)}
         \label{fig:traj:playroom}
     \end{subfigure}
     \begin{subfigure}[b]{0.245\linewidth}
         \centering
         \includegraphics[width=\linewidth]{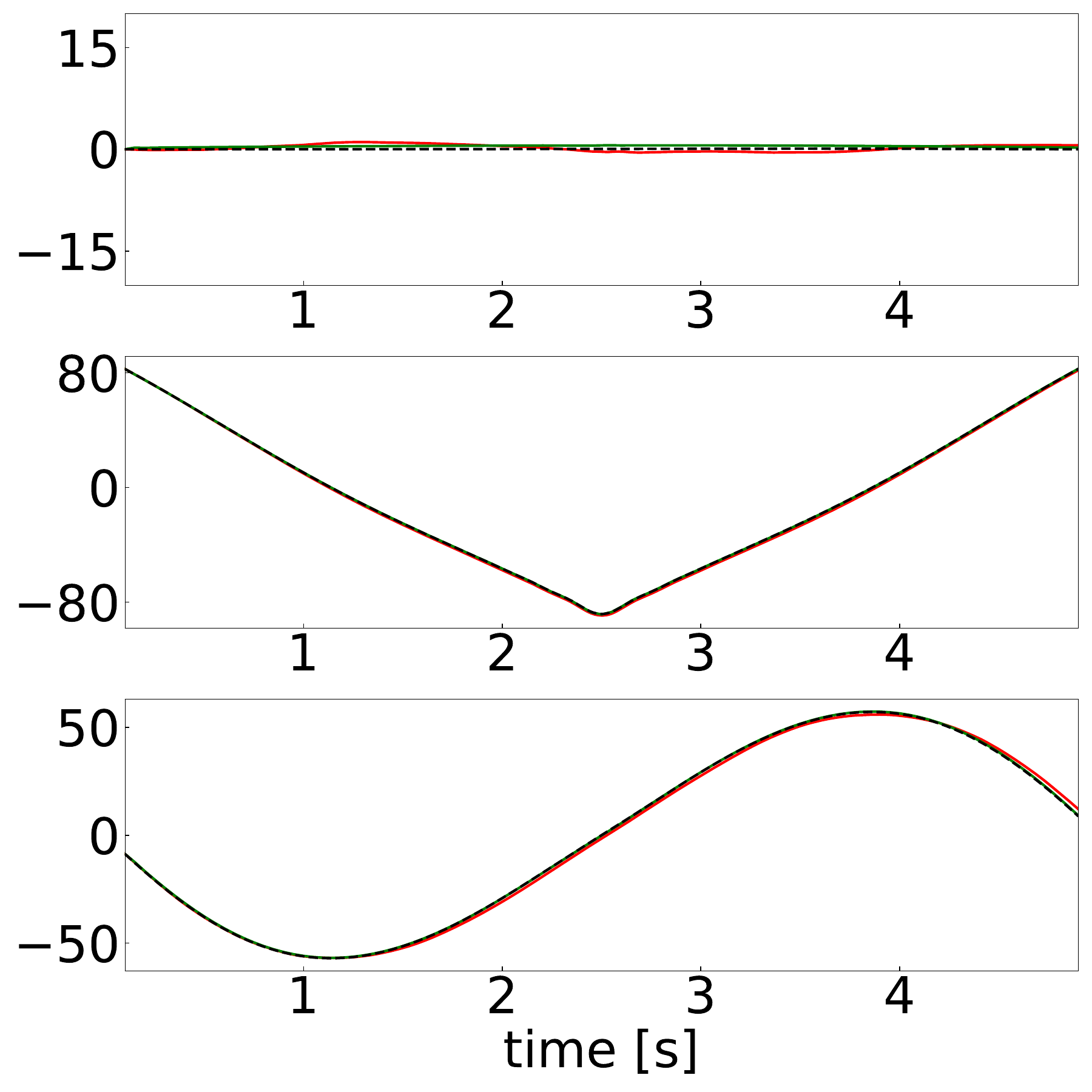}
         \caption{\bicycle{} (synth)}
         \label{fig:traj:bicycle}
     \end{subfigure}
     \begin{subfigure}[b]{0.245\linewidth}
         \centering
         \includegraphics[width=\linewidth]{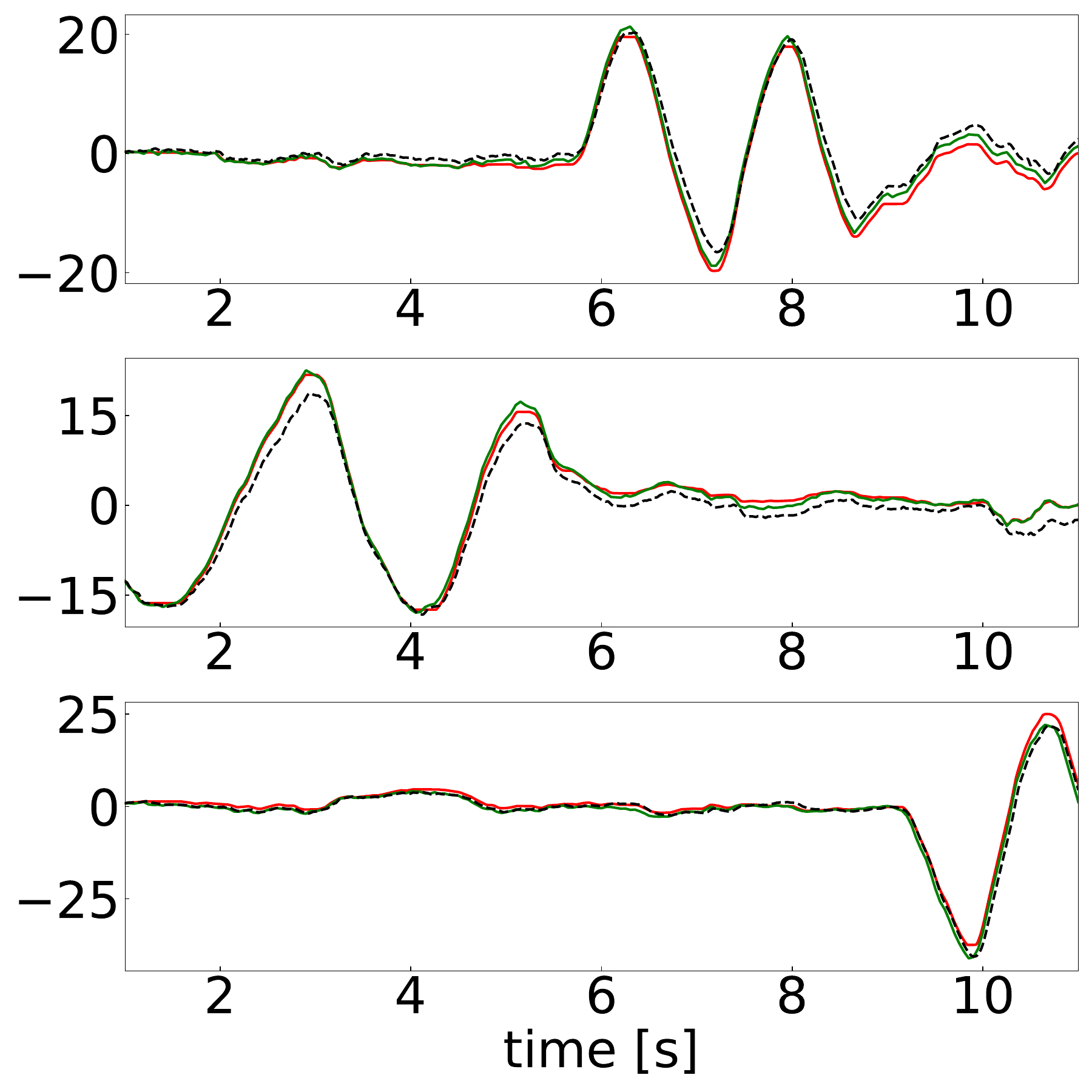}
         \caption{\boxes{} (real)}
         \label{fig:traj:boxes}
    \end{subfigure}
    \begin{subfigure}[b]{0.245\linewidth}
         \centering
         \includegraphics[width=\linewidth]{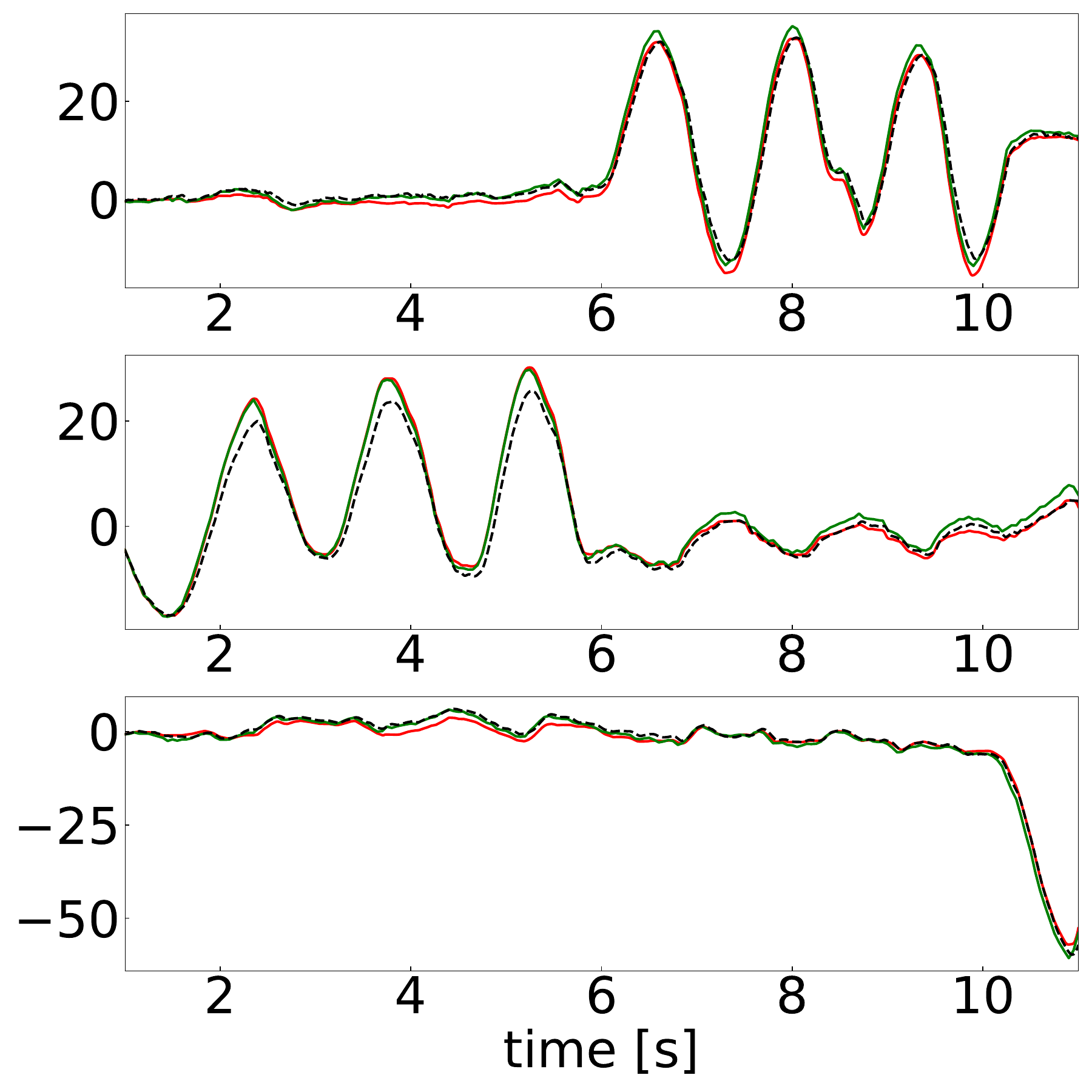}
         \caption{\dynamic{} (real)}
         \label{fig:traj:dynamic}
    \end{subfigure}
    \vspace{-1ex}
    \caption{Camera rotation degrees-of-freedom (DOFs) before refinement (``\cmaxw{}'') and after refinement (``\cmaxw{}+EPBA''). 
    The rotations are obtained by minimizing the quadratic loss function in \eqref{eq:NLLSGenericProblem}.
    }
    \label{fig:traj}
\end{figure*}

We thoroughly evaluate the proposed method.
First, we introduce the experimental setup (\cref{sec:experim:setup}) (datasets, initialization and evaluation metrics).
Second, we present the results on synthetic data (\cref{sec:experim:synth_data}) and real-world data (\cref{sec:experim:real_data}). 
Then we report the computational effort (\cref{sec:experim:runtime}) and characterize the sensitivity of EPBA (\cref{sec:experim:sensitivity}).
We also show experiments in the wild (without ground truth) (\cref{sec:experim:wild}),
in high-speed scenarios and challenging illumination conditions (\cref{sec:experim:fast_hdr}),
and discuss super-resolution (\cref{sec:experim:super_resolution}) and map-only refinement (\cref{sec:experim:map_only}).
Please see also the accompanying \textbf{video}.

\subsection{Experimental Setup}
\label{sec:experim:setup}

\subsubsection{Datasets}
\label{sec:experim:setup:datasets}

We test EPBA on six synthetic sequences released in \cite{Guo24tro} and on four real-world sequences from the standard dataset \cite{Mueggler17ijrr}.
All these sequences consist of events, frames (not used), IMU data and ground truth (GT) poses.
In addition, we utilize the Event-based Fast Rotation Dataset (EFRD)\footnote{\url{https://github.com/shicy17/VAVM}} to demonstrate the performance of \EPBA{} in high-speed and challenging illumination conditions (HDR and low light).

The synthetic sequences in the ECRot dataset \cite{Guo24tro} cover a variety of scenes (indoor, outdoor, daylight, night, human-made and natural) and their resolutions vary from 2K (\playroom{}), 4K (\bicycle{}), 6K (\city{} and \street{}), to 7K (\town{} and \bay{}).
These sequences were generated with a DAVIS240C camera model (240 $\times$ 180 px) and a duration of 5~s, with the only exception of \playroom{}, whose camera model is a DVS128 ($128 \times 128$ px) and duration is 2.5~s.

The Event Camera Dataset (ECD) \cite{Mueggler17ijrr} provides four hand-held rotational motion sequences: \shapes{}, \poster{}, \boxes{} and \dynamic{},
which feature indoor scenes with different texture complexity.
The GT poses %
are output at \mbox{200~Hz} from a motion capture system (mocap).
For accuracy evaluation (\cref{sec:experim:real_data}), we use the ECD data from 1 to 11~s, where the camera translation is relatively small (see 
Appendix~\ref{sec:suppl:camera_trans}). %
For fast-motion experiments (\cref{sec:experim:fast_hdr}), we use the highest-speed segments (50--55~s) of \poster{} and \boxes{}, where the camera's angular velocity reaches up to 800 and 600 $^\circ/s$, respectively.

The EFRD dataset is recorded using a DAVIS346 camera.
We use four sequences (\bicycles{}, \building{}, \staircase{} and \miscellany{}) in our experiments.
They exhibit fast rotation (\bicycles{}, \staircase{} and \miscellany{}), HDR (\bicycles{} and \building{}) and low-light conditions (\staircase{} and \miscellany{}).

\subsubsection{Initialization}
For bootstrapping, we first feed the event data into one of four front-end methods, namely \esmt{} \cite{Kim18phd}, \rtpt{} \cite{Reinbacher17iccp}, \cmaxgae{} \cite{Kim21ral} and \cmaxw{} \cite{Gallego17ral} (see the comparison in \cite{Guo24tro}).
Then these front-end--estimated camera rotations are passed (together with the events) to the mapping module of \esmt{}, as implemented in \cite{Guo24tro}, which produces an initial intensity map (e.g., top row of \cref{fig:map_comp}).
We interpolate the front-end rotations at 1~kHz and align them to the GT ones at $t = t_0$ ($t_0 = 0.1$~s for synthetic data and $t_0 = 1$~s for real data) before they are used to obtain initial maps and initialize EPBA.
Unless otherwise specified, the map size is set to $1024 \times 512$ px and the control pose frequency $f$ is set to 20~Hz.
We also show in \cref{sec:experim:wild,sec:experim:map_only} that the map initialization is not strictly needed. 
\EPBA{} can recover the intensity map from scratch (e.g., zeros or random noise).

\subsubsection{Evaluation Metrics}
\label{sec:experim:setup:metrics}
We evaluate EPBA using the Absolute Rotation Error (ARE), which measures the accuracy of the estimated camera rotations, and the Photometric Error (PhE), which assesses the consistency of the event data with the refined camera rotations and map.

\emph{Absolute Rotation Error (ARE)}.
At timestamp $t_k$, the error between the estimated rotation $\Rot_k$ and the corresponding GT rotation $\Rot_k'$ (computed by linear interpolation), is defined by the angle of their difference $\Delta \Rot_k = \Rot_k'^\top \Rot_k$ \cite{Barfoot15book}.
Because the output rotations of each front-end method have different frequencies, we calculate the errors at such timestamps and compute the Root Mean Square Error (RMSE) to quantify the accuracy \cite{Guo24tro}.
The refined rotations share the same control pose timestamps (regardless of the front-ends), where errors are calculated.

\emph{Photometric Error (PhE)}.
The PhE measures the goodness of fit between the event data, the estimated variables and the sensor model.
It is a standard criterion to assess the performance of BA algorithms.
In our case, it is computed by means of \eqref{eq:ObjFunc}, by aggregating the per-event photometric errors into a single value.

Note that the ARE and PhE values in \cref{tab:synth:cg:small:rmse,tab:synth:cg:small:phe,tab:real:cg:small:rmse,tab:real:cg:small:phe} are obtained using the CG solver. 
The results corresponding to the Cholesky solver are given in Appendix \ref{sec:suppl:cholesky}.

\subsection{Experiments on Synthetic Data}
\label{sec:experim:synth_data}

\begin{table}
\centering
\caption{\label{tab:synth:cg:small:rmse}
Absolute rotation RMSE [deg] on synthetic sequences.  
The best results per sequence are in bold.
``-'' means the method fails on that sequence, 
and ``N/A'' indicates that EPBA is not applicable because the corresponding front-end fails on this sequence.
\rtpt{} is not shown because it fails on all sequences. 
For the analysis of front-end failures, see \cite{Guo24tro}.}
\adjustbox{max width=\linewidth}{
\setlength{\tabcolsep}{3pt}
\begin{tabular}{ll*{6}{S[table-format=1.3,table-number-alignment=center]}}
\toprule

Front-end & Trajectory & \text{playroom} & \text{bicycle} & \text{city} & \text{street} & \text{town} & \text{bay} \\
\midrule
\multirow{6}{*}{\shortstack{\esmt{}}}
& before BA & 5.861 & 1.466 & 1.692 & 3.441 & 4.322 & 2.5 \\
& \EMBA{} & 6.094 & 1.182 & 1.675 & 3.456 & 4.400 & 2.412 \\
& Ours (Quad) & 4.757 & 0.814 & 1.411 & 2.857 & 4.062 & 2.359 \\
& Ours (Huber) & 5.012 & 0.558 & 0.543 & 2.65 & 4.22 & 3.433 \\
& Ours (Cauchy) & 5.268 & 0.558 & 0.544 & 2.622 & 4.241 & 3.472 \\

\midrule
\multirow{6}{*}{\shortstack{\cmaxgae{}}}
& before BA & 4.628 & 1.651 & \novalue & \novalue & 4.656 & \novalue \\
& \EMBA{} & 4.419 & 1.496 & \NA{} & \NA{} & 4.534 & \NA{} \\
& Ours (Quad) & 3.538 & 1.248 & \NA{} & \NA{} & 4.295 & \NA{} \\
& Ours (Huber) & 1.946 & 1.227 & \NA{} & \NA{} & 3.762 & \NA{} \\
& Ours (Cauchy) & 2.093 & 1.22 & \NA{} & \NA{} & 3.786 & \NA{} \\

\midrule
\multirow{6}{*}{\shortstack{\cmaxw{}}}
& before BA & 3.223 & 1.69 & 1.532 & 0.965 & 1.905 & 1.797 \\
& \cmaxslam{} & 0.763 & 0.327 & 0.509 & 0.470 & 0.553 & 0.617 \\
& \EMBA{} & 2.856 & 0.923 & 0.973 & 0.744 & 0.858 & 1.409 \\
& Ours (Quad) & 1.066 & 0.195 & 0.571 & 0.543 & 1.19 & 1.395 \\
& Ours (Huber) & 0.587 & 0.213 & \bnum{0.154} & \bnum{0.152} & \bnum{0.188}& 0.934 \\
& Ours (Cauchy) & \bnum{0.548} & \bnum{0.193} & 0.156 & 0.155 & 0.196 & \bnum{0.893} \\

\bottomrule
\end{tabular}
}
\end{table}

\begin{table}
\centering
\caption{\label{tab:synth:cg:small:phe}
Squared photometric error [$\times 10^6$] on synthetic sequences.
To measure PhE values for \cmaxslam{} and \EMBA{} using \eqref{eq:ObjFunc}, 
we adopt map-only \EPBA{} (\cref{sec:experim:map_only}).
It reconstruct maps (from the output rotations of \cmaxslam{} and \EMBA{}) on which PhE values are calculated.}
\adjustbox{max width=\linewidth}{
\setlength{\tabcolsep}{3pt}
\begin{tabular}{ll*{6}{S[table-format=1.3,table-number-alignment=center]}}
\toprule

Front-end & Trajectory & \text{playroom} & \text{bicycle} & \text{city} & \text{street} & \text{town} & \text{bay} \\
\midrule
\multirow{6}{*}{\shortstack{\esmt{}}}
& before BA & 0.683 & 0.458 & 0.963 & 0.782 & 0.685 & 0.698 \\
& \EMBA{} & 0.182 & 0.197 & 0.527 & 0.467 & 0.425 & 0.477 \\
& Ours (Quad) & 0.119 & 0.115 & 0.444 & 0.386 & 0.392 & 0.435 \\
& Ours (Huber) & 0.176 & 0.121 & 0.434 & 0.465 & 0.447 & 0.491 \\
& Ours (Cauchy) & 0.229 & 0.119 & 0.435 & 0.478 & 0.455 & 0.497 \\

\midrule
\multirow{6}{*}{\shortstack{\cmaxgae{}}}
& before BA & 0.675 & 0.680 & \novalue & \novalue & 0.806 & \novalue \\
& \EMBA{} & 0.148 & 0.188 & \NA{} & \NA{} & 0.400 & \NA{} \\
& Ours (Quad) & 0.113 & 0.178 & \NA{} & \NA{} & 0.337 & \NA{} \\
& Ours (Huber) & 0.129 & 0.199 & \NA{} & \NA{} & 0.394 & \NA{} \\
& Ours (Cauchy) & 0.151 & 0.201 & \NA{} & \NA{} & 0.395 & \NA{} \\

\midrule
\multirow{6}{*}{\shortstack{\cmaxw{}}}
& before BA & 0.913 & 0.632 & 2.121 & 1.571 & 1.406 & 1.764 \\
& \cmaxslam{} & 0.079 & 0.167 & 0.477 & 0.362 & 0.349 & 0.395 \\
& \EMBA{} & 0.104 & 0.201 & 0.477 & 0.375 & 0.398 & 0.424 \\
& Ours (Quad) & 0.088 & 0.113 & 0.390 & 0.293 & 0.316 & 0.393 \\
& Ours (Huber) & 0.102 & 0.124 & 0.428 & 0.321 & 0.336 & 0.445 \\
& Ours (Cauchy) & 0.116 & 0.120 & 0.430 & 0.322 & 0.338 & 0.462 \\

\bottomrule
\end{tabular}
}
\end{table}

First, let us present the results of EPBA on synthetic data.
The comparison of initial and refined rotations on \playroom{} and \bicycle{} are depicted in \cref{fig:traj:playroom,fig:traj:bicycle}, respectively,
where the refined rotations fit better with the GT than the initial ones. 
The accuracy of camera rotations before and after EPBA refinement is quantified in \cref{tab:synth:cg:small:rmse}, 
where the results of \cmaxslam{} \cite{Guo24tro} and \EMBA{} \cite{Guo24eccv} are also reported for comparison.

Due to EPBA refinement, the errors decrease on all synthetic sequences.
The effect is most noticeable when initialized by \cmaxw{}.
For example, the RMSE of the \cmaxw{} rotations on \playroom{} decreases from 3.223$^\circ$ to 1.066$^\circ$ (quadratic loss),
and that of \bicycle{} is reduced from 1.69$^\circ$ to 0.195$^\circ$.
The improvements (percentage decrease) for \playroom{}, \bicycle{} and \city{} are more than $60\%$, while those of \street{} and \town{} are also around $50\%$.
For each front-end, the accuracy of \EPBA{} refined rotations is better on almost all sequences than those of \cite{Guo24eccv}.
The only exception is the refinement of \cmaxw{} rotations on the \town{} sequence.
Compared to \cmaxslam{}, \EPBA{} reports smaller rotation error on all sequences except \playroom{}.

\def\figWidth{0.32\linewidth}
\begin{figure*}[t]
	\centering
    {\small
    \setlength{\tabcolsep}{1pt}
	\begin{tabular}{
	>{\centering\arraybackslash}m{0.35cm} 
	>{\centering\arraybackslash}m{\figWidth} 
	>{\centering\arraybackslash}m{\figWidth}
	>{\centering\arraybackslash}m{\figWidth}
        }
        
        \rotatebox{90}{\makecell{Scene}}
		&\includegraphics[width=\linewidth]{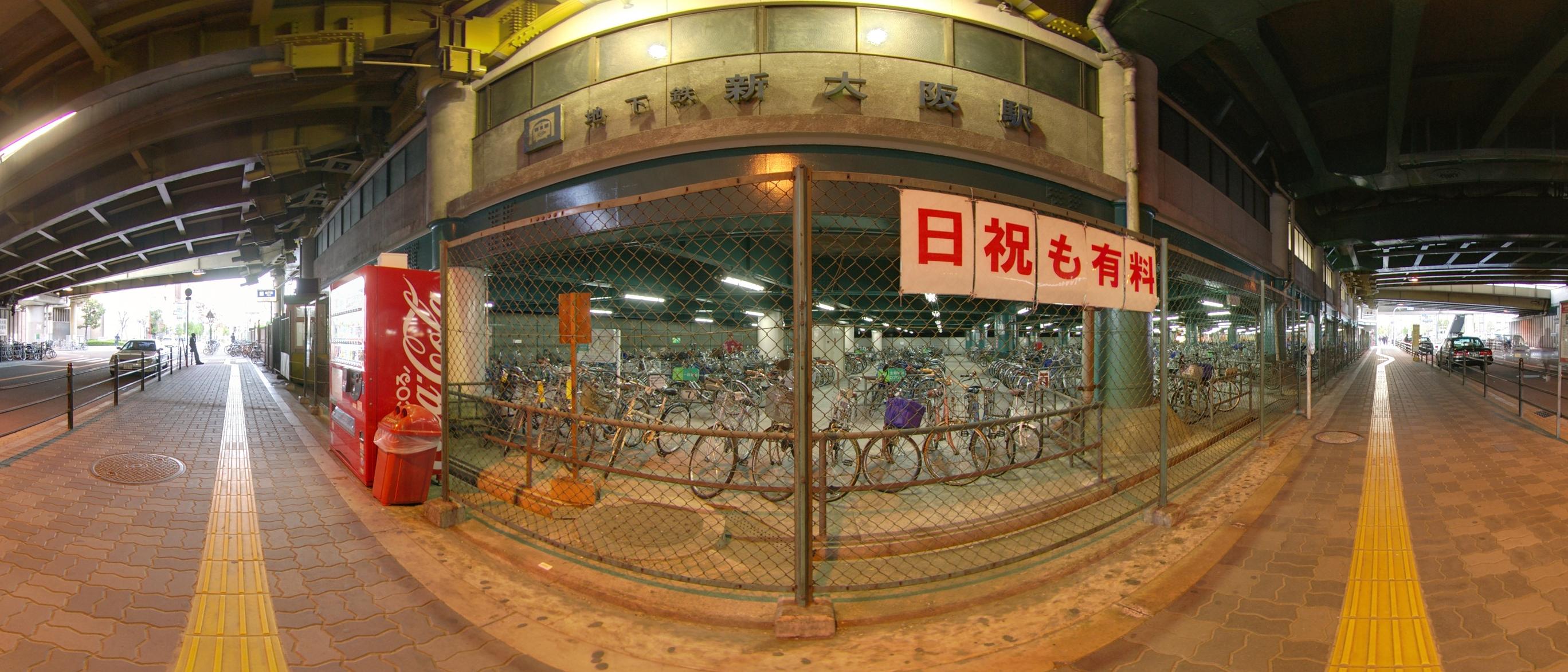}
            &\includegraphics[width=\linewidth]{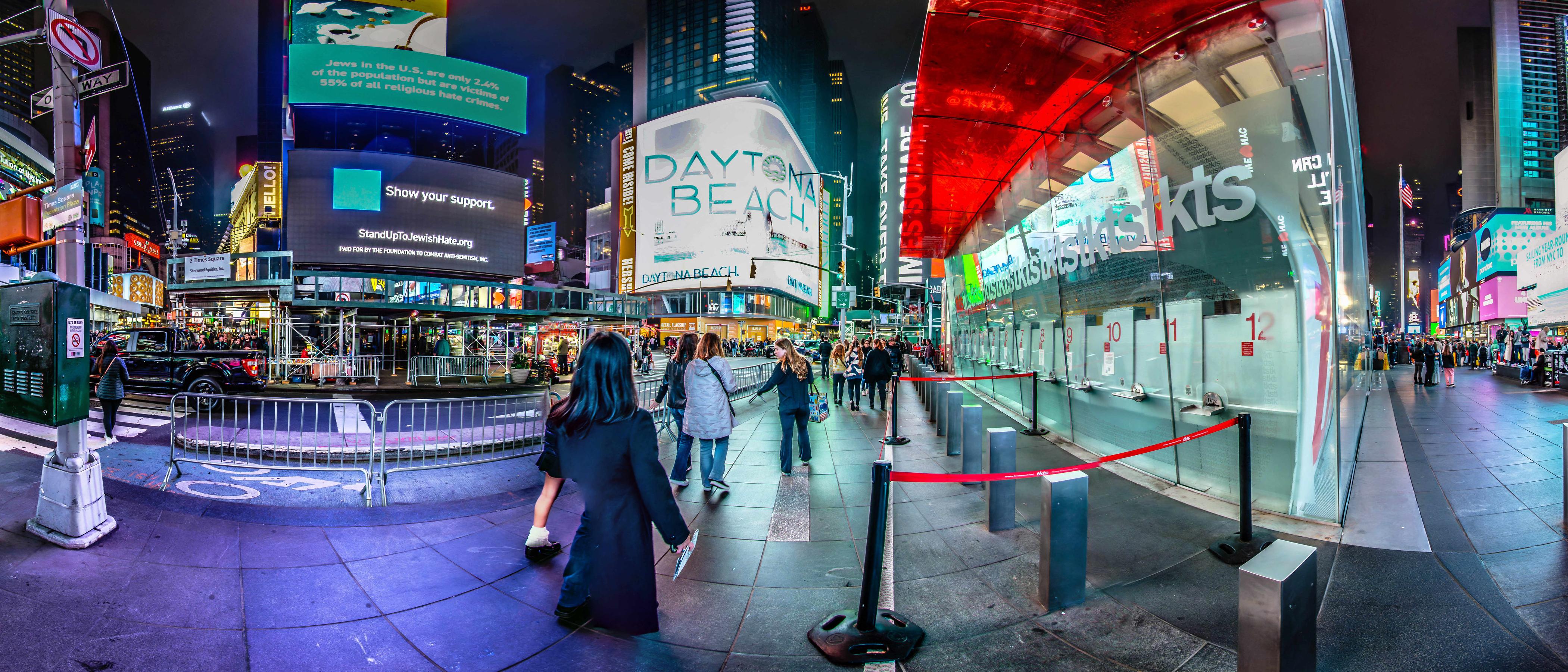}
            &\includegraphics[width=\linewidth]{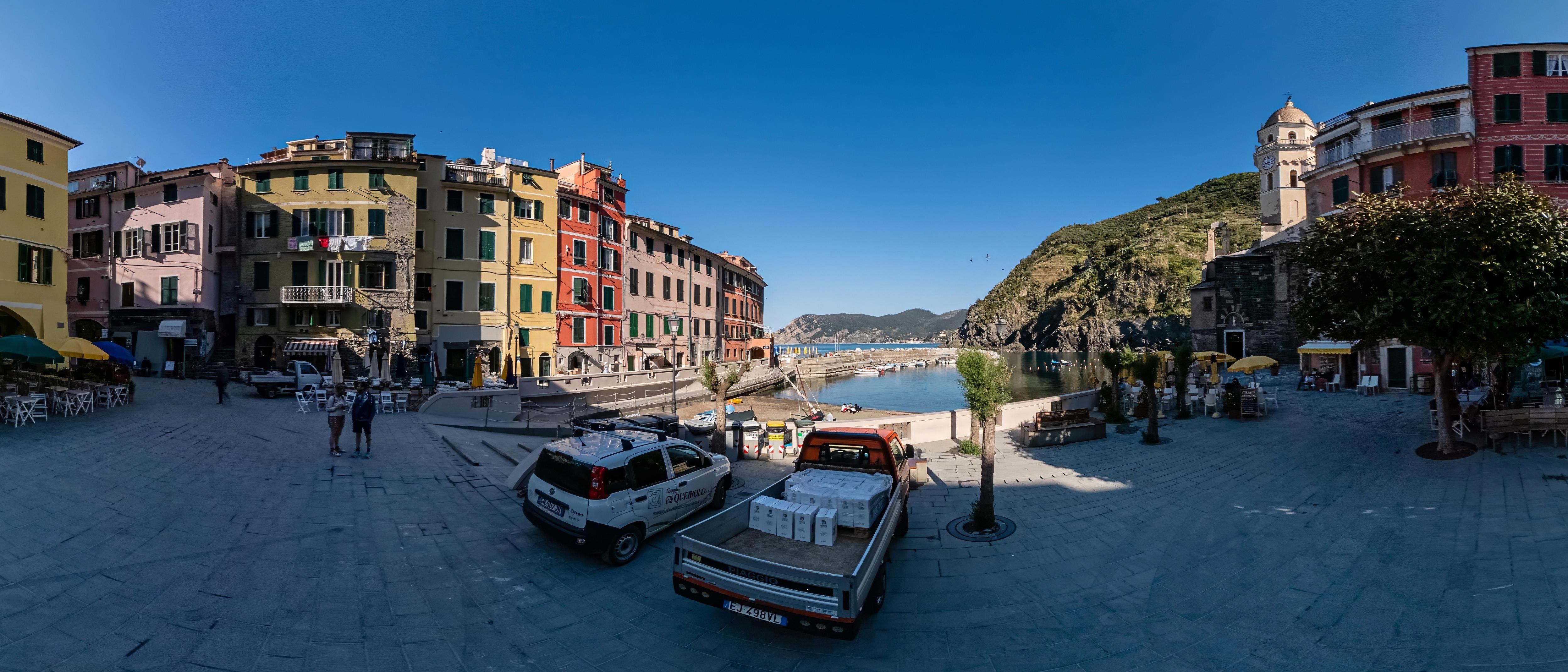}
		\\

        \rotatebox{90}{\makecell{Initial map}}
		&\includegraphics[width=\linewidth]{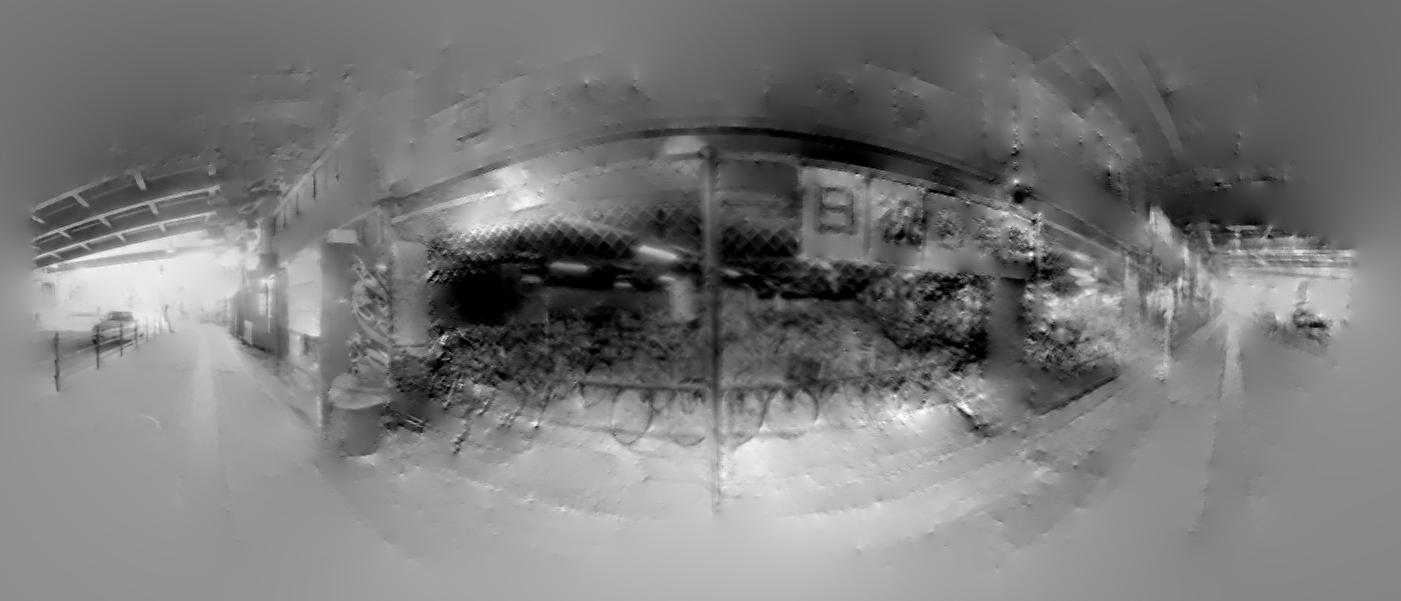}
		&\includegraphics[width=\linewidth]{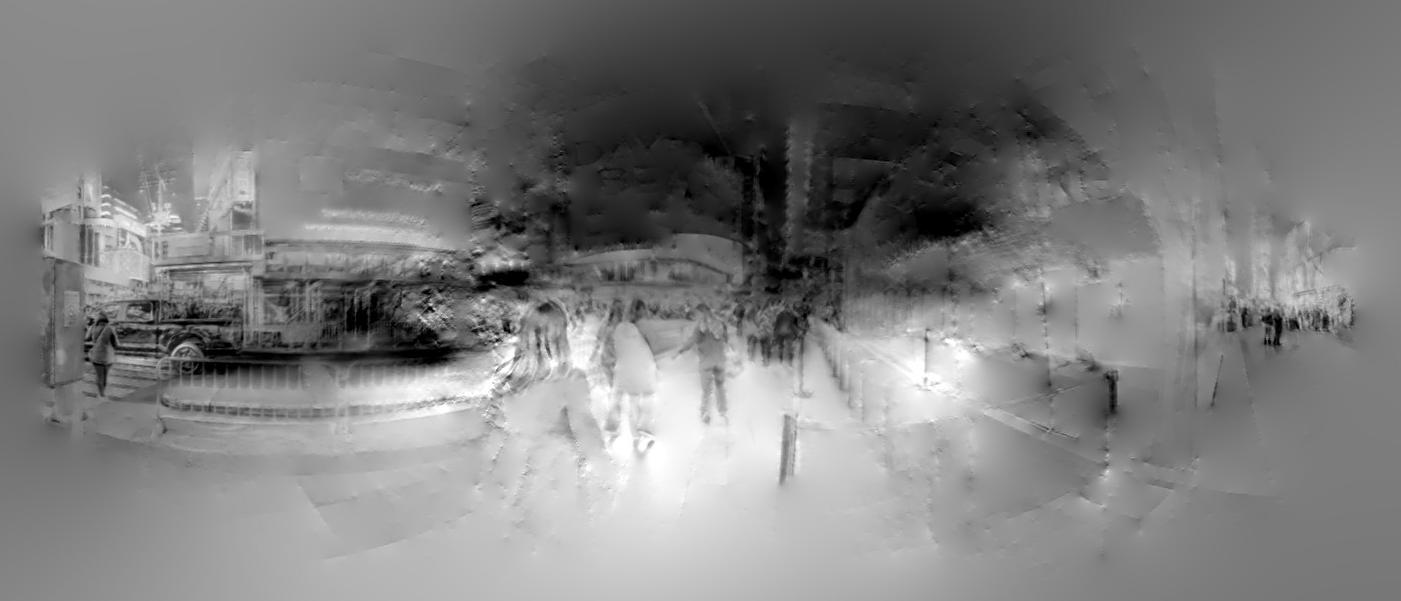}
            &\includegraphics[width=\linewidth]{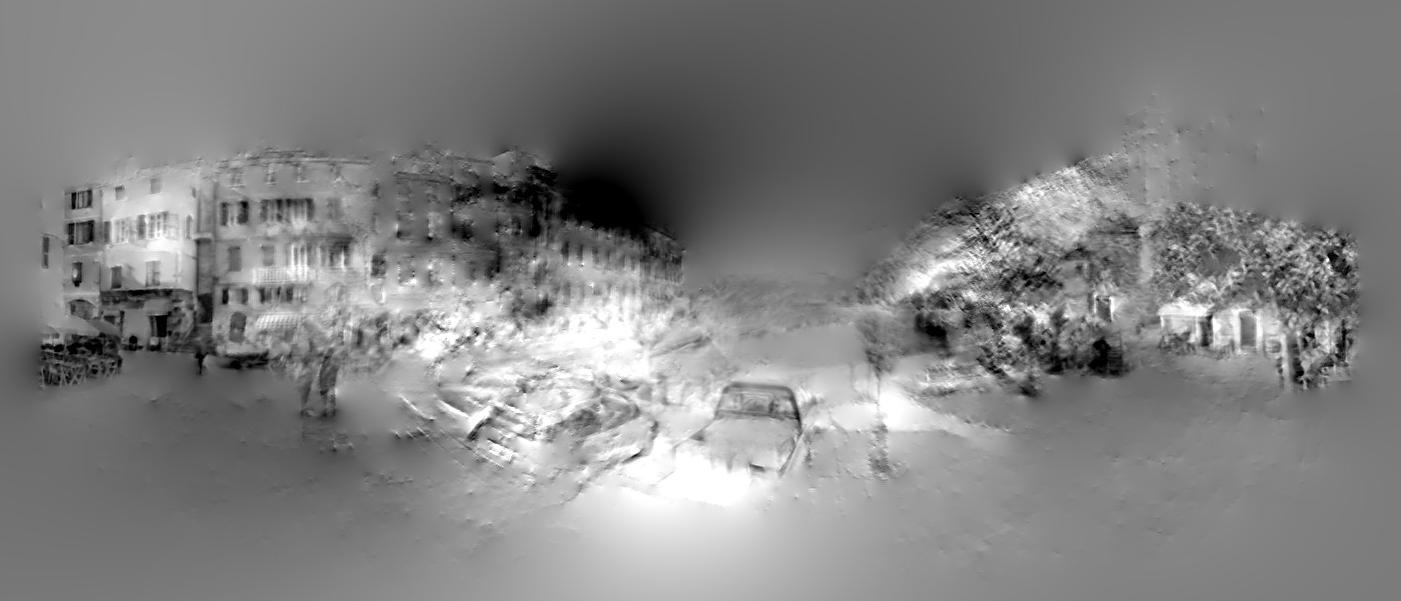}
		\\
  
        \rotatebox{90}{\makecell{Refined map}}
		&\includegraphics[width=\linewidth]{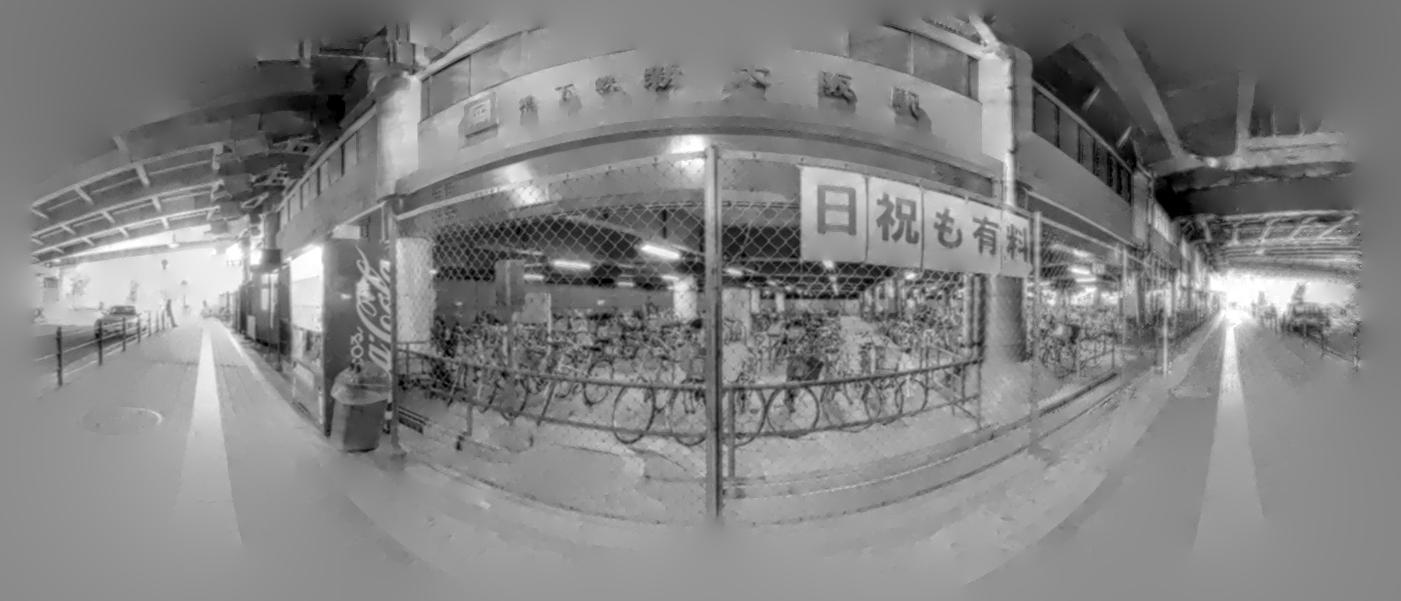}
		&\includegraphics[width=\linewidth]{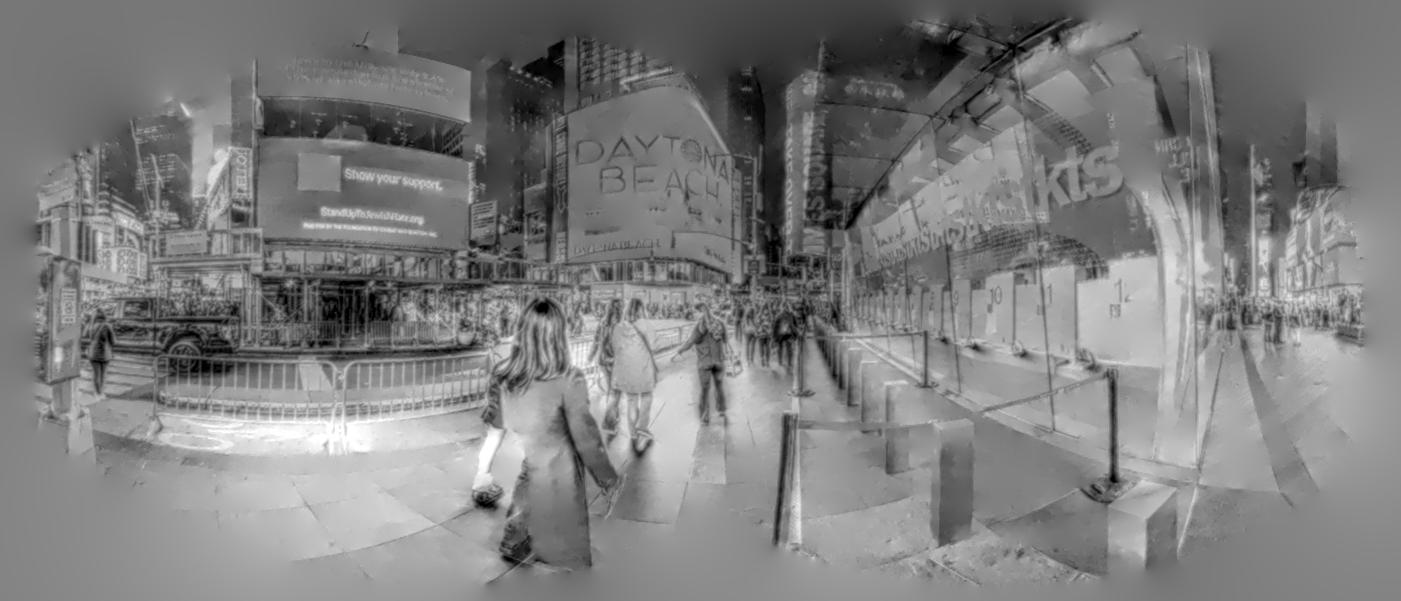}
            &\includegraphics[width=\linewidth]{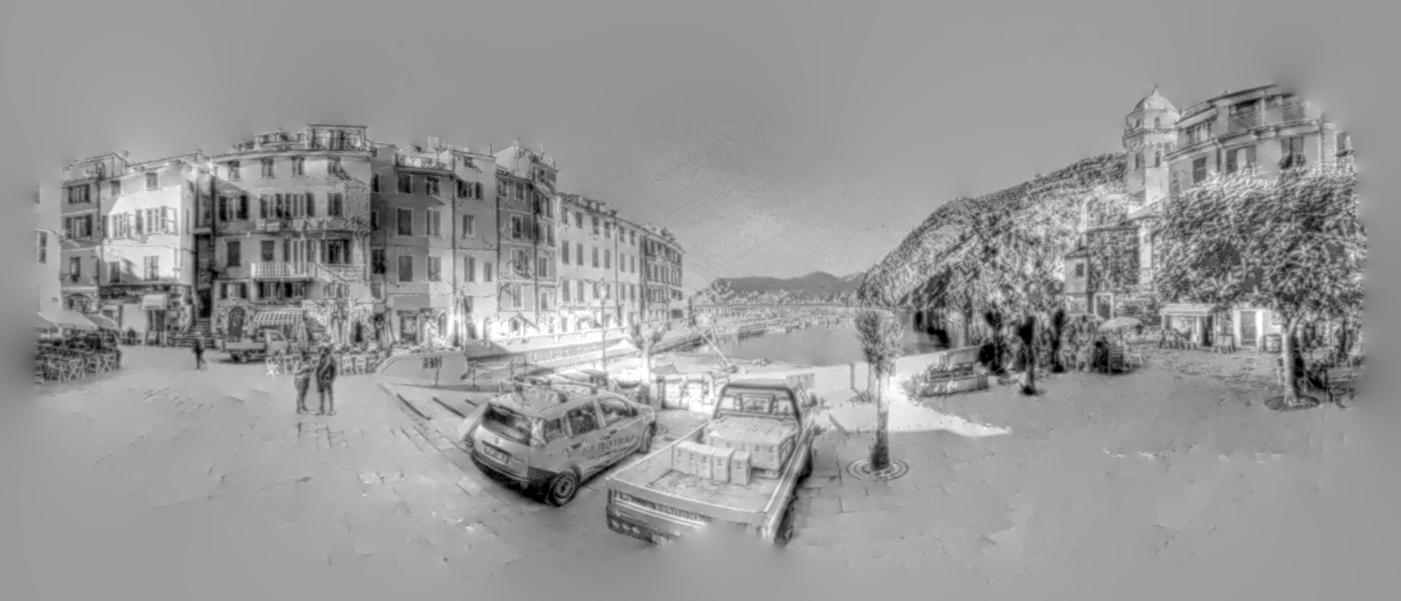}
		\\

        & (a) \bicycle{}
        & (b) \street{}
        & (c) \town{}
        \\
	\end{tabular}
	}
    \vspace{-1ex}
    \caption{EPBA results on synthetic data from \cite{Guo24tro}. 
    Estimated maps have $2048 \times 1024$ px. 
    Initial camera rotations are obtained by integrating the angular velocities estimated 
    using \cmaxw{}~\cite{Gallego17ral}.
    \label{fig:synth_refine}
    }
\end{figure*}

\begin{figure}[t]
     \centering
     \begin{subfigure}[b]{0.48\linewidth}
         \centering
         \includegraphics[width=\linewidth]{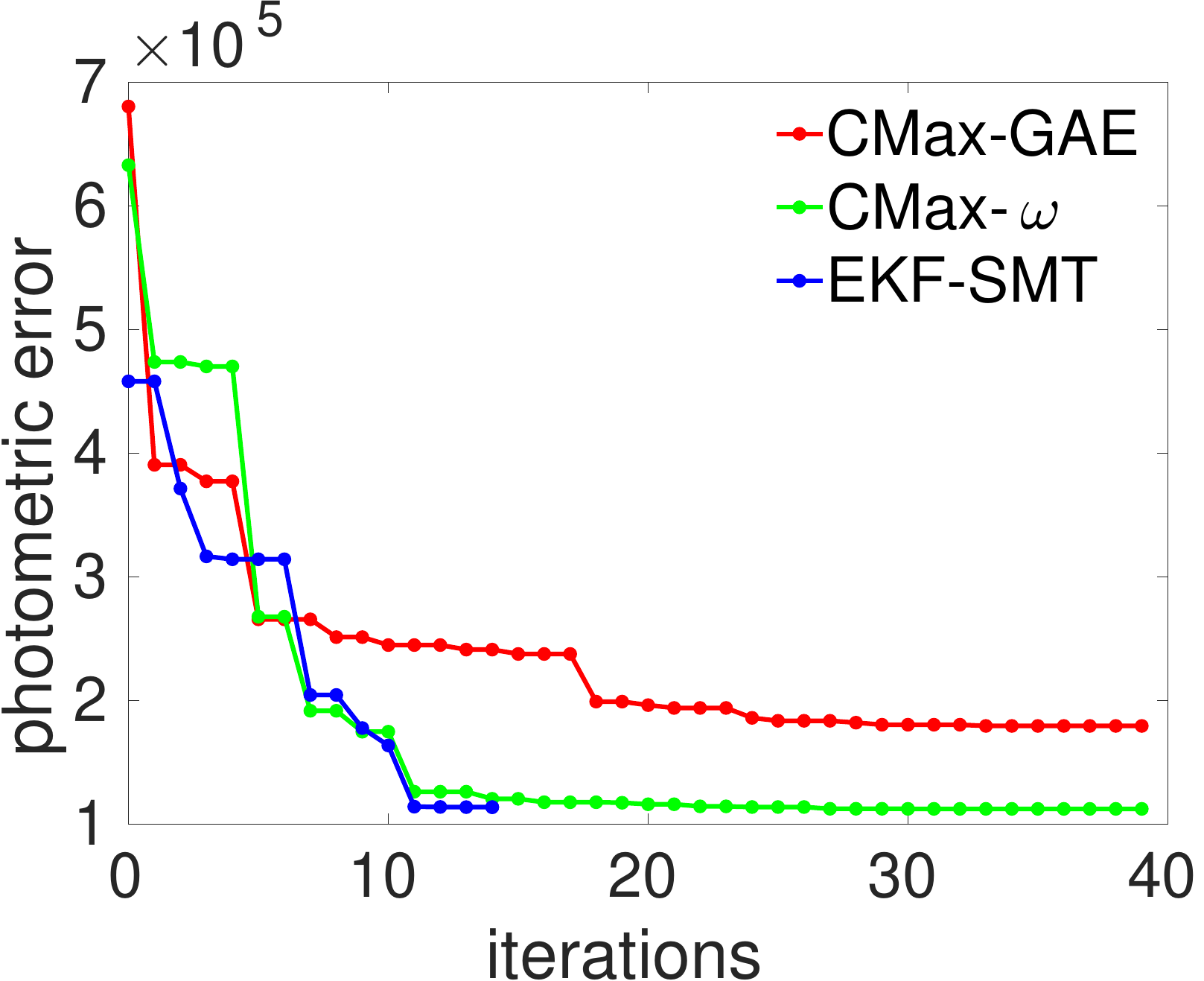}
         \caption{Evolution of loss value.}
         \label{fig:error_evolution:decreasing}
     \end{subfigure}\;\;
     \begin{subfigure}[b]{0.46\linewidth}
         \centering
         \includegraphics[width=\linewidth]{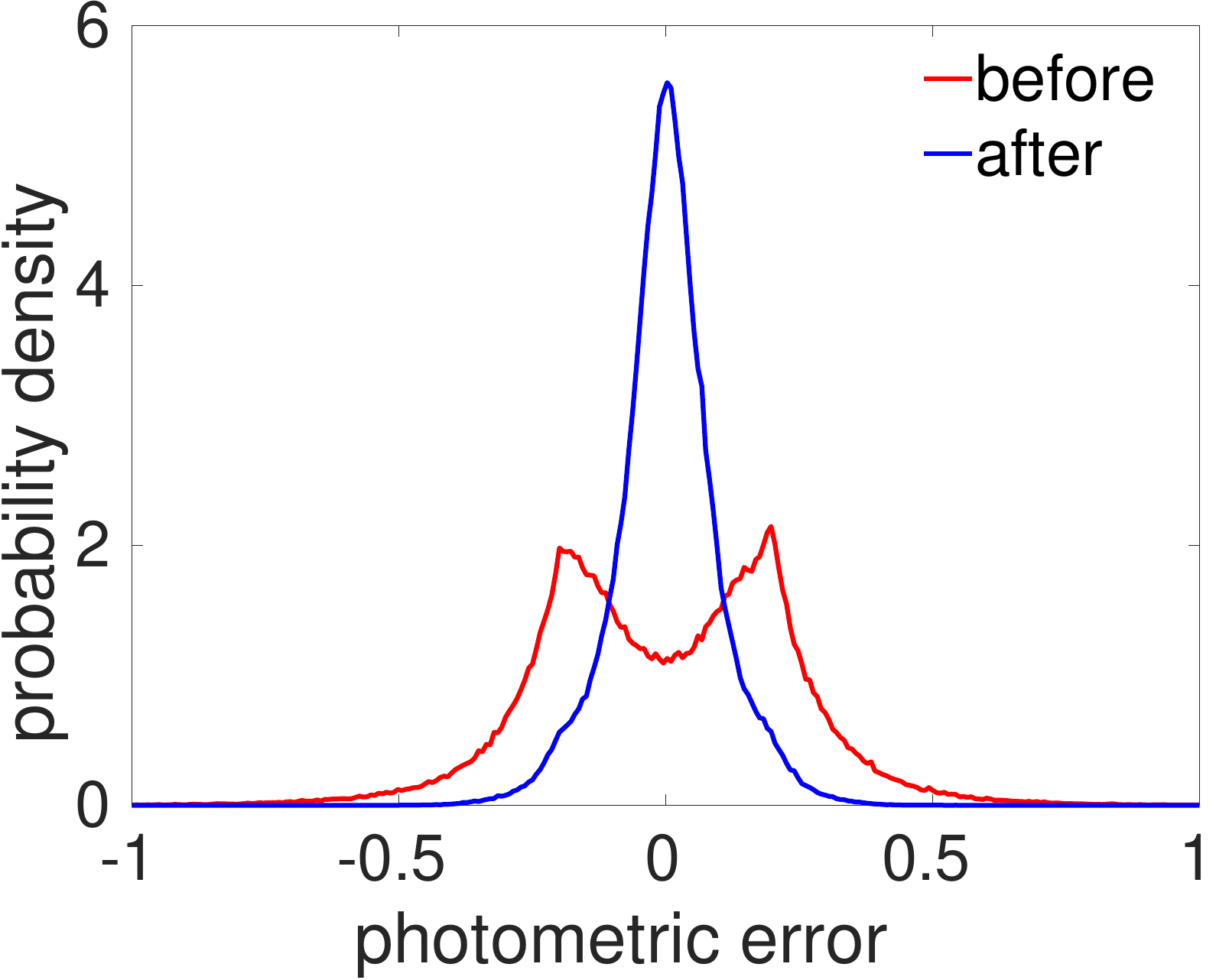}
         \caption{Error PDFs for \cmaxw{}.}
         \label{fig:error_evolution:pdf}
     \end{subfigure}
        \caption{Evolution of the loss value on the \bicycle{} sequence using EPBA 
        initialized with camera rotations from three front-end methods.
        \label{fig:error_evolution}
        }
\end{figure}

The improvement effect of EPBA is also obvious in the PhE and the map quality: \cref{fig:map_comp,fig:synth_refine}, and \cref{tab:synth:cg:small:phe}.
In most tests, the PhE is reduced by more than $50\%$ (\cref{tab:synth:cg:small:phe}).
The maximal relative decrease is bigger than $90\%$, e.g., refining the \cmaxw{} rotations on \playroom{}.
\Cref{tab:synth:cg:small:phe} also reveals that \EPBA{} (quadratic loss) achieves smaller PhE than \cite{Guo24eccv} for all front-ends and on all sequences,
as well as for \cmaxslam{} on five out of six sequences (only except \playroom{}).
The evolution of the PhE on \bicycle{} with three different initializations is displayed in \cref{fig:error_evolution:decreasing}.
It shows that the PhE drops rapidly in the first ten iterations, then it slowly decreases for fine adjustment until convergence.
The refinement of \cmaxgae{} seems to get stuck in a local minimum, which may be due to the initialization.
In addition, the initial and refined probability density functions (PDFs) of the PhE for \cmaxw{} on \bicycle{} are displayed in \cref{fig:error_evolution:pdf}: 
the PDF of the initial PhE has two peaks around $\pm C$ ($\pm 0.2$), while that of the refined PhE is better behaved, with a single concentrated peak at zero.

A comparison of initial and refined maps of $2048 \times 1024$ px size of several synthetic sequences is given in \Cref{fig:synth_refine}.
\EPBA{} achieves significant improvements in the visual quality of these maps:
blurred regions are optimized to become sharp or smooth,
unbalanced brightness (due to Poisson integration at initialization via \esmt{}) is overcome,
and the fine details that are not visible in the initial maps are revealed,
such as the billboards in \street{}, the \bicycle{} wheels, and the tree leaves in \town{}.

Furthermore, we also investigate the effect of the quadratic, Huber and Cauchy loss functions (\cref{sec:method:robust}) in \cref{tab:synth:cg:small:rmse,tab:synth:cg:small:phe}.
It turns out that the Huber and Cauchy loss functions further improve \EPBA{}'s capability for decreasing the rotation error.
For instance, the RMSE of the \cmaxw{} rotations on \town{} is reduced from 1.19$^\circ$ (quadratic -- ``Quad'') to 0.188$^\circ$ (Huber) and 0.196$^\circ$ (Cauchy), where the relative decrease is $>80\%$.
When initialized by \cmaxw{}, both Huber and Cauchy loss functions achieve errors $<1^\circ$ (in the range 0.15--0.94$^\circ$).
Taking into account the variations with robust cost functions,
\EPBA{} achieves the best rotation accuracy on all sequences.
For the impact of different loss functions on the refined map, which is difficult to show with still images, we provide some intuitive animations in the accompanying video.
On the other hand, the refined PhE of the Huber and Cauchy loss functions is slightly bigger than the quadratic one.
This is expected, as the objective function has changed to a reweighed squared PhE,
where the weights of the outliers are reduced \cite{Barfoot15book}.
However, the refined camera rotations and maps are better than those of the quadratic loss.

In a word, on synthetic data, \EPBA{} achieves a comprehensive refinement in terms of rotation accuracy, map quality and photometric error.

\begin{table}
\centering
\caption{
\label{tab:real:cg:small:rmse} 
Absolute rotation RMSE [deg] on real data.
\esmt{} is not shown since this front-end fails on all sequences~\cite{Guo24tro}.
}
\adjustbox{max width=\linewidth}{
\setlength{\tabcolsep}{6pt}
\begin{tabular}{ll*{4}{S[table-format=1.3,table-number-alignment=center]}}
\toprule
Front-end & Trajectory & \text{shapes} & \text{poster} & \text{boxes} & \text{dynamic} \\
\midrule
\multirow{6}{*}{\shortstack{\rtpt{}}}
& before BA & 2.187 & 3.802 & 1.743 & 2.000 \\
& \EMBA{} & 2.850 & 3.958 & 2.319 & 2.285 \\
& Ours (Quad) & 2.796 & 4.059 & 2.089 & 2.537 \\
& Ours (Huber) & 2.926 & 4.075 & 2.128 & 2.788 \\
& Ours (Cauchy) & 2.933 & 4.074 & 2.559 & 2.722 \\

\midrule
\multirow{6}{*}{\shortstack{\cmaxgae{}}}
& before BA & 2.512 & 3.625 & 2.018 & 1.698 \\
& \EMBA{} & 2.691 & 4.094 & 2.400 & 2.004 \\
& Ours (Quad) & 3.099 & 4.627 & 2.255 & 2.560 \\
& Ours (Huber) & 2.944 & 4.446 & 2.851 & 2.680 \\
& Ours (Cauchy) & 2.940 & 4.458 & 2.940 & 2.726 \\

\midrule
\multirow{6}{*}{\shortstack{\cmaxw{}}}
& before BA & 4.111 & 4.072 & 3.224 & 3.126 \\
& \cmaxslam{} & 4.953 & 5.653 & 5.418 & 3.380 \\
& \EMBA{} & 4.441 & 4.196 & 2.866 & 2.791 \\
& Ours (Quad) & 3.020 & 4.122 & 2.785 & 2.987 \\
& Ours (Huber) & 2.951 & 4.127 & 2.740 & 2.858 \\
& Ours (Cauchy) & 2.959 & 4.123 & 2.737 & 2.847 \\

\bottomrule
\end{tabular}
}
\end{table}

\begin{table}
\centering
\caption{\label{tab:real:cg:small:phe}
Squared photometric error [$\times 10^5$] on real data from \cite{Mueggler17ijrr}.
Same evaluation procedure as \cref{tab:synth:cg:small:phe}.
}
\adjustbox{max width=\linewidth}{
\setlength{\tabcolsep}{6pt}
\begin{tabular}{ll*{4}{S[table-format=1.3,table-number-alignment=center]}}
\toprule
Front-end & Trajectory & \text{shapes} & \text{poster} & \text{boxes} & \text{dynamic} \\
\midrule
\multirow{6}{*}{\shortstack{\rtpt{}}}
& before BA & 0.723 & 5.535 & 4.792 & 3.474 \\
& \EMBA{} & 0.292 & 3.217 & 2.796 & 2.302 \\
& Ours (Quad) & 0.192 & 1.955 & 2.842 & 2.266 \\
& Ours (Huber) & 0.208 & 2.083 & 3.256 & 2.585 \\
& Ours (Cauchy) & 0.208 & 2.077 & 3.221 & 2.685 \\

\midrule
\multirow{6}{*}{\shortstack{\cmaxgae{}}}
& before BA & 0.750 & 5.782 & 4.667 & 3.539 \\
& \EMBA{} & 0.445 & 3.483 & 2.873 & 2.461 \\
& Ours (Quad) & 0.264 & 3.050 & 2.877 & 2.345 \\
& Ours (Huber) & 0.229 & 3.667 & 3.300 & 2.791 \\
& Ours (Cauchy) & 0.237 & 3.706 & 3.384 & 2.860 \\

\midrule
\multirow{6}{*}{\shortstack{\cmaxw{}}}
& before BA & 0.553 & 4.345 & 3.736 & 2.914 \\
& \cmaxslam{} & 0.326 & 2.567 & 2.302 & 2.099 \\
& \EMBA{} & 0.253 & 3.255 & 2.768 & 2.212 \\
& Ours (Quad) & 0.192 & 1.953 & 1.620 & 1.645 \\
& Ours (Huber) & 0.208 & 2.077 & 1.699 & 1.799 \\
& Ours (Cauchy) & 0.208 & 2.072 & 1.694 & 1.813 \\

\bottomrule
\end{tabular}
}
\end{table}

\def\figWidth{0.24\linewidth}
\begin{figure*}[t]
	\centering
    {\small
    \setlength{\tabcolsep}{1pt}
	\begin{tabular}{
	>{\centering\arraybackslash}m{0.4cm} 
	>{\centering\arraybackslash}m{\figWidth} 
	>{\centering\arraybackslash}m{\figWidth}
        >{\centering\arraybackslash}m{\figWidth}
        >{\centering\arraybackslash}m{\figWidth}}

        \rotatebox{90}{\makecell{Map from GT rots.}}
            &\includegraphics[width=\linewidth]{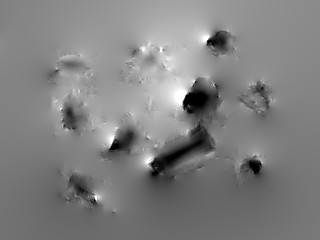}
		&\includegraphics[width=\linewidth]{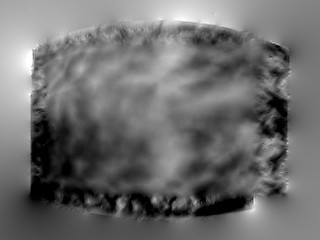}
            &\includegraphics[width=\linewidth]{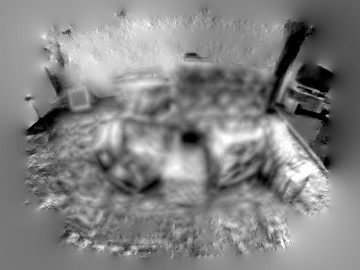}
            &\includegraphics[width=\linewidth]{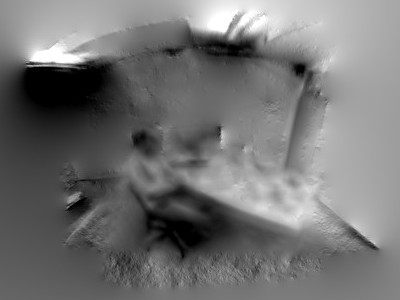}
		\\
  
        \rotatebox{90}{\makecell{Initial map}}
		&\includegraphics[width=\linewidth]{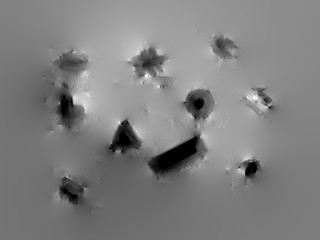}
		&\includegraphics[width=\linewidth]{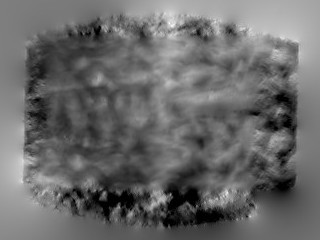}
            &\includegraphics[width=\linewidth]{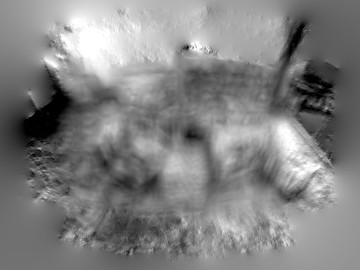}
            &\includegraphics[width=\linewidth]{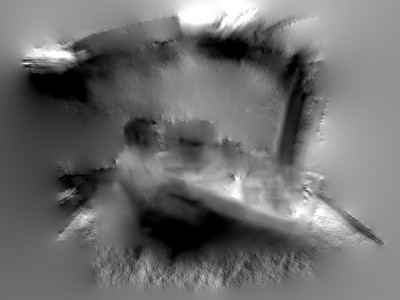}
		\\
  
        \rotatebox{90}{\makecell{Refined map}}
		&\includegraphics[width=\linewidth]{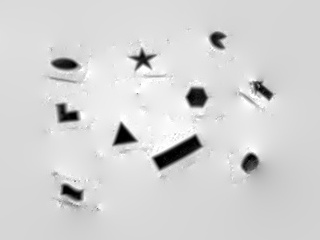}
		&\includegraphics[width=\linewidth]{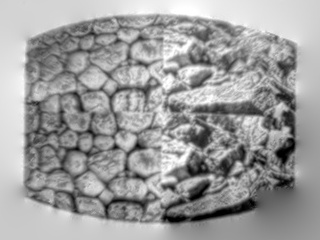}
		&\includegraphics[width=\linewidth]{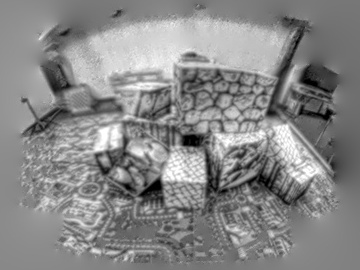}
            &\includegraphics[width=\linewidth]{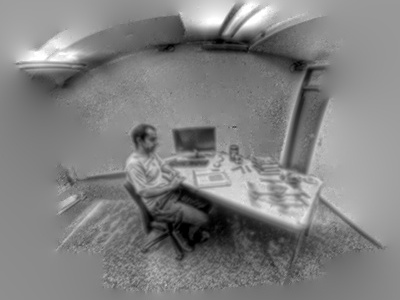}
		\\

        & (a) \shapes{}
        & (b) \poster{}
        & (c) \boxes{}
        & (d) \dynamic{}
        \\
	\end{tabular}
	}
    \vspace{-1ex}
    \caption{Results of EPBA on real-world data from \cite{Mueggler17ijrr}. 
    The maps in the top two rows are obtained using the mapping module of SMT \cite{Kim14bmvc}, 
    by feeding the GT camera rotations or the rotations estimated using \cmaxw{}, respectively. 
    Crops from $1024 \times 512$~px panoramic maps.
    \label{fig:real_refine}
    }
\end{figure*}

\subsection{Experiments on Real-world Data}
\label{sec:experim:real_data}
As pointed out in \cite{Guo24tro}, the main difficulty of real-world evaluation lies in utilizing real data that conforms with the purely rotational motion assumption of the problem. 
Real-world sequences are recorded hand-held and contain residual translations, 
which affect the events, and in turn affect the estimated rotations. 
For evaluation purposes, comparing such estimated motion to the GT rotations from a \mbox{6-DOF} mocap system \cite{Mueggler17ijrr} can be misleading if the residual translations are non-negligible.
Therefore, in photometric BA, we turn to the PhE as an overall sensible figure of merit.

\Cref{fig:traj:boxes,fig:traj:dynamic} show the initial and refined \cmaxw{} rotations on \boxes{} and \dynamic{}, respectively, whose differences are small at this scale.
\Cref{tab:real:cg:small:rmse} reports the ARE of all front-ends with respect to the mocap GT rotations;
the refined ARE fluctuates slightly above and below the initial values.
In all these results there are no big differences between the camera rotations before and after refinement because all of them contain compensation for the translational motion component.
\EPBA{} reports similar ARE as \cite{Guo24eccv}, while shows smaller ARE than \cmaxslam{}.

In contrast, the advantages of \EPBA{} are demonstrated in the reduction of the PhE values (\cref{tab:real:cg:small:phe}) and the promotion of the map quality (\cref{fig:real_refine}). 
On real-world data, the decrease of PhE varies between $30\%$ and $70\%$. 
In \cref{tab:real:cg:small:phe}, \EPBA{} achieves better PhE reduction than \cite{Guo24eccv} in nearly all trials, while \cmaxslam{} shows markedly worse performance.
In terms of map quality, a notable visual improvement is achieved (see \cref{fig:real_refine}). 
After \EPBA{} refinement, the maps become much sharper and smoother than the initial ones. 
Some subtle details (e.g., the textures on the stones in \poster{} and the patterns on the carpet in \boxes{}) are recovered. 
For comparison, to show the influence of camera translation, we input the GT rotations from the mocap into the mapping module of \esmt{}, and display the reconstructed maps (which appear to be blurred) in the top row of \cref{fig:real_refine}.

The impact of the choice of loss function is quantitatively reported in \cref{tab:real:cg:small:rmse,tab:real:cg:small:phe}.
However, due to the inherent difficulties of real-world evaluation mentioned above, it is hard to conclude whether the Huber and Cauchy loss functions promote rotation accuracy (ARE); they remain about the same.
Similar to the synthetic data, the refined squared PhE of the Huber and Cauchy loss are bigger due to the change of objective function.
The refined maps of different loss functions are compared in the supplementary video,
where the Huber/Cauchy loss results in a better map than the quadratic one in most cases.

In short, although there are some difficulties in rotation accuracy evaluation on real-world data, the PhE criterion and the map quality still prove \EPBA{}'s effectiveness.

\subsection{Complexity Analysis and Runtime}
\label{sec:experim:runtime}

There are three main steps in \EPBA{}:
($i$) the evaluation of the objective function %
and its derivatives, %
whose complexity is $O(\numEvents)$;
($ii$) the formation of the normal equations, %
whose complexity is also $O(\numEvents)$,
and ($iii$) the solution of the %
normal equations.
For the Cholesky-based solver, the cost of solving the normal equations depends in a complicated way on the number of valid pixels $\numPixels$ and the amount of scene texture.
In the best case ($\mA_{22}$ is very sparse), the cost of Cholesky decomposition grows approximately linearly with the size of $\mA_{22}$, i.e., $\numPixels$ \cite{Boyd04book}.
In the worst case ($\mA_{22}$ is almost dense), the complexity is close to $O(\numPixels^3)$.
In general, the higher the texture, the more the events and valid pixels, so that EPBA becomes more expensive.
For the CG solver, sparsity does not play a role as strong as in Cholesky \cite{Hager06survey}. 
Convergence depends on the condition number of the system matrix.
The CG solver is iterative; and we set a tolerance of $10^{-6}$ on the norm of the residual of the linear system as termination criterion.

\begin{table}[t]
\centering
\caption{
\label{tab:runtime}
Runtime evaluation of EPBA's main steps [s].
Scene complexity order: \shapes{} $<$ \dynamic{} $<$ \boxes{} $<$ \poster{}.
}
\adjustbox{max width=\linewidth}{
\setlength{\tabcolsep}{3pt}
\begin{tabular}{lrrrr}
\toprule 
ECD sequence  & shapes & poster & boxes & dynamic \\
\midrule
Obj. func. & 1.171 & 9.143 & 7.949 & 6.234 \\
Forming Normal Eqs. & 0.306 & 4.074 & 3.936 & 2.966 \\
Solving Normal Eqs. (CG) & 0.261 & 2.430 & 2.521 & 1.938 \\
Solving Normal Eqs. (Chol.) & 3.282 & 28.044 & 50.226 & 77.458 \\
\midrule
$N_p$ (active pixels) & 14115 & 53786 & 62832 & 62760 \\
$N_e$ (number of events) & 1.78M & 12.59M & 10.76M & 8.80M \\
\bottomrule
\end{tabular}
}
\end{table}

We conduct a runtime evaluation to support the above analysis.
\Cref{tab:runtime} reports the average runtime of each step for different scenes (e.g., texture complexity), on a standard laptop (Intel Core i7-1165G7 CPU @ 2.80GHz).
Regarding the Cholesky solver, the most expensive step is solving the normal equations,
more specifically, the Cholesky decomposition of $\mA_{22}$ is the most costly step.
For \shapes{}, whose texture is simple and $\numPixels$ is small, %
EPBA runs fast.
For \boxes{}, whose texture is complicated and $\numPixels$ is large, %
EPBA becomes slower.
Sample sparsity patterns are provided in Appendix \ref{sec:suppl:sparsity}.
The CG solver does not require decomposing or inverting matrices, which is much faster than the Cholesky solver.
For example, for \dynamic{}, CG is 40 times faster than Cholesky.
The runtime of the CG solver is approximately proportional to the size of the system matrix, i.e., the number of valid pixels $\numPixels$, hence it is not as sensitive as the Cholesky solver.

\subsection{Sensitivity and Ablation Analyses}
\label{sec:experim:sensitivity}

Let us characterize the sensitivity of EPBA with respect to its main parameters: 
the contrast threshold $C$ (\cref{sec:sensit:Cth}) and control pose frequency $f$ (\cref{sec:sensit:Posefreq}).
We also show the effect of short-time linearizing the EGM (\cref{sec:sensit:linearization}).
We use the \bicycle{} sequence in the following studies. 
The quadratic loss is adopted as objective, the map size is $1024 \times 512$ px, and the initial rotations are computed using \cmaxw{}.

\subsubsection{Contrast Threshold}
\label{sec:sensit:Cth}
The results of running \EPBA{} with varying values of $C=\{0.05, 0.1, 0.2, 0.5\}$ in the objective function are reported in \cref{tab:sensitivity:contrast},
where $C=0.2$ is the true value used for creating \bicycle{}, and $f=20$~Hz.
Note that the value of the PhE changes as the value of $C$ varies. 
Therefore, for a meaningful comparison, we use the PhE at $C=0.2$ as reference and calculate the equivalent PhE for the other $C$ values.

It turns out that \EPBA{} achieves smaller ARE and PhE in the trials of $C=\{0.1,0.2\}$. 
Nevertheless, those of $C=\{0.05,0.5\}$ still show a good refinement performance (with respect to 1.69$^\circ$ ARE and $6.3 \cdot 10^5$ PhE, in \cref{tab:synth:cg:small:rmse,tab:synth:cg:small:phe}),
which reveals the fact that EPBA is robust to the value of $C$.
This is of great significance to the practicality of \EPBA{} because the $C$ values of real event cameras are difficult to determine and may change greatly during operation (even within the same dataset) \cite{Stoffregen20eccv}.
Future work could look into including $C$ as a variable in the estimation problem, although the effect does not seem to be significant (as per \cref{tab:sensitivity:contrast}) and it would moderately alter the solver.

\begin{table}[t]
\centering
\caption{\label{tab:sensitivity:contrast} 
Sensitivity analysis on the camera's contrast threshold $C$.
Top: absolute rotation error (ARE), in RMSE form.
Bottom: equivalent squared photometric error (PhE).}
\adjustbox{max width=\linewidth}{
\setlength{\tabcolsep}{8pt}
\begin{tabular}{lrrrr}
\toprule 
$C$  & 0.05 & 0.1 & 0.2 & 0.5 \\
\midrule
ARE [$^\circ$] & 0.871 & 0.379 & 0.449 & 0.747 \\
Equiv. PhE  [$\cdot 10^5$] & 1.200 & 1.130 & 1.123 & 1.171 \\
\bottomrule
\end{tabular}
}
\end{table}%
\begin{table}[t]
\centering
\caption{\label{tab:sensitivity:ctrl_pose_freq} 
Sensitivity on the control pose frequency $f$.
}
\adjustbox{max width=\linewidth}{
\setlength{\tabcolsep}{8pt}
\begin{tabular}{lrrrr}
\toprule 
$f$~[Hz]  & 5 & 10 & 20 & 100 \\
\midrule
ARE [$^\circ$] & 0.662 & 0.635 & 0.449 & 0.505 \\
PhE [$\cdot 10^5$] & 1.336 & 1.284 & 1.123 & 1.124 \\
\bottomrule
\end{tabular}
}
\end{table}

\begin{figure}[t]
     \centering
     \begin{subfigure}{0.45\linewidth}
         \centering
         \includegraphics[trim={250px 70px 100px 50px},clip,width=\linewidth]{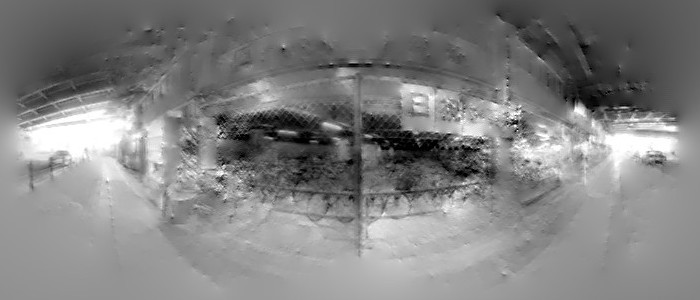}
         \caption{Linearized EGM.}
         \label{fig:sensitivity:LEGM_effect:SMT}
     \end{subfigure}
     \begin{subfigure}{0.45\linewidth}
         \centering
         \includegraphics[trim={250px 70px 100px 50px},clip,width=\linewidth]{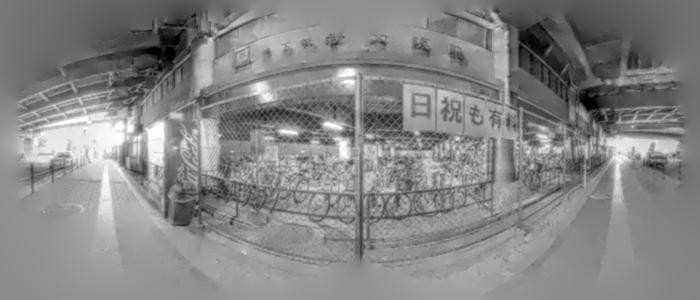}
         \caption{EGM (EPBA).}
         \label{fig:sensitivity:LEGM_effect:EPBA}
     \end{subfigure}
        \caption{%
        Two maps obtained with same camera rotations from \EPBA{}:
        (a) using \esmt{}'s mapping module,
        (b) using EGM.
        Crops from $1024 \times 512$ px panoramic maps.
        \label{fig:sensitivity:LEGM_effect}}
\end{figure}

\def\figWidth{0.95\linewidth}
\begin{figure*}[t]
	\centering
    {\small
    \setlength{\tabcolsep}{1pt}
	\begin{tabular}{
	>{\centering\arraybackslash}m{0.4cm} 
        >{\centering\arraybackslash}m{\figWidth}}
        \rotatebox{90}{\makecell{\atrium{}}}
            &\includegraphics[width=\linewidth]{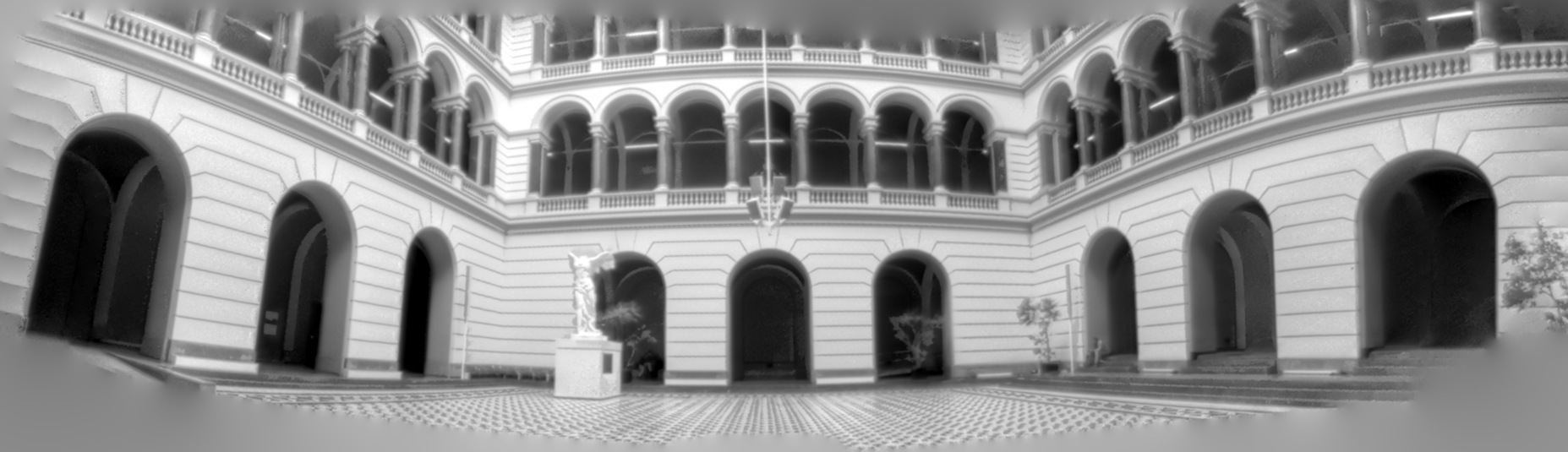}
		\\
        \rotatebox{90}{\makecell{\crossroad{}}}
            &\includegraphics[width=\linewidth]{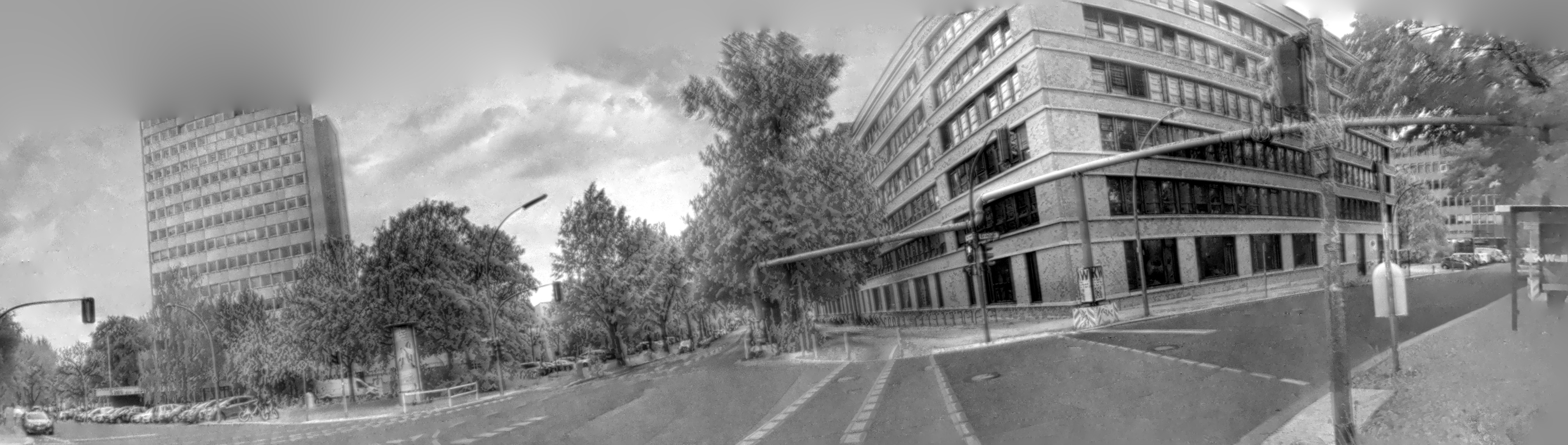}
	\end{tabular}
	}
    \caption{Results of \EPBA{} in the wild (without ground truth).
        The intensity maps are recovered from scratch (zeros).
        (a) \atrium{} is recorded by a Prophesee EVK4 (1 Mpixel camera), where initial rotations are provided by \cmaxw{} angular velocity integration. 
        Crop from a $4096 \times 2048$ px map.
        (b) \crossroad{} is recorded using a DVXplorer (VGA resolution), and initial rotations are provided by IMU angular velocity integration.
        Crop from an $8192 \times 4096$ px map.
        \label{fig:wild_experim}
    }
\end{figure*}

\def\figWidth{0.158\linewidth}
\begin{figure*}[t]
	\centering
    {\small
    \setlength{\tabcolsep}{1pt}
	\begin{tabular}{
	>{\centering\arraybackslash}m{0.3cm} 
	>{\centering\arraybackslash}m{\figWidth} 
	>{\centering\arraybackslash}m{\figWidth}
	>{\centering\arraybackslash}m{\figWidth}
        >{\centering\arraybackslash}m{\figWidth}
        >{\centering\arraybackslash}m{\figWidth}
        >{\centering\arraybackslash}m{\figWidth}}

        \rotatebox{90}{\makecell{Refined map}}
            &\includegraphics[width=\linewidth]{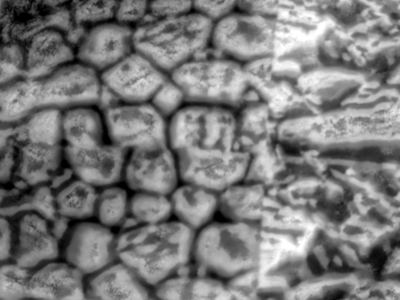}
		&\includegraphics[width=\linewidth]{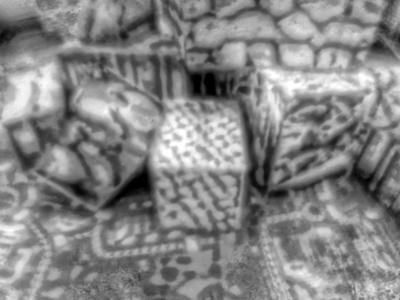}
            &\includegraphics[width=\linewidth]{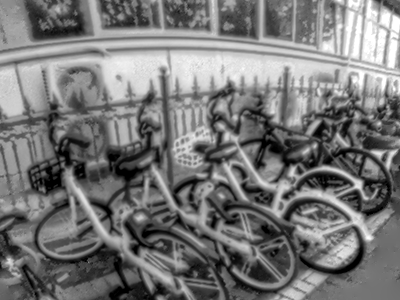}
            &\includegraphics[width=\linewidth]{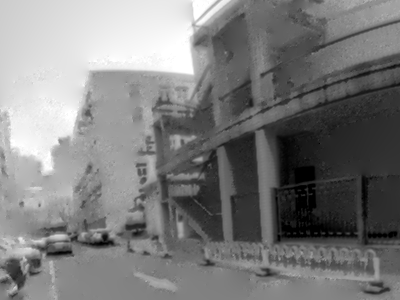}
            &\includegraphics[width=\linewidth]{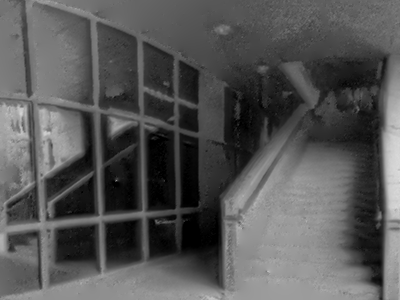}
            &\includegraphics[width=\linewidth]{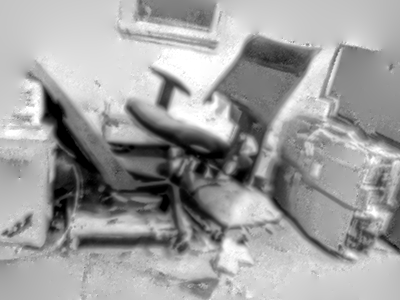}
		\\
  
        \rotatebox{90}{\makecell{Reference}}
		&\includegraphics[width=\linewidth]{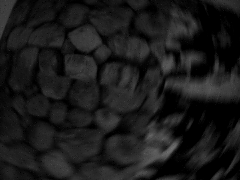}
		&\includegraphics[width=\linewidth]{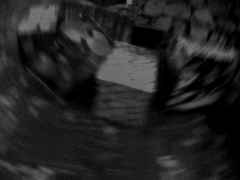}
            &\includegraphics[width=\linewidth]{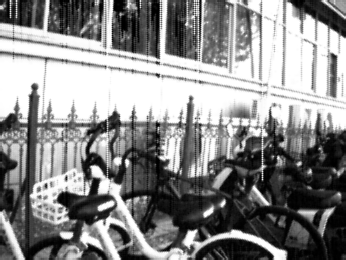}
            &\gframe{\includegraphics[width=\linewidth]{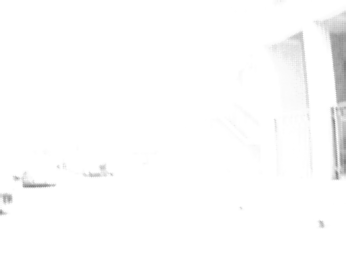}}
            &\includegraphics[width=\linewidth]{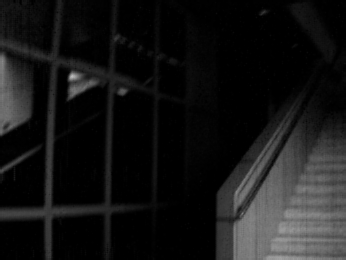}
            &\includegraphics[width=\linewidth]{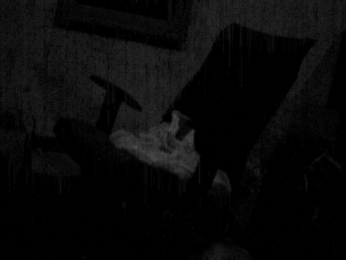}
		\\

        & (a) \poster{}
        & (b) \boxes{}
        & (c) \bicycles{}
        & (d) \building{}
        & (e) \staircase{}
        & (f) \miscellany{}
        \\
	\end{tabular}
	}
    \vspace{-1ex}
    \caption{Results of EPBA in fast-motion, low-light and HDR scenarios. 
    The intensity map is recovered from scratch (zeros), using rotations estimated by CMax-SLAM \cite{Guo24tro}.
    Crops from $2048 \times 1024$~px panoramic maps.
    \label{fig:fast_hdr}}
\end{figure*}

\def\figWidth{0.462\linewidth} \def\zoominWidth{0.165\linewidth}
\begin{figure*}[!ht]
	\centering
    {\small
    \setlength{\tabcolsep}{1pt}
	\begin{tabular}{
	>{\centering\arraybackslash}m{0.4cm} 
        >{\centering\arraybackslash}m{\figWidth}
        >{\centering\arraybackslash}m{\zoominWidth}
        >{\centering\arraybackslash}m{\zoominWidth}
        >{\centering\arraybackslash}m{\zoominWidth}
        }
        \rotatebox{90}{\makecell{1K}}
            &\includegraphics[width=\linewidth]{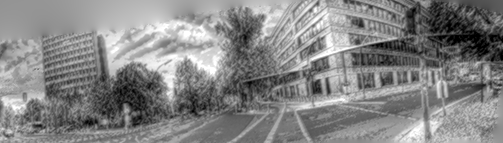}
            &\includegraphics[width=\linewidth]{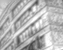}
            &\includegraphics[width=\linewidth]{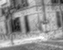}
            &\includegraphics[width=\linewidth]{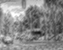}
		\\ [-0.3ex]
        \rotatebox{90}{\makecell{2K}}
            &\includegraphics[width=\linewidth]{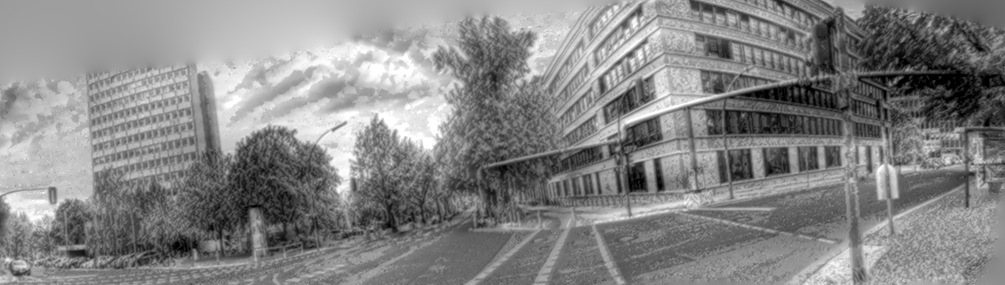}
            &\includegraphics[width=\linewidth]{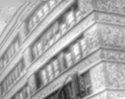}
            &\includegraphics[width=\linewidth]{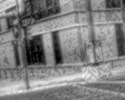}
            &\includegraphics[width=\linewidth]{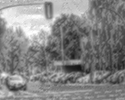}
		\\ [-0.3ex]
        \rotatebox{90}{\makecell{4K}}
            &\includegraphics[width=\linewidth]{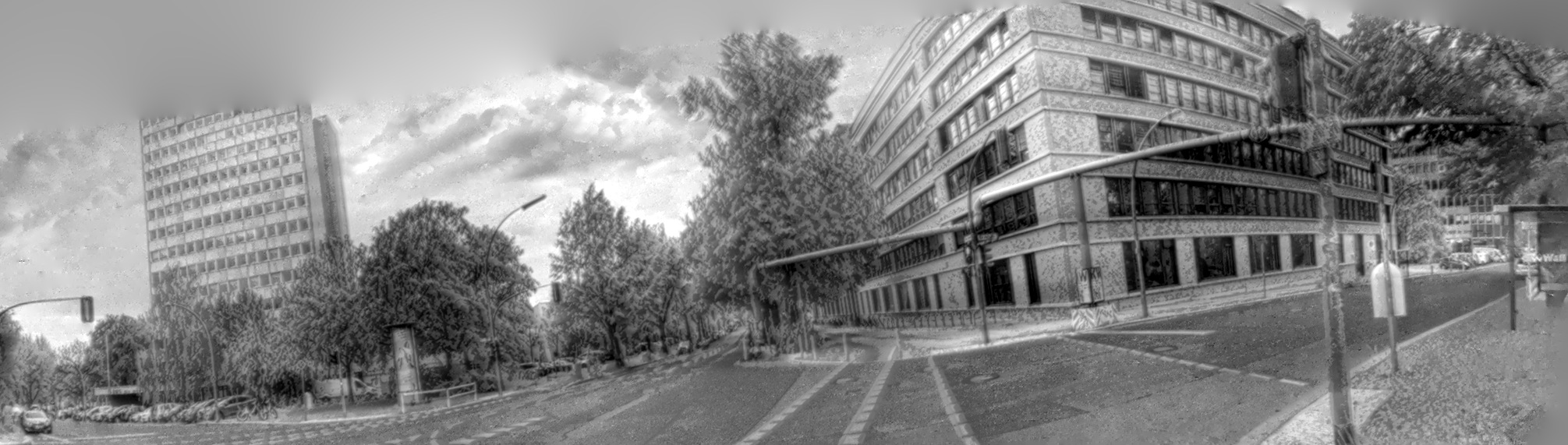}
            &\includegraphics[width=\linewidth]{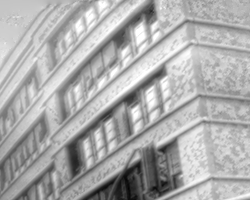}
            &\includegraphics[width=\linewidth]{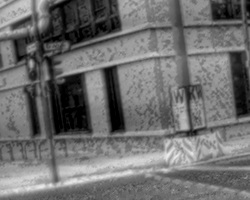}
            &\includegraphics[width=\linewidth]{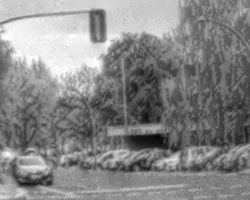}
		\\ [-0.3ex]
        \rotatebox{90}{\makecell{8K}}
            &\includegraphics[width=\linewidth]{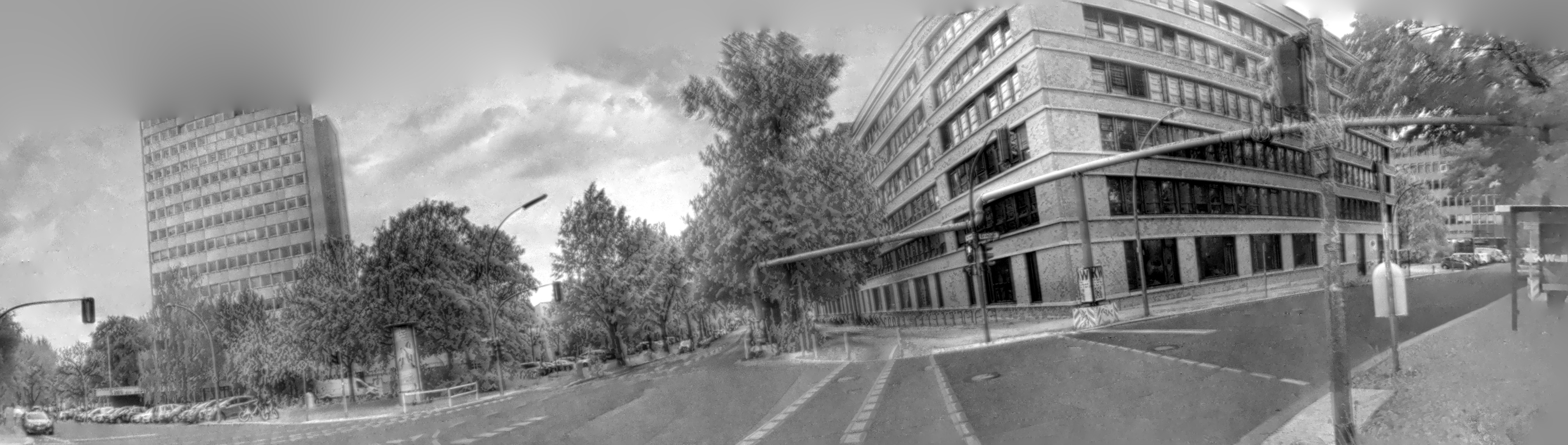}
            &\includegraphics[width=\linewidth]{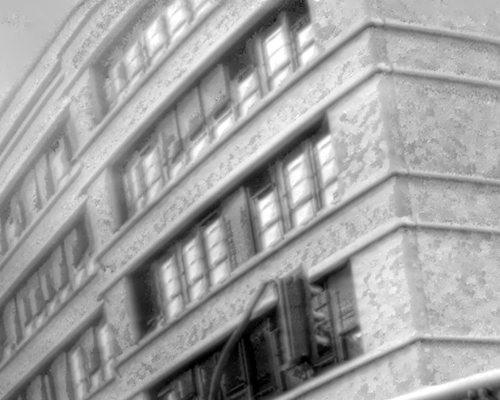}
            &\includegraphics[width=\linewidth]{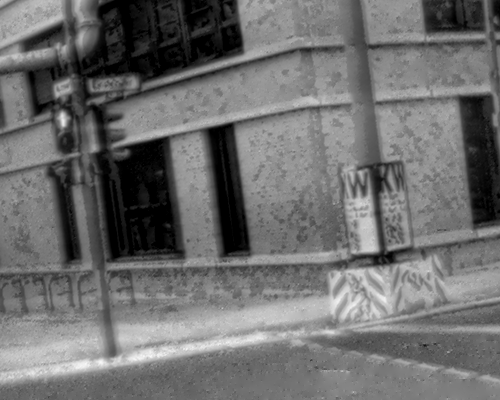}
            &\includegraphics[width=\linewidth]{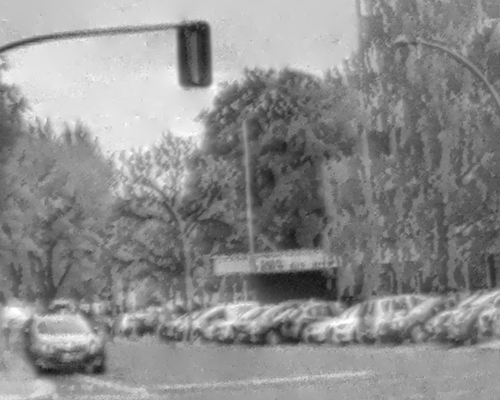}
		\\
	\end{tabular}
	}
    \caption{Results of running \EPBA{} at several map resolutions, from 1K ($1024 \times 512$ px) to 8K ($8192 \times 4096$ px). 
    The intensity maps are recovered from scratch (zeros), and the initial rotations are provided by IMU dead-reckoning.
    The insets have a width of 500 px at the 8K scale.
    \label{fig:super_resolution}
    }
    \vspace{2ex}
    \def\figWidth{0.97\linewidth}
    {\small
    \setlength{\tabcolsep}{1pt}
	\begin{tabular}{
	>{\centering\arraybackslash}m{0.4cm} 
        >{\centering\arraybackslash}m{\figWidth}}
        \rotatebox{90}{\makecell{\bay{}}}
            &\includegraphics[width=\linewidth]{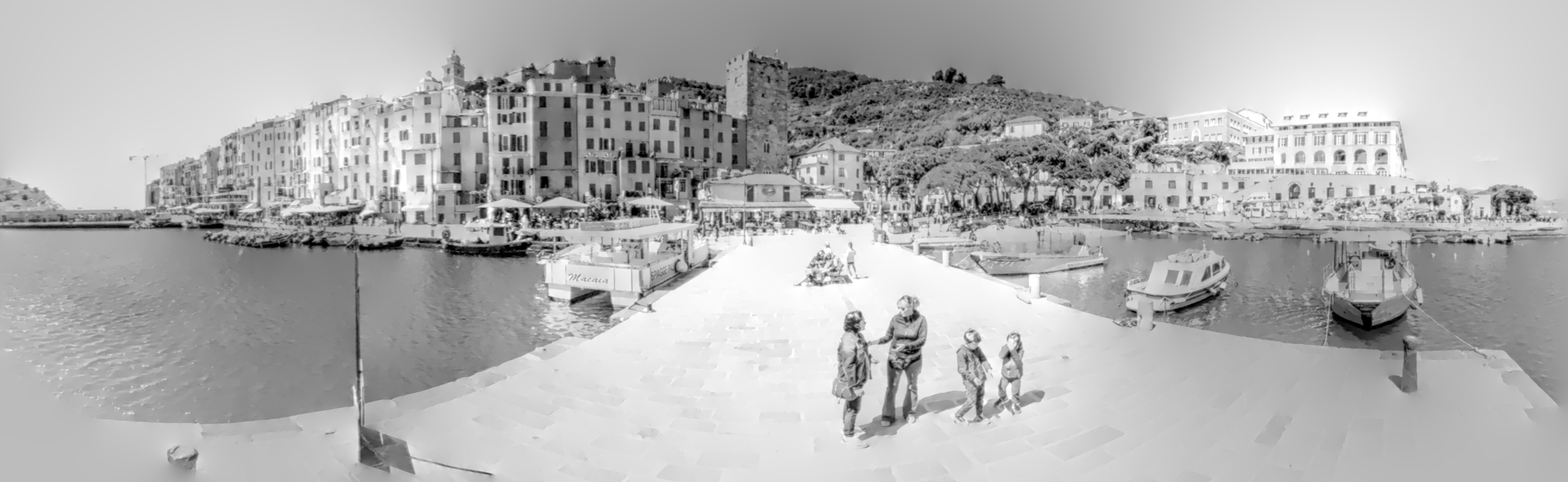}
		\\
        \rotatebox{90}{\makecell{\city{}}}
            &\includegraphics[width=\linewidth]{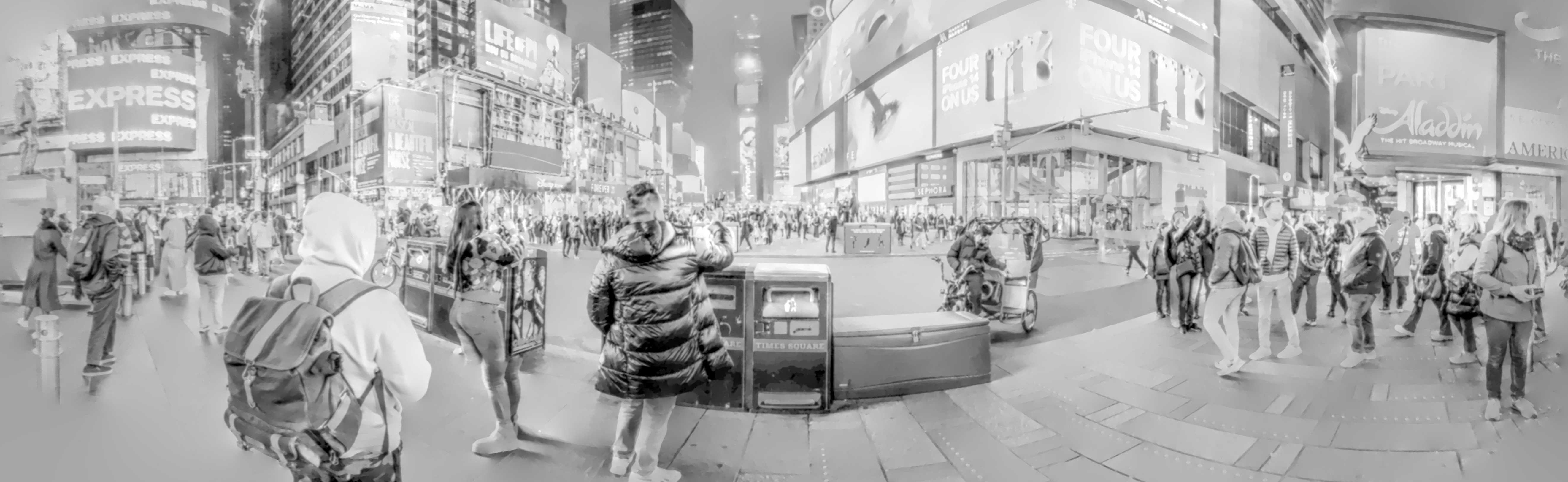}
		\\
	\end{tabular}
	}
    \caption{Results of map-only \EPBA{} on synthetic sequences from the ECRot dataset \cite{Guo24tro}.
        The intensity maps are recovered from scratch (zeros), using GT rotations.
        Crops from $8192 \times 4096$ px maps.
        \label{fig:map_only:synth}
    }
\end{figure*}

\begin{figure}[t]
    \centering
    \includegraphics[width=\linewidth]{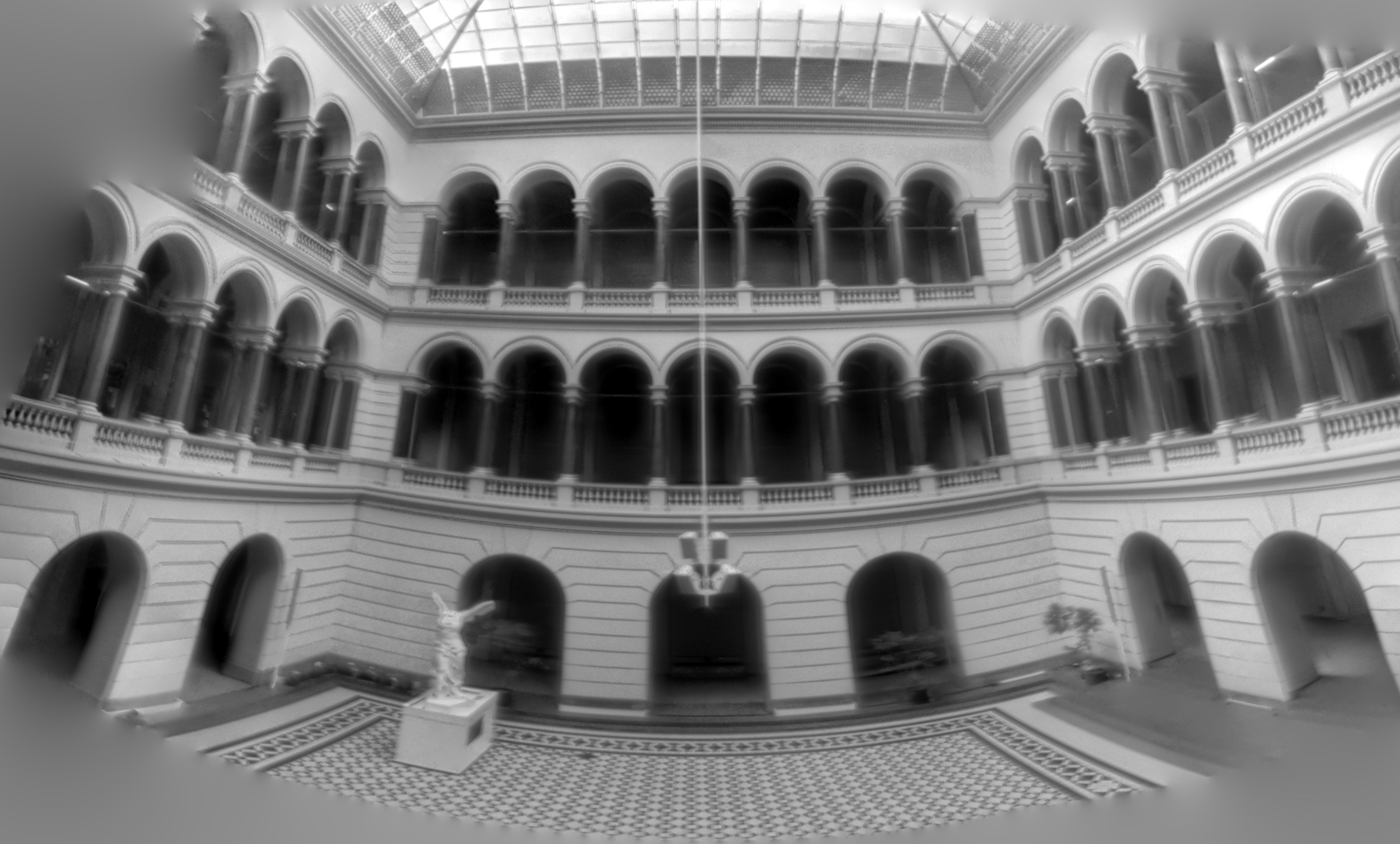}
    \caption{Results of map-only \EPBA{} on data from a Prophesee EVK4 (1 Mpixel) event camera. 
    The intensity map is recovered from scratch (zeros), using rotations estimated by \cmaxslam{} \cite{Guo24tro}.
    Crop from an $8192 \times 4096$ px map.
    \label{fig:map_only:real}
    }
\end{figure}

\subsubsection{Control Pose Frequency}
\label{sec:sensit:Posefreq}
The results of testing EPBA with different control pose frequencies $f=\{5,10,20,100\}$~Hz are presented in \cref{tab:sensitivity:ctrl_pose_freq},
where $C=0.2$ is set to its true value.
Overall, \EPBA{} shows robustness to the choice of $f$.
As $f$ increases from 5 to 20~Hz, both ARE and PhE shrink slightly and reach a minimum at $f=20$~Hz.
When $f$ further grows to $100$~Hz, the errors increase marginally, which implies that no significantly better refinement is achieved by choosing a high frequency (while it incurs in a computational cost).

\subsubsection{Effect of Short-time Linearization of the EGM}
\label{sec:sensit:linearization}
We additionally show the difference between the original EGM and the short-time linearized EGM (LEGM \cite{Gehrig19ijcv}) %
in terms of reconstructing grayscale maps.
Note that LEGM refers to a different linearization from that in \eqref{eq:ErrorsLinearized}; 
it refers to the linearization of \eqref{eq:EGM} by means of the brightness constancy assumption (optical-flow constraint equation) and Taylor's approximation during a short time interval to write an event's brightness increment as the dot product of a brightness gradient and the optical flow \cite[Eq.~(4)]{Gallego20pami}.

We input the refined rotations from EPBA into the mapping module of \esmt{}, %
which is formulated using the LEGM \cite{Kim14bmvc,Kim18phd}, and compare the resulting grayscale panoramic maps, as displayed in \cref{fig:sensitivity:LEGM_effect}.
It is evident that the map reconstructed by the LEGM-based method is blurred and has many artifacts, while the map from EPBA (EGM-based) has a higher quality.%

\subsection{Experiments in the ``Wild'' (without Ground Truth) }
\label{sec:experim:wild}

To demonstrate the applicability of \EPBA{} for panoramic imaging (mosaicing), we recorded data with two of the latest high-resolution event cameras, namely, a DVXplorer (VGA resolution, $640 \times 480$ px) and a Prophesee EVK4 (HD resolution, $1280 \times 720$ px) \cite{Finateu20issccShort}.
Both cameras produced a massive amount of events (millions of events/s) due to their high spatial resolution.
They were hand-held, so there is inevitably translational motion.
We show that \EPBA{} is tolerant to these small translations and is still able to produce sharp panoramas,
thanks to its capability for jointly refining camera motion and map, that is, compensating for small translations to obtain a consistent, sharp intensity map. 

In the case of the DVXplorer, which is equipped with an inertial measurement unit (IMU), 
\EPBA{} is initialized by IMU angular velocity integration (i.e., dead-reckoning).
For the Prophesee EVK4, which has no IMU integrated, initial camera rotations are estimated from event data using \cmaxw{}. 
Regarding map initialization, we use a zero map to initialize \EPBA{} (CG), 
that is, \EPBA{} is capable of recovering the intensity map from scratch.
The results are displayed in \cref{fig:wild_experim}, with maps produced at 8K that reveal subtle textures, such as the statue in \atrium{}, and the cars and bicycles in \crossroad{}.
The maps are inherently HDR, due to the high dynamic range properties of events and the fact that this property is not spoiled by the data processing pipeline.
The results show that \EPBA{} is capable of recovering high-quality panoramas without prior information on the scene, which significantly extends its applicability.

\subsection{Experiments in High-speed and HDR Scenarios}
\label{sec:experim:fast_hdr}

To highlight the advantages of event cameras over traditional cameras 
we also demonstrate EPBA in high-speed, low-light and HDR conditions, whose results are displayed in \cref{fig:fast_hdr}.
The initial rotations are computed using \cmaxslam{}, while the map is reconstructed from scratch.

Regarding fast motion, we test \EPBA{} with the highest-speed segments (50 - 55 s) of \poster{} and \boxes{}, where the frames captured by the DAVIS camera show large motion blur (\cref{fig:fast_hdr}, 2nd row, columns a-b).
In contrast, the reconstructed maps exhibit detailed textures (patterns on the poster, boxes and carpet).
The high-quality reconstruction of EPBA can also be observed despite the fast motion of some of the EFRD sequences (columns c, e and f in \cref{fig:fast_hdr}).

\EPBA{} also works well in challenging illumination conditions, e.g., HDR and low light.
For the former, \EPBA{} reveals the details that are concealed in the DAVIS frames, such as the window reflection in \bicycles{} and almost all objects in \building{} (columns c and d in \cref{fig:fast_hdr}).
For the latter, we can clearly see the objects in the dark in the reconstructed maps, 
such as the window frames in \staircase{} or the chair and the suitcase in \miscellany{}, 
while they are invisible in the DAVIS frames (columns e and f in \cref{fig:fast_hdr}).

In summary, \EPBA{} manages to unlock the high-speed, HDR and low-light characteristics of event cameras. 
This enables scene reconstruction in conditions where traditional cameras perform poorly.

\subsection{Super Resolution}
\label{sec:experim:super_resolution}

To some extent, the resolution of the panoramic map is independent of the event camera resolution.
In principle, we are free to choose the map size as needed, as memory allows.
This enables \EPBA{} to produce super-resolution panoramas that display details that are hidden at low resolutions. 

\Cref{fig:super_resolution} shows the results of running \EPBA{} on \crossroad{} at different resolutions. 
As the map size grows, the panorama becomes smoother and more fine details are recovered,
which are more evident in the zoomed-in insets on the right.
Comparing sensor and map resolutions:
($i$) at the 4K map, a sensor pixel occupies roughly the same area as a map pixel.
($ii$) At the 2K map (coarser resolution), a map pixel receives events from $\approx 4$ sensor pixels.
($iii$) At the 8K map, a sensor pixel covers $\approx 4$ map pixels (fine- or super-resolution);
there are more map pixels in the current field of view than sensor pixels, but the continuous motion of the camera and the high temporal resolution of the data fills in those extra map pixels, thus effectively converting the high temporal resolution of the event camera into the map's high spatial resolution.

It is worth noting that the variable space of the problem changes as the map size varies (the bigger the map, the larger the number of map parameters, and therefore variables in the problem), which may affect the convergence of this NLLS optimization.
Consequently, \EPBA{} may get stuck in a local optima at some very low or high map resolutions.
In other words, a higher map resolution does not always result in better visual quality. 
A coarse-to-fine approach could be adopted to increase robustness against getting trapped in local minima.

\subsection{Map-only Bundle Adjustment}
\label{sec:experim:map_only}

In cases where accurate camera motion is known, the bundle adjustment goal can be simplified to only recovering the intensity map (a similar idea is leveraged in \cite{Zhang22pami} for known optical flow).
\EPBA{} also admits a variation, namely map-only \EPBA{}, for scene panorama reconstruction from given camera rotations.
It is less time- and memory-consuming than full \EPBA{} 
because the derivatives on the camera motion parameters are not needed any more 
and the complexity of forming and solving the normal equations is also reduced. 
Now \eqref{eq:NormalEqsFirstPartitioning} is simplified, becoming 
$\mA_{22} \Delta\bP_{\bbeta}^{\ast} = \bb_{2}$.

We test the map-only \EPBA{} on both synthetic and real-world data.
For the former, we use the GT rotations for initialization.
For the latter, we input events to \cmaxslam{} \cite{Guo24tro} to obtain accurate camera rotations.
The results are presented in \cref{fig:map_only:synth,fig:map_only:real}, where the maps are produced with 8K resolution (super-resolution).
The fine details, e.g., the sea waves and windows in \bay{}, the people and billboards in \city{}, as well as the floor tiles and roof of the \atrium{}, are notably reconstructed.

\ifclearsectionlook\cleardoublepage\fi \section{Limitations}
\label{sec:limitations}

A concern shared by many event-based algorithms is the dependency on \emph{texture}.
High texture produces many events, which can slow down the algorithm. 
On the other end, too little texture can lead to failure of the front-end method, and therefore \EPBA{}'s initialization failure.
If the event camera is equipped with an IMU, then EPBA's initial rotations can be obtained by integration of the gyroscope's output, while the map may be initialized from scratch.
In a broader sense, the texture-dependency concern can be overcome by tuning the camera's $C$ value and/or resampling events.

The assumptions of \emph{brightness constancy} and a static scenario are commonplace among the surveyed event-based rotational SLAM methods.
Hence, events triggered by flickering lights or by independently moving objects may cause inaccuracies or failure if they are a considerable proportion.

\begin{figure}[t]
    \centering
    \includegraphics[width=0.48\linewidth]{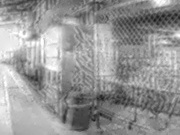}
    \includegraphics[width=0.48\linewidth]{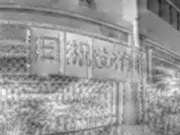}
    \caption{Local minima obtained using \EPBA{} initialized by \cmaxgae{} on the \bicycle{} sequence (red curve in \cref{fig:error_evolution:decreasing}).
    The map contains blurred and double edges (cf. \cref{fig:synth_refine}a).}
    \label{fig:failure}
\end{figure}

\emph{Local convergence} of the LM method is also a limitation: 
\EPBA{} can get stuck in local minima of the very high-dimensional search space if the initialization is not sufficiently good (see \cref{fig:failure}).
This limitation is shared by frame-based BA, and overcoming it and providing guarantees of convergence to the desired solution regularly sparks novel ideas.

As a back-end, \EPBA{} processes the data from a (long) time duration, and event cameras output considerable amount of data 
(VGA and higher spatial resolution event cameras can produce 1Gev/s \cite{Finateu20issccShort}).
These bring considerable pressure to \EPBA{} in terms of \emph{memory and computational resources}.
Despite the tailored LM solver, the method does not run in real time on a laptop. 
\EPBA{} could be optimized for speed at the expense of some accuracy loss; this is a subject for future work.

\begin{figure}
    \centering
    \def\figWidth{0.94\linewidth}
    {\small
    \setlength{\tabcolsep}{1pt}
	\begin{tabular}{
	>{\centering\arraybackslash}m{0.4cm} 
        >{\centering\arraybackslash}m{\figWidth}}
        \rotatebox{90}{\makecell{\sliderfar{}}}
            &\includegraphics[width=\linewidth]{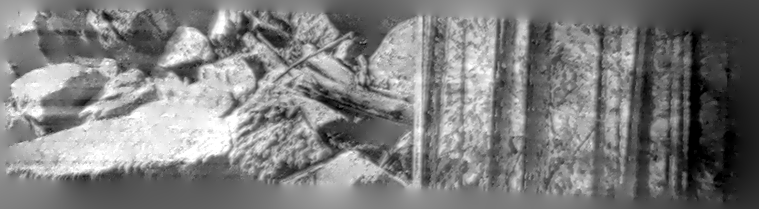}
		\\
        \rotatebox{90}{\makecell{\sliderdepth{}}}
            &\includegraphics[width=\linewidth]{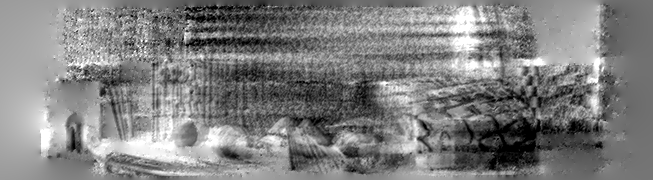}
		\\
	\end{tabular}
	}
    \caption{Reconstructed maps on two translational motion sequences from \cite{Mueggler17ijrr}.
    For \sliderfar{}, initial camera rotations are computed using \cmaxslam{}.
    For \sliderdepth{}, \cmaxslam{} failed, so initial camera rotations are created with a constant angular velocity (of 23$^\circ/s$) around the $y$ axis.
    Crops from $2048 \times 1024$ px panoramic maps.
    \label{fig:trans_test}}
\end{figure}

In \cref{sec:experim:real_data}, we have shown that \EPBA{} is capable of compensating for small camera translations to achieve delicate panoramas.
To further explore the influence of \emph{camera translation}, we test \EPBA{} on translational sequences from \cite{Mueggler17ijrr}: \sliderfar{} and \sliderdepth{}, where the camera translates for 1~m at a constant velocity.
The resulting maps are displayed in \cref{fig:trans_test}.
Note that both the initial rotations from the front-ends and the refined ones from \EPBA{} in this experiment are not the real camera motion.
They are rotations found to compensate for translational components.
As shown in \cref{fig:trans_test},
precise maps are recovered when the front-end (i.e., \cmaxslam{}) estimates rotations that approximately compensate for camera translations (\sliderfar{}).
Conversely, when the front-end fails to provide reasonable initial rotations, \EPBA{} cannot reconstruct the scene (\sliderdepth{}).
This confirms that reasonable initialization is essential for the success of BA algorithms.

The method developed in \cref{sec:method:rotmotion} does not consider \emph{scene depth}. 
However, this assumption was chosen on purpose to explore the capabilities of a first direct method for event-only BA. 
The resulting panoramic intensity maps have better quality than anything previously seen, 
and the camera rotations have also been considerably refined (as much as in \cite{Guo24tro}).
They can inspire the extension of direct event-only BA methods for rigid-body motions. 
Current efforts in modeling scene appearance in rigid-body-motion sequences follow learning-based NeRF or Gaussian splatting paradigms \cite{Klenk23ral,Rudnev23cvpr,kerbl23tog}; however, so far, these rely on the strong assumption of accurately known camera poses (e.g., provided by external motion capture systems).
Future research could look into combining the capabilities of both approaches, to refine appearance, camera trajectories and scene depth with arbitrary camera motions and texture.

\section{Conclusion}
\label{sec:conclusion}

We have introduced the first event-only photometric bundle adjustment approach that jointly refines the motion of a rotating camera and the panoramic intensity map. 
We formulated the bundle adjustment problem from first principles (the event generation model and avoiding linearization errors), which allowed us to exploit the unique space-time characteristics of events (e.g., sparsity and high temporal resolution).
To the best of our knowledge, no prior work on the same task has considered the simultaneous refinement of camera orientations and scene map, without converting events into frames.
A second-order solver has been tailored for this problem, to make it tractable on a standard laptop, for which we have adopted a cumulative way to form normal equations.
On both synthetic and real-world data, the proposed method achieves great improvements in terms of both the camera rotations and map quality from four front-end estimators.
We have shown the applicability of the method to produce high quality panoramas at various resolutions from data recorded by both old and modern event cameras (even 1 Mpixel resolution) despite small translations, e.g. due to hand-held motion.
We release the code and hope that our work helps bring maturity to panoramic image reconstruction and event-based SLAM, among the potential applications of bundle adjustment.

\ifarxiv
\appendices
\section*{Supplementary Material}

\section{Quantitative Results with the Cholesky Solver}
\label{sec:suppl:cholesky}
For completeness, \cref{tab:synth:chol:small:rmse,tab:real:chol:small:rmse} report the rotation and photometric errors using the Cholesky solver instead of the CG solver (\cref{tab:synth:cg:small:rmse,tab:synth:cg:small:phe,tab:real:cg:small:rmse,tab:real:cg:small:phe}). 
In general, the accuracy numbers obtained with the direct linear solver (Cholesky) are slightly better than with the iterative solver (CG), 
but the former solver does not scale as well with the problem size as the latter (see \cref{tab:runtime}).
Therefore, we decided to report the CG solver in the main manuscript and the Cholesky solver in this Appendix.

\begin{table*}[ht]
\centering
\caption{\label{tab:synth:chol:small:rmse} 
Absolute rotation RMSE [deg] (ARE)
and squared photometric error [$\times 10^6$] (PhE)
on synthetic sequences \cite{Guo24tro} (Cholesky solver, 1024 $\times$ 512 px map).}
\adjustbox{max width=\linewidth}{
\begin{tabular}{ll*{14}{S[table-format=1.3,table-number-alignment=center]}}
\toprule
&& \multicolumn{4}{c}{\esmt{}} &\emptycol& \multicolumn{4}{c}{\cmaxgae{}} &\emptycol& \multicolumn{4}{c}{\cmaxw{}}\\
\cmidrule(l{2mm}r{2mm}){3-6}
\cmidrule(l{2mm}r{2mm}){8-11}
\cmidrule(l{2mm}r{2mm}){13-16}
&Sequence & \text{before} & \text{Quad} & \text{Huber} & \text{Cauchy} 
&\emptycol& \text{before} & \text{Quad} & \text{Huber} & \text{Cauchy} 
&\emptycol& \text{before} & \text{Quad} & \text{Huber} & \text{Cauchy} \\
\midrule
\multirow{6}{*}{\begin{turn}{90}
ARE
\end{turn}}
&playroom & 5.861 & 5.485 & 5.05 & 5.32 & \emptycol & 4.628 & 3.428 & 1.937 & 2.07 & \emptycol & 3.223 & 1.089 & 0.594 & 0.549\\
&bicycle & 1.466 & 0.56 & 0.558 & 0.558 & \emptycol & 1.651 & 1.298 & 1.141 & 1.119 & \emptycol & 1.69 & 0.449 & 0.178 & 0.169\\
&city & 1.692 & 1.389 & 0.979 & 0.542 & \emptycol & \novalue & \NA{} & \NA{} & \NA{} & \emptycol & 1.532 & 0.515 & 0.152 & 0.155\\
&street & 3.441 & 2.818 & 2.631 & 2.61 & \emptycol & \novalue & \NA{} & \NA{} & \NA{} & \emptycol & 0.965 & 0.529 & 0.153 & 0.155\\
&town & 4.322 & 4.096 & 4.211 & 4.235 & \emptycol & 4.656 & 4.318 & 3.724 & 3.766 & \emptycol & 1.905 & 0.767 & 0.182 & 0.186\\
&bay & 2.5 & 2.418 & 3.606 & 3.445 & \emptycol & \novalue & \NA{} & \NA{} & \NA{} & \emptycol & 1.797 & 1.411 & 0.877 & 0.834\\
\midrule
\multirow{6}{*}{\begin{turn}{90}
PhE
\end{turn}}
&playroom & 0.683 & 0.12 & 0.176 & 0.231 & \emptycol & 0.675 & 0.112 & 0.13 & 0.153 & \emptycol & 0.913 & 0.088 & 0.102 & 0.116\\
&bicycle & 0.458 & 0.113 & 0.121 & 0.118 & \emptycol & 0.68 & 0.179 & 0.197 & 0.198 & \emptycol & 0.632 & 0.112 & 0.123 & 0.12\\
&city & 0.963 & 0.421 & 0.453 & 0.434 & \emptycol & \novalue & \NA{} & \NA{} & \NA{} & \emptycol & 2.121 & 0.388 & 0.428 & 0.43\\
&street & 0.782 & 0.387 & 0.464 & 0.474 & \emptycol & \novalue & \NA{} & \NA{} & \NA{} & \emptycol & 1.571 & 0.294 & 0.321 & 0.323\\
&town & 0.685 & 0.398 & 0.448 & 0.454 & \emptycol & 0.806 & 0.369 & 0.388 & 0.395 & \emptycol & 1.406 & 0.305 & 0.337 & 0.339\\
&bay & 0.698 & 0.46 & 0.484 & 0.495 & \emptycol & \novalue & \NA{} & \NA{} & \NA{} & \emptycol & 1.764 & 0.401 & 0.449 & 0.464\\
\bottomrule
\end{tabular}
}
\end{table*}

\begin{table*}[ht]
\centering
\caption{\label{tab:real:chol:small:rmse} 
Absolute rotation RMSE [deg] (ARE)
and squared photometric error [$\times 10^6$] (PhE)
on real sequences \cite{Mueggler17ijrr} 
(Cholesky solver, 1024 $\times$ 512 px map).}
\adjustbox{max width=\linewidth}{
\begin{tabular}{ll*{14}{S[table-format=1.3,table-number-alignment=center]}}
\toprule
&& \multicolumn{4}{c}{\rtpt{}} &\emptycol& \multicolumn{4}{c}{\cmaxgae{}} &\emptycol& \multicolumn{4}{c}{\cmaxw{}}\\
\cmidrule(l{2mm}r{2mm}){3-6}
\cmidrule(l{2mm}r{2mm}){8-11}
\cmidrule(l{2mm}r{2mm}){13-16}
&Sequence & \text{before} & \text{Quad} & \text{Huber} & \text{Cauchy} 
&\emptycol& \text{before} & \text{Quad} & \text{Huber} & \text{Cauchy} 
&\emptycol& \text{before} & \text{Quad} & \text{Huber} & \text{Cauchy} \\
\midrule
\multirow{4}{*}{\begin{turn}{90}
ARE
\end{turn}}
&shapes & 2.187 & 2.991 & 2.933 & 2.946 & \emptycol & 2.512 & 3.111 & 2.944 & 2.933 & \emptycol & 4.111 & 3.021 & 2.971 & 2.974\\
&poster & 3.802 & 4.06 & 4.064 & 4.066 & \emptycol & 3.625 & 4.313 & 4.376 & 4.358 & \emptycol & 4.072 & 4.119 & 4.124 & 4.124\\
&boxes & 1.743 & 2.141 & 2.506 & 3.003 & \emptycol & 2.018 & 1.999 & 2.981 & 3.109 & \emptycol & 3.224 & 2.784 & 2.734 & 2.736\\
&dynamic & 2 & 2.816 & 2.86 & 2.748 & \emptycol & 1.698 & 2.582 & 2.696 & 2.71 & \emptycol & 3.126 & 2.97 & 2.853 & 2.849\\
\midrule
\multirow{4}{*}{\begin{turn}{90}
PhE
\end{turn}}
&shapes & 0.723 & 0.192 & 0.208 & 0.208 & \emptycol & 0.75 & 0.267 & 0.228 & 0.238 & \emptycol & 0.553 & 0.192 & 0.208 & 0.208\\
&poster & 5.535 & 1.954 & 2.083 & 2.077 & \emptycol & 5.782 & 3.124 & 3.661 & 3.487 & \emptycol & 4.345 & 1.954 & 2.078 & 2.073\\
&boxes & 4.792 & 3 & 3.232 & 3.242 & \emptycol & 4.667 & 3.263 & 3.389 & 3.553 & \emptycol & 3.736 & 1.619 & 1.701 & 1.694\\
&dynamic & 3.474 & 2.313 & 2.548 & 2.685 & \emptycol & 3.539 & 2.497 & 2.801 & 2.858 & \emptycol & 2.914 & 1.646 & 1.8 & 1.814\\
\bottomrule
\end{tabular}
}
\end{table*}

\section{Camera Translation in the ECD Sequences}
\label{sec:suppl:camera_trans}
In \cref{sec:experim}, we mentioned that the four sequences from the ECD dataset \cite{Mueggler17ijrr} were recorded by a hand-held event camera, so the camera motion inevitably contains translations,
which affects all involved front-end methods as well as our BA approach.
\Cref{fig:suppl:cam_trans} displays the translational component of the GT poses provided by the mocap system.
It shows that the magnitude of the translational motion grows, as time progresses and the speed of the motion increases.
We use the first part of the sequences, where the translational motion is still small (about less than 10 cm) for the desk/room-sized scenes.

\begin{figure}[h]
    \centering
    \begin{subfigure}[b]{0.46\linewidth}
         \centering
         \includegraphics[width=\linewidth]{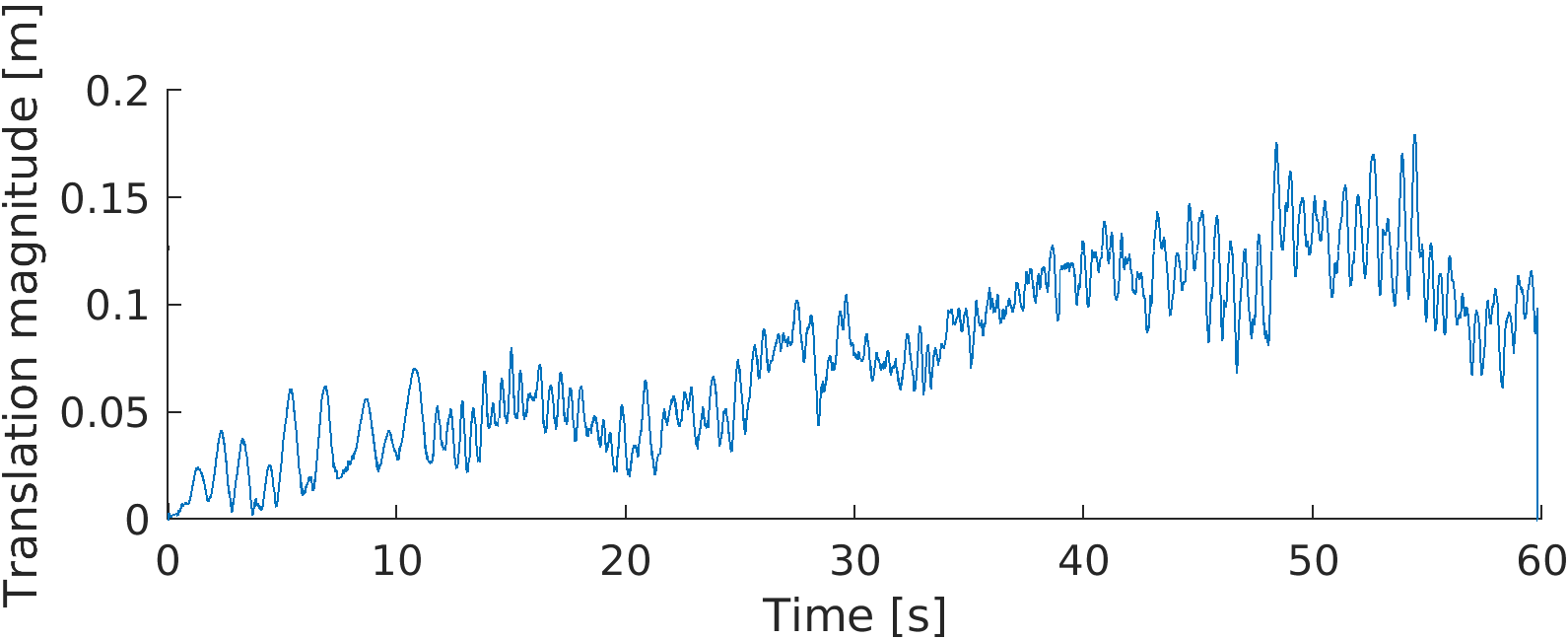}
         \caption{\shapes{}}
         \label{fig:suppl:cam_trans:shapes}
     \end{subfigure}\;\;\;\;
     \begin{subfigure}[b]{0.46\linewidth}
         \centering
         \includegraphics[width=\linewidth]{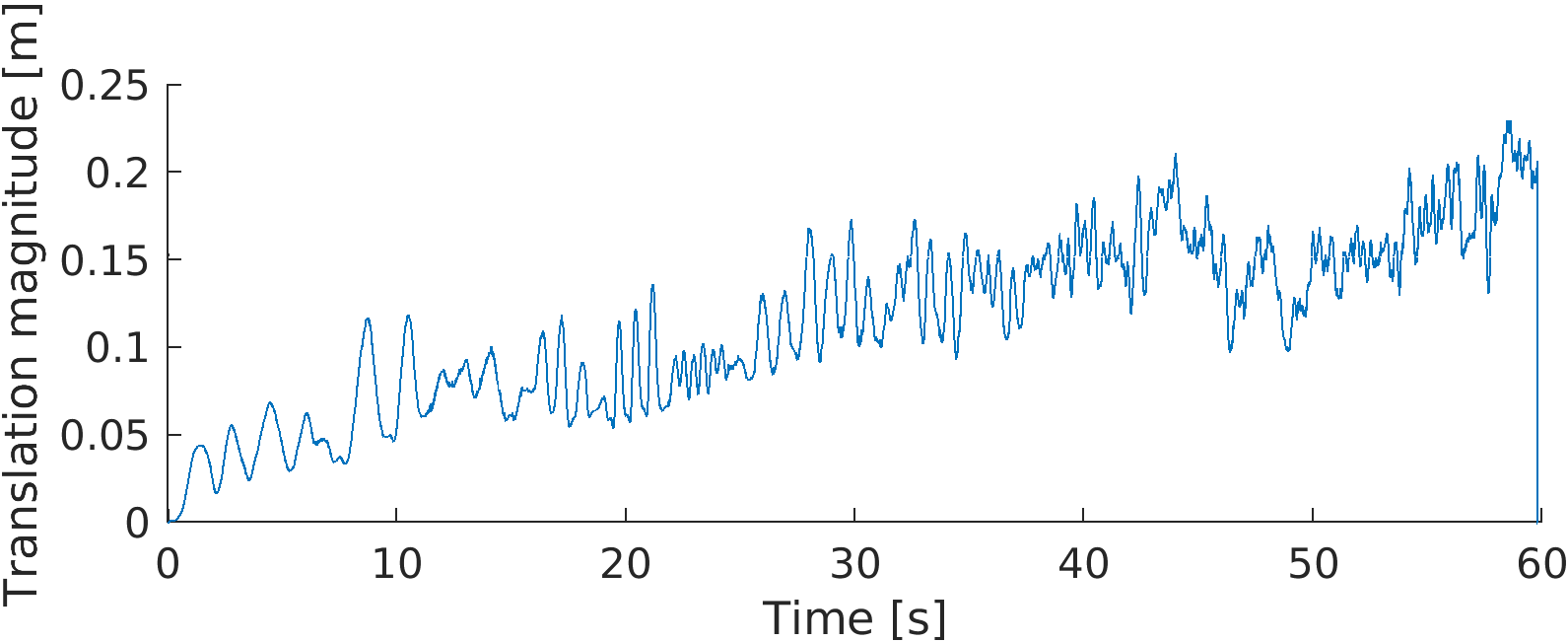}
         \caption{\poster{}}
         \label{fig:suppl:cam_trans:poster}
     \end{subfigure}
     \begin{subfigure}[b]{0.46\linewidth}
         \centering
         \includegraphics[width=\linewidth]{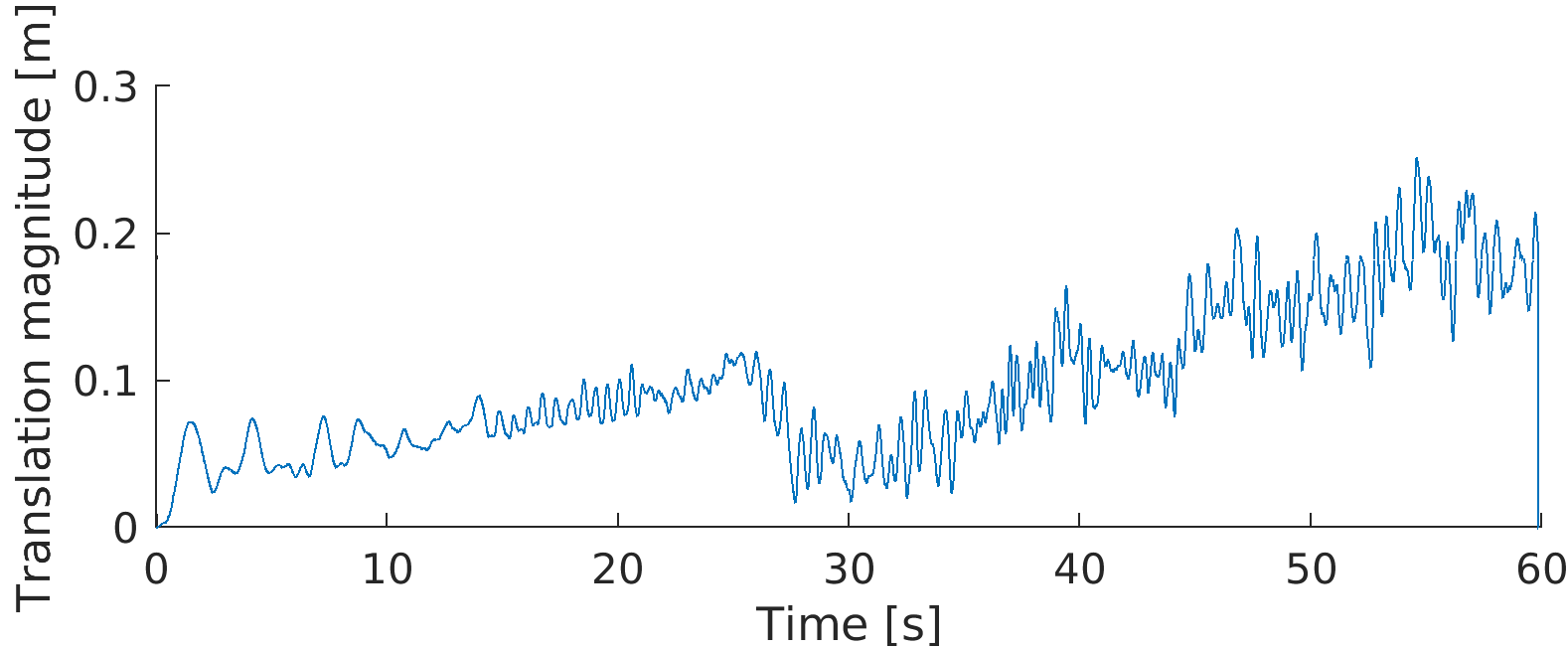}
         \caption{\boxes{}}
         \label{fig:suppl:cam_trans:boxes}
    \end{subfigure}\;\;\;\;
    \begin{subfigure}[b]{0.46\linewidth}
         \centering
         \includegraphics[width=\linewidth]{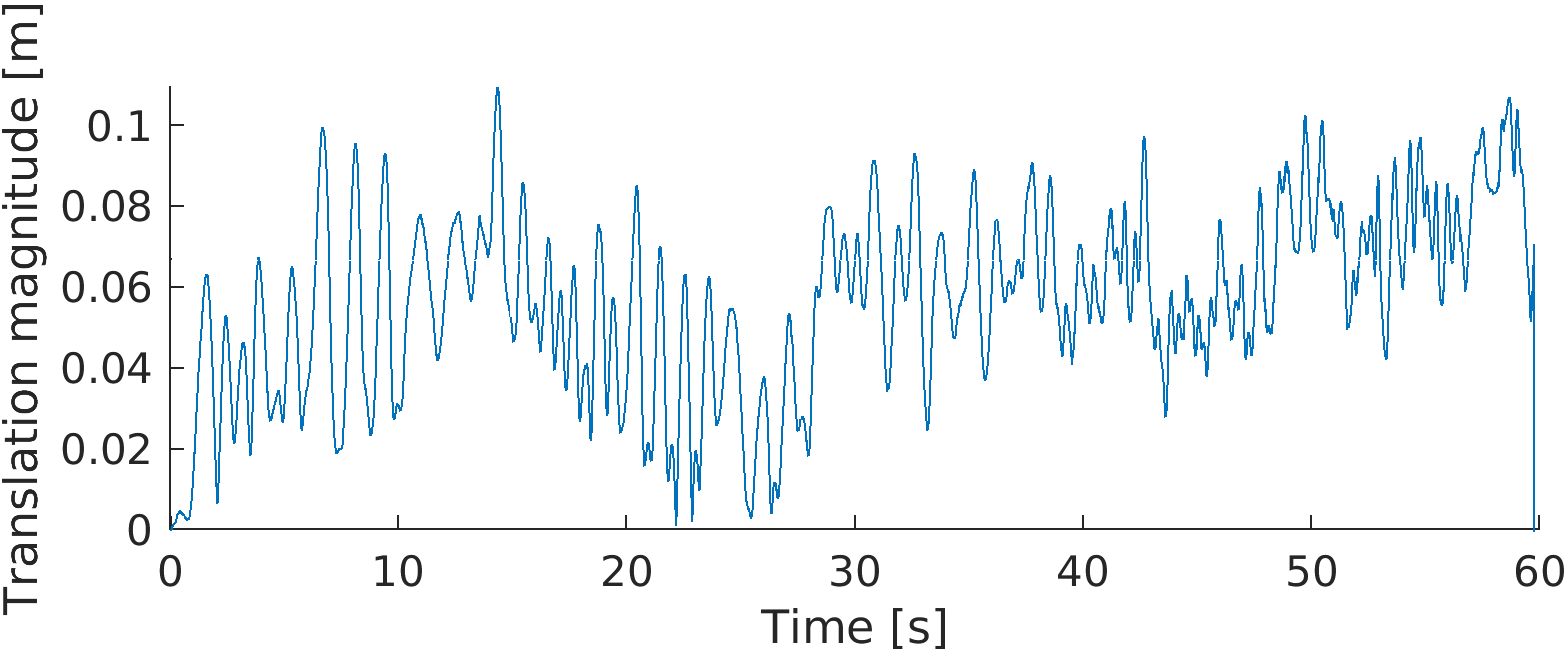}
         \caption{\dynamic{}}
         \label{fig:suppl:cam_trans:dynamic}
    \end{subfigure}
    \vspace{-1ex}
    \caption{From the motion capture system: ground truth camera translation magnitude of the four ECD sequences \cite{Mueggler17ijrr}.
    }
    \label{fig:suppl:cam_trans}
\end{figure}

\section{Sparsity Patterns}
\label{sec:suppl:sparsity}
\Cref{fig:sparsity:ecd} displays the sparsity patterns on the four ECD sequences, which support the results of the runtime evaluation with the Cholesky solver in \cref{tab:runtime}.
\EPBA{} solves the normal equations fastest on \shapes{} because the size of the $\mA_{22}$ matrix is the smallest.
\poster{} is the second fastest also because it results in a smaller $\mA_{22}$ matrix (given by $\numPixels$) than the other two.
\boxes{} and \dynamic{} have a similar number of valid pixels $\numPixels$, but the $\mA_{22}$ matrix of \boxes{} has a smaller bandwidth than that of \dynamic{} (better sparsity pattern).
Therefore, it takes a shorter time for \boxes{} to solve the normal equations than \dynamic{},
even though the $\mA_{22}$ matrix of the latter has a smaller number of nonzero elements (nz).
This verifies our statement in \cref{sec:experim:runtime}: the time consumption of the factorization of $\mA_{22}$ depends in a complicated way on the matrix size, number of nonzero elements and sparsity pattern \cite{Boyd04book}.

\def\figWidth{0.45\linewidth}
\begin{figure}[t]
    \centering
    \begin{subfigure}[b]{\figWidth}
         \centering
         {\includegraphics[trim={2cm, 0, 2cm, 0},clip,width=\linewidth]{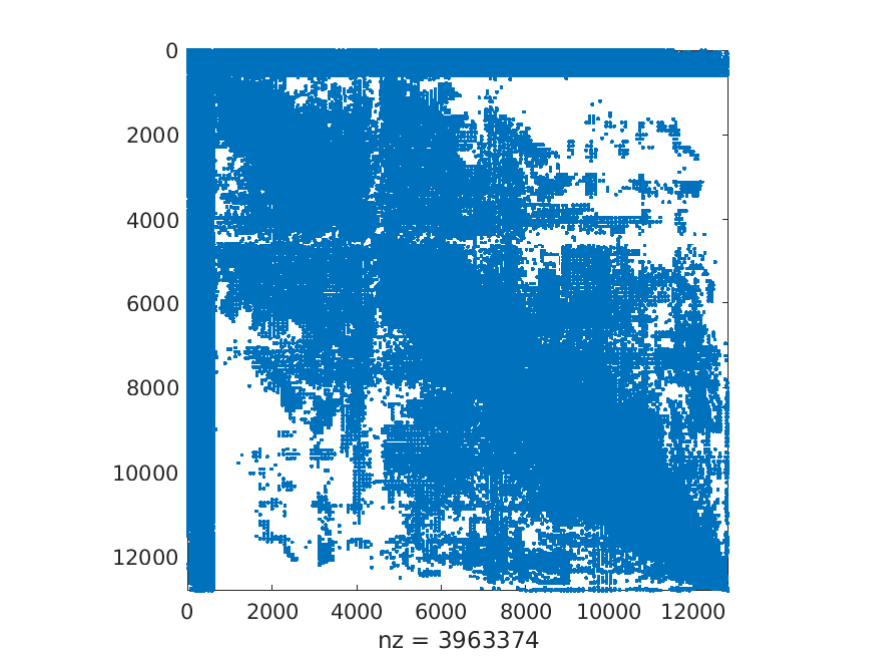}}
         \caption{\shapes{}, $\numPixels=14115$}
         \label{fig:sparsity:shapes}
     \end{subfigure}
     \begin{subfigure}[b]{\figWidth}
         \centering
         {\includegraphics[trim={2cm, 0.2cm, 2cm, 0},clip,width=\linewidth]{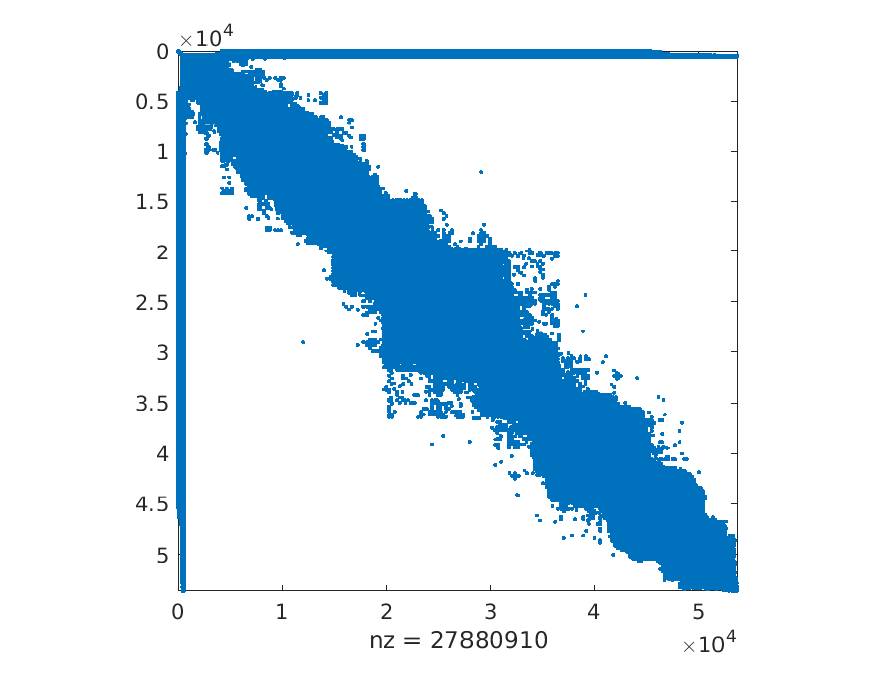}}
         \caption{\poster{}, $\numPixels=53786$}
         \label{fig:sparsity:poster}
     \end{subfigure}
     \begin{subfigure}[b]{\figWidth}
         \centering
         {\includegraphics[trim={2cm, 0, 2cm, 0},clip,width=\linewidth]{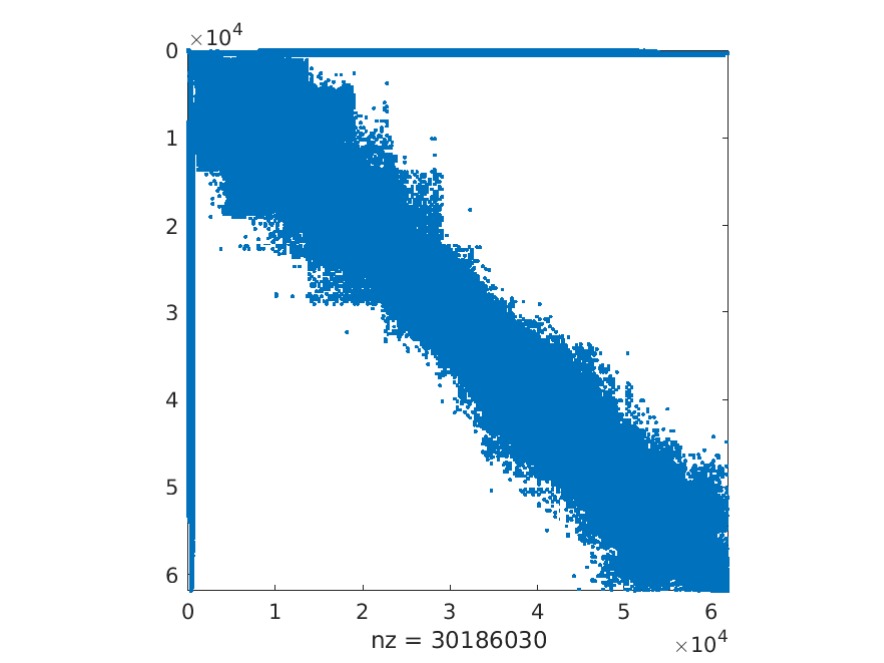}}
         \caption{\boxes{}, $\numPixels=62832$}
         \label{fig:sparsity:boxes}
    \end{subfigure}
    \begin{subfigure}[b]{\figWidth}
         \centering
         {\includegraphics[trim={2cm, 0, 2cm, 0},clip,width=\linewidth]{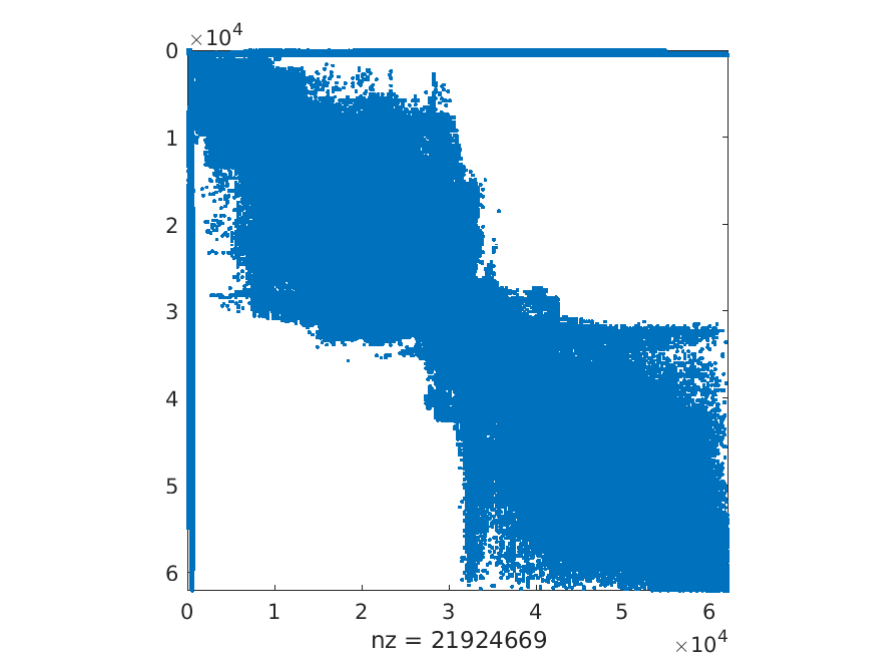}}
         \caption{\dynamic{}, $\numPixels=62760$}
         \label{fig:sparsity:dynamic}
    \end{subfigure}
    \caption{Sparsity pattern of matrix $\mA \in \Real^{(3\numPoses+\numPixels) \times (3\numPoses+\numPixels)}$ in the normal equations during the first iteration of EPBA, for each of the four ECD \cite{Mueggler17ijrr} sequences. 
    Note the difference in size ($\numPixels$), number of non-zero entries (nz) and bandwidth.
    }
    \label{fig:sparsity:ecd}
\end{figure}

\section{Linearization of Photometric Error Terms}
\label{sec:suppl:linear}

In the chain of transformations \eqref{eq:chainoftransformations}, there is a geometric part (how to transform the coordinates of a sensor pixel to the coordinates of a map point) and a photometric part (reading out the intensities at the map point(s)).
The photometric errors are the compositions of both parts (geometric and photometric), thus applying the chain rule, the derivative of the photometric error is the product of the derivatives of each part (product of Jacobian matrices).
Let us first linearize the geometric part (\cref{sec:LinearizeGeometricPart}), then the photometric part (\cref{sec:LinearizePhotometricPart}), 
and combine them to show how to obtain the linerization of the per-event photometric error~\eqref{eq:LinErrorOrigEGM}.

\subsection{Linearization of the geometric part}
\label{sec:LinearizeGeometricPart}

First we can compute how the map point changes as the camera rotation changes.
Let us substitute the rotation perturbation \eqref{eq:poseandmapperturbation} in the formula for the warped event,
approximate the matrix exponential using the first two terms in the series expansion, 
and use first order Taylor's expansion and the properties of the cross product:
\begin{align}
\bp(t) & =\pi(\Rot(t)\Kint^{-1}\bx^{h})\nonumber \\
 & \stackrel{\eqref{eq:poseandmapperturbation}}{=} \pi\bigl(\exp(\delta\boldsymbol{\varphi}^{\wedge})\Rot^{\text{op}}(t)\Kint^{-1}\bx^{h}\bigr)\nonumber \\
 & \approx\pi((\boldsymbol{1}+\delta\boldsymbol{\varphi}^{\wedge})\underbrace{\Rot^{\text{op}}(t)\Kint^{-1}\bx^{h}}_{\doteq\bz_{\text{op}}(t)})\nonumber \\
 & =\pi(\bz_{\text{op}}+\delta\boldsymbol{\varphi}^{\wedge}\bz_{\text{op}})\nonumber \\
 & =\pi(\bz_{\text{op}}-\bz_{\text{op}}^{\wedge}\delta\boldsymbol{\varphi})\label{eq:MapPointLinearizePerturbtAux} \\
 & \stackrel{\text{Taylor}}{\approx}\pi(\bz_{\text{op}})-\underbrace{\left.\prtl{\pi}{\bz}\right|_{\bz_{\text{op}}}\bz_{\text{op}}^{\wedge}}_{\doteq\mE_{\text{op}}}\delta\boldsymbol{\varphi}\nonumber \\
 & =\pi(\bz_{\text{op}})-\mE_{\text{op}}\delta\boldsymbol{\varphi}. 
 \label{eq:MapPointLinearizePerturbt}
\end{align}

Next, we relate $\delta\boldsymbol{\varphi}$ (at time $t$) to the perturbations in the control poses (i.e., rotations) $\Delta\bP_{\balpha}$ \eqref{eq:Perturb}--\eqref{eq:PerturbVecPartitioning}.
To this end, we use the results in \cite{Barfoot15book}, around Faulhaber\textquoteright s formula. 
Using linear interpolation of the control poses, with $u(t)\in[0,1]$,
\[
\Rot(t)=\exp\left(u(t)\log\left(\Rot_{i+1}\Rot_{i}^{-1}\right)\right)\Rot_{i}
\]
Starting from \cite[(8.140)]{Barfoot15book},
\[
\exp(\delta\boldsymbol{\varphi}^{\wedge})\Rot^{\text{op}}(t)=\exp\left(u(t)\log\left(\Rot_{i+1}\Rot_{i}^{-1}\right)\right)\Rot_{i}
\]
we arrive at
\begin{equation}
\delta\boldsymbol{\varphi}=(\boldsymbol{1}-\boldsymbol{A}(u(t),\boldsymbol{\phi}))\delta\boldsymbol{\phi}_{1}+\boldsymbol{A}(u(t),\boldsymbol{\phi})\delta\boldsymbol{\phi}_{2},
\label{eq:BarfootdeltaPerturbRotAtt}
\end{equation}
where $\exp(\boldsymbol{\phi}^{\wedge})=\Rot_{i+1}\Rot_{i}^{-1}$
and $\boldsymbol{A}(u,\boldsymbol{\phi})=u\boldsymbol{J}(u\boldsymbol{\phi})\boldsymbol{J}^{-1}(\boldsymbol{\phi}).$
Formula \eqref{eq:BarfootdeltaPerturbRotAtt} mirrors the usual linear interpolation scheme. 
When $\boldsymbol{\phi}$ is small, then $\boldsymbol{A}(u,\boldsymbol{\phi})\approx u\boldsymbol{1}$.

Substituting \eqref{eq:BarfootdeltaPerturbRotAtt} into \eqref{eq:MapPointLinearizePerturbt},
we obtain the formula for how the map point changes as the control poses are perturbed:
\begin{align}
\bp(t) & \approx\bp_{\text{op}}(t)-\left(\mE_{\text{op}}(\boldsymbol{1}-\boldsymbol{A}(u(t),\boldsymbol{\Delta\phi}_{i,i+1}))\right)\delta\boldsymbol{\phi}_{i}\\
&\qquad\qquad -\left(\mE_{\text{op}}\boldsymbol{A}(u(t),\boldsymbol{\Delta\phi}_{i,i+1})\right)\delta\boldsymbol{\phi}_{i+1},
\end{align}
where 
\[
\boldsymbol{A}(u(t),\boldsymbol{\Delta\phi}_{i,i+1})=u(t)\boldsymbol{J}(u(t)\boldsymbol{\Delta\phi}_{i,i+1})\boldsymbol{J}^{-1}(\boldsymbol{\Delta\phi}_{i,i+1}).
\]
and the incremental angles $\boldsymbol{\Delta\phi}_{i,i+1}$ are defined by
$\exp(\boldsymbol{\Delta\phi}_{i,i+1}^{\wedge})=\Rot_{i+1}\Rot_{i}^{-1}$.

\subsection{Linearization of the photometric part}
\label{sec:LinearizePhotometricPart}

Stemming from \eqref{eq:chainoftransformations}--\eqref{eq:mappointwarp}, let $\bp_{\text{op}}(t)$ be the ``operating point'' of the map point at time~$t$,
\[
\bp_{\text{op}}(t)\doteq\pi(\Rot_{\text{op}}(t)\Kint^{-1}\bx^{h}).
\]
Then, perturbing both the map and the rotation,
\begin{align}
M(\bp(t)) 
 & \stackrel{\eqref{eq:chainoftransformations}}{=} M(\pi(\Rot(t)\Kint^{-1}\bx^{h})) \nonumber\\
 & \stackrel{\eqref{eq:poseandmapperturbation}}{=} (M_{\text{op}}+\Delta M)\left(\pi(\exp(\delta\boldsymbol{\varphi}^{\wedge})\Rot_{\text{op}}(t)\Kint^{-1}\bx^{h})\right) \nonumber\\
 & \stackrel{\eqref{eq:MapPointLinearizePerturbtAux}}{\approx} M_{\text{op}}\left(\pi(\bz_{\text{op}}-\bz_{\text{op}}^{\wedge}\delta\boldsymbol{\varphi})\right)+\Delta M\left(\pi(\bz_{\text{op}}-\bz_{\text{op}}^{\wedge}\delta\boldsymbol{\varphi})\right) \nonumber\\
 & \stackrel{\text{Taylor}}{\approx} (M_{\text{op}}\circ\pi)(\bz_{\text{op}}) + \underbrace{\left.\prtl{(M_{\text{op}}\circ\pi)}{\bz}\right|_{\bz_{\text{op}}}(-\bz_{\text{op}}^{\wedge}\delta\boldsymbol{\varphi})}_{\text{linear in }\delta\boldsymbol{\varphi}} \nonumber\\
 &\qquad +\underbrace{(\Delta M\circ\pi)(\bz_{\text{op}})}_{\text{linear in }\Delta M} +\underbrace{\cancel{\left.\prtl{\Delta M\circ\pi}{\bz}\right|_{\bz_{\text{op}}}(-\bz_{\text{op}}^{\wedge}\delta\boldsymbol{\varphi})}}_{\text{higher order term}}.
\end{align}

Next, apply the chain rule of function composition
(for photometric function $f$ and geometric function $\pi$) 
\[
\prtl{(f\circ\pi)}{\bz} = (\nabla f)^{\top}\prtl{\pi}{\bz},
\]
to simplify the term that is linear in $\delta\boldsymbol{\varphi}$,
\begin{align}
\left.\prtl{(M_{\text{op}}\circ\pi)}{\bz}\right|_{\bz_{\text{op}}} \bz_{\text{op}}^{\wedge}\delta\boldsymbol{\varphi}
& = \left(\nabla M_{\text{op}}(\bp_{\text{op}}(t))\right)^{\top}\left.\prtl{\pi}{\bz}\right|_{\bz_{\text{op}}} \bz_{\text{op}}^{\wedge}\delta\boldsymbol{\varphi} \nonumber\\
& = \left(\nabla M_{\text{op}}(\bp_{\text{op}}(t))\right)^{\top}\mE_{\text{op}}\delta\boldsymbol{\varphi},
\end{align}
and obtain
\begin{align}
M(\bp(t)) 
 & \approx M_{\text{op}}(\bp_{\text{op}}(t))-\underbrace{\left(\nabla M_{\text{op}}(\bp_{\text{op}}(t))\right)^{\top}\mE_{\text{op}}\delta\boldsymbol{\varphi}}_{\text{linear in }\delta\boldsymbol{\varphi}} \nonumber\\
 &\quad +\underbrace{\Delta M(\bp_{\text{op}}(t))}_{\text{linear in }\Delta M}.
 \label{eq:LinearizationMatp}
\end{align}

The next step consists of relating $\delta\boldsymbol{\varphi}$ and $\Delta M$ to the parameterizing
perturbations $\Delta\bP_{\balpha}$ and $\Delta\bP_{\bbeta}$ in \eqref{eq:Perturb}--\eqref{eq:PerturbVecPartitioning}.
\begin{itemize}
\item $\Delta\bP_{\balpha}$: We use \eqref{eq:BarfootdeltaPerturbRotAtt} to write $\delta\boldsymbol{\varphi}$ in terms of the perturbations of the control poses $\{\delta\boldsymbol{\phi}_{j}\}$ and the incremental angle $\boldsymbol{\Delta\phi}_{i,i+1}=(\log(\Rot_{i+1}\Rot_{i}^{-1}))^{\vee}$.
Letting $\bq_{\text{op}}^{\top}(t) \doteq \left(\nabla M_{\text{op}}(\bp_{\text{op}}(t))\right)^{\top}\mE_{\text{op}}(t)$,
we have:
\begin{align}
\bq_{\text{op}}^{\top}(t) \delta\boldsymbol{\varphi} 
& = \underbrace{\bq_{\text{op}}^{\top}(t)(\boldsymbol{1}-\boldsymbol{A}(u(t),\boldsymbol{\Delta\phi}_{i,i+1}))}_{1\times3}\delta\boldsymbol{\phi}_{i} \nonumber\\
& \quad +\underbrace{\bq_{\text{op}}^{\top}(t)\boldsymbol{A}(u(t),\boldsymbol{\Delta\phi}_{i,i+1})}_{1\times3}\delta\boldsymbol{\phi}_{i+1}.
\end{align}

\item $\Delta\bP_{\bbeta}$: (Specifying $\Delta M(\bp_{\text{op}}(t))$ in terms of $\Delta\bP_{\bbeta}$). 
Assuming $\Delta M$ is an image of the same size as $M$, the pixels of $M$ have values $\bbeta$ 
(e.g., $\bbeta$ is a ``flattened'' version of $M$) and the pixels of $\Delta M$ have values $\Delta\bbeta$. 
Then, $\Delta M(\bp_{\text{op}}(t))$ is an incremental brightness at map point $\bp_{\text{op}}(t)$, which we write in terms of the values of $\Delta M$.
For implementation simplicity, we use just one value (i.e., nearest neighbor map point), instead of four (in bilinear interpolation). 
\end{itemize}

\subsection{Linearization of the error at each event (i.e., two points)}
\label{sec:LinearizePerEventError}

We may now derive the linearization for the two terms involved in computing the photometric error \eqref{eq:ObsOrigEGM} at each event.
Substituting \eqref{eq:LinearizationMatp} into \eqref{eq:ObsOrigEGM} gives,
\begin{align}
\epsilon_{k} & \approx M_{\text{op}}(\bp_{\text{op}}(t_{k}))-\bq_{\text{op}}^{\top}(t_{k})\delta\boldsymbol{\varphi}+\Delta M(\bp_{\text{op}}(t_{k}))\nonumber \\
 & \quad -M_{\text{op}}(\bp_{\text{op}}(t_{k}-\Delta t_{k}))+\bq_{\text{op}}^{\top}(t_{k}-\Delta t_{k})\delta\tilde{\boldsymbol{\varphi}} \nonumber \\
 & \quad -\Delta M(\bp_{\text{op}}(t_{k}-\Delta t_{k})) -\pol_{k} C, \nonumber
\end{align}
which yields \eqref{eq:LinErrorOrigEGM}.

\ifclearsectionlook\cleardoublepage\fi
\bibliographystyle{IEEEtran_link}

\vspace{-3.5ex}
\begin{IEEEbiography}
[{\includegraphics[width=1in,height=1.25in,clip,keepaspectratio]{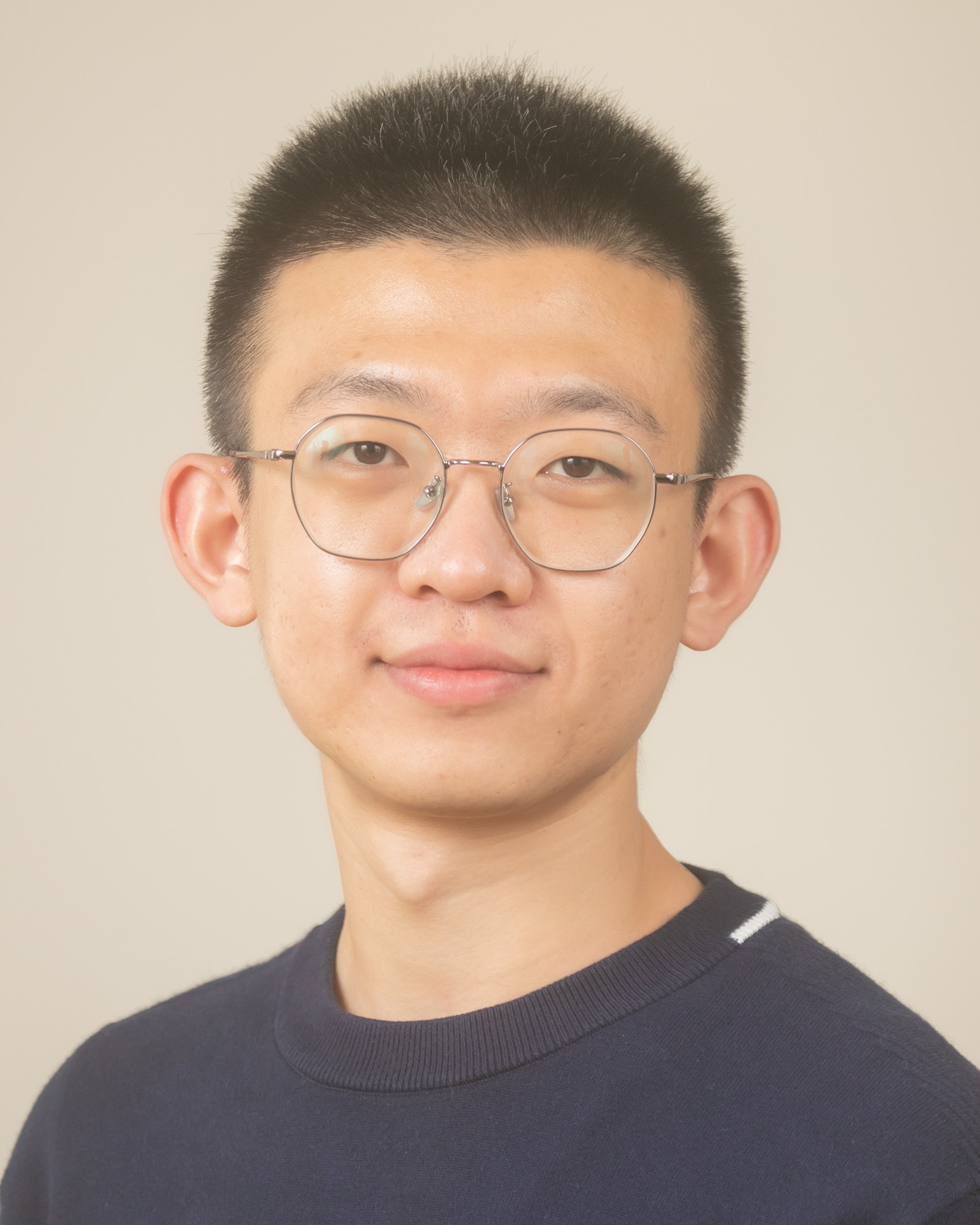}}]{Shuang Guo}
received the B.Sc. and M.Sc. degrees in aerospace engineering from Harbin Institute of Technology, China, in 2019 and 2021, respectively.
He is currently working towards the Ph.D. degree in the Robotic Interactive Perception Laboratory at Technische Universit\"at Berlin, 
in the Dept. of Electrical Engineering and Computer Science.
His research interests lie at the intersection of computer vision, event-based vision, deep learning and robotics.
\end{IEEEbiography}

\vspace{-4ex}
\begin{IEEEbiography}[{\includegraphics[width=1in,height=1.25in,clip,keepaspectratio]{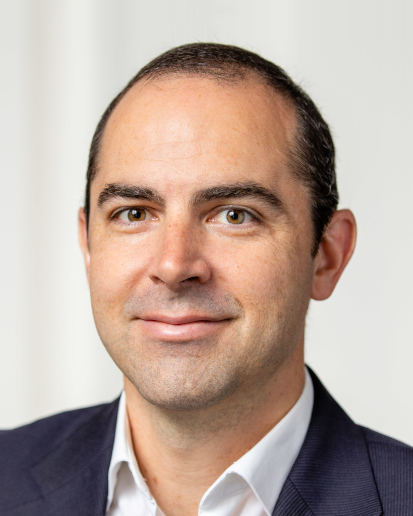}}]{Guillermo Gallego} (SM'19) 
is Full Professor at Technische Universit\"at Berlin, in the Dept. of EECS, 
and at the Einstein Center Digital Future, Berlin, where he leads the Robotic Interactive Perception Laboratory.
He is also a Principal Investigator at the Science of Intelligence Excellence Cluster 
and the Robotics Institute Germany.
He received the PhD degree in Electrical and Computer Engineering from the Georgia Institute of Technology, USA, in 2011, supported by a Fulbright Scholarship.
From 2011 to 2014 he was a Marie Curie researcher with Universidad Politecnica de Madrid, Spain, and from 2014 to 2019 he was a postdoctoral researcher at the Robotics and Perception Group, University of Zurich and ETH Zurich, Switzerland.
He serves as Associate Editor for IEEE T-PAMI, RA-L and IJRR, and guest Editor for IEEE T-RO.
His research interests include robotics, perception, %
optimization and geometry. 
\end{IEEEbiography}

\else
\ifclearsectionlook\cleardoublepage\fi
\bibliographystyle{IEEEtran}
\bibliography{all}

\vspace{-3.5ex}
\begin{IEEEbiography}
[{\includegraphics[width=1in,height=1.25in,clip,keepaspectratio]{images/authors/Guo}}]{Shuang Guo}
received the B.Sc. and M.Sc. degrees in aerospace engineering from Harbin Institute of Technology, China, in 2019 and 2021, respectively.
He is currently working towards the Ph.D. degree in the Robotic Interactive Perception Laboratory at Technische Universit\"at Berlin, 
in the Dept. of Electrical Engineering and Computer Science.
His research interests lie at the intersection of computer vision, event-based vision, deep learning and robotics.
\end{IEEEbiography}

\vspace{-4ex}
\begin{IEEEbiography}[{\includegraphics[width=1in,height=1.25in,clip,keepaspectratio]{images/authors/Gallego}}]{Guillermo Gallego} (SM'19) 
is Full Professor at Technische Universit\"at Berlin, in the Dept. of EECS, 
and at the Einstein Center Digital Future, Berlin, where he leads the Robotic Interactive Perception Laboratory.
He is also a Principal Investigator at the Science of Intelligence Excellence Cluster 
and the Robotics Institute Germany.
He received the PhD degree in Electrical and Computer Engineering from the Georgia Institute of Technology, USA, in 2011, supported by a Fulbright Scholarship.
From 2011 to 2014 he was a Marie Curie researcher with Universidad Politecnica de Madrid, Spain, and from 2014 to 2019 he was a postdoctoral researcher at the Robotics and Perception Group, University of Zurich and ETH Zurich, Switzerland.
He serves as Associate Editor for IEEE T-PAMI, RA-L and IJRR, and guest Editor for IEEE T-RO.
His research interests include robotics, perception, %
optimization and geometry. 
\end{IEEEbiography}

\clearpage 

\fi

\end{document}